\documentclass{article} 
\usepackage{iclr2027_conference,times}

\usepackage{hyperref}
\usepackage{url}
\usepackage[utf8]{inputenc} 
\usepackage[T1]{fontenc}    
\usepackage{hyperref}       
\usepackage{url}            
\usepackage{booktabs}       
\usepackage{amsfonts}       
\usepackage{nicefrac}       
\usepackage{microtype}      
\usepackage[table,dvipsnames,svgnames,x11names]{xcolor}         
\usepackage{graphicx}
\usepackage{wrapfig}
\usepackage{float}
\usepackage{capt-of}
\usepackage{amsmath}
\usepackage{makecell}

\usepackage[utf8]{inputenc}
\usepackage{hyperref}
\usepackage{fancyhdr}
\usepackage{enumitem}
\usepackage{longtable}
\usepackage{booktabs}
\usepackage{tabularx}
\usepackage{array}
\usepackage[most]{tcolorbox}
\usepackage{listings}
\usepackage{xurl}
\usepackage{etoc}

\setlist[itemize]{topsep=4pt, itemsep=2pt, leftmargin=1.5em}
\renewcommand{\arraystretch}{1.10}

\definecolor{Accent}{HTML}{1F4E79}
\definecolor{AccentSoft}{HTML}{EEF5FB}
\definecolor{TableHead}{HTML}{DDECF7}
\definecolor{TableStripe}{HTML}{F5F9FC}
\definecolor{MutedBorder}{HTML}{C8CDD4}
\definecolor{WarnFill}{HTML}{FFF8E6}
\definecolor{EffectFill}{HTML}{FDEEEE}

\newcommand{\filepath}[1]{\texttt{\scriptsize\detokenize{#1}}}
\newcommand{\codes}[1]{\textsf{#1}}

\newcolumntype{L}[1]{>{\raggedright\arraybackslash}p{#1}}
\newcolumntype{C}[1]{>{\centering\arraybackslash}p{#1}}

\renewcommand{\arraystretch}{1.05}

\definecolor{Accent}{HTML}{1F4E79}
\definecolor{AccentSoft}{HTML}{EEF5FB}
\definecolor{SystemFill}{HTML}{EEF5FD}
\definecolor{UserFill}{HTML}{EEF9F0}
\definecolor{SchemaFill}{HTML}{FFF4E8}
\definecolor{TaskFill}{HTML}{F4EEFB}
\definecolor{AlignFill}{HTML}{ECFAF8}
\definecolor{NegativeFill}{HTML}{FDEEEE}
\definecolor{CodeFill}{HTML}{F6F7F9}
\definecolor{MutedBorder}{HTML}{C8CDD4}

\lstdefinestyle{promptstyle}{
  basicstyle=\ttfamily\small,
  breaklines=true,
  breakatwhitespace=false,
  columns=fullflexible,
  keepspaces=true,
  showstringspaces=false,
  upquote=true,
  tabsize=2,  
  breakautoindent=false,
  breakindent=0pt
}

\tcbset{
  prompt/base/.style={
    enhanced,
    breakable,
    listing engine=listings,
    listing only,
    frame hidden,
    boxrule=0pt,
    arc=2mm,
    outer arc=2mm,
    left=2mm,
    right=2mm,
    top=1.3mm,
    bottom=1.2mm,  
    listing options={style=promptstyle},
    borderline west={1.5mm}{0pt}{Accent},
    colback=CodeFill
  },
  prompt/system/.style={prompt/base, colback=SystemFill, borderline west={1.5mm}{0pt}{blue!70!black}},
  prompt/user/.style={prompt/base, colback=UserFill, borderline west={1.5mm}{0pt}{green!45!black}},
  prompt/schema/.style={prompt/base, colback=SchemaFill, borderline west={1.5mm}{0pt}{orange!85!black}},
  prompt/task/.style={prompt/base, colback=TaskFill, borderline west={1.5mm}{0pt}{violet!70!black}},
  prompt/align/.style={prompt/base, colback=AlignFill, borderline west={1.5mm}{0pt}{teal!70!black}},
  prompt/wrapper/.style={prompt/base, colback=AccentSoft, borderline west={1.5mm}{0pt}{Accent}},
  prompt/negative/.style={prompt/base, colback=NegativeFill, borderline west={1.5mm}{0pt}{red!70!black}},
  prompt/code/.style={prompt/base, colback=CodeFill, borderline west={1.5mm}{0pt}{black!45}},
  sourcebox/.style={
    enhanced,
    breakable,
    colback=black!2,
    colframe=MutedBorder,
    boxrule=0.45pt,
    arc=1.6mm,
    left=1.5mm,
    right=1.5mm,
    top=0.8mm,
    bottom=0.8mm
  }
}

\newtcblisting{PromptBlock}[1][]{prompt/code,#1}
\newtcolorbox{NoteBox}{enhanced, breakable, colback=yellow!10, colframe=yellow!45!black, boxrule=0.6pt, arc=2mm, left=2mm, right=2mm, top=1.2mm, bottom=1.2mm}

\title{A Simulation-Grounded Agentic VLM Framework for Wildfire Monitoring and Reporting}

\author{%
\parbox[t]{\dimexpr\textwidth-2\tabcolsep\relax}{%
    \normalfont
    \textbf{Duowen Chen\textsuperscript{1},
    Yuchen Sun\textsuperscript{1},
    Zhiqi Li\textsuperscript{1},
    Yuxuan Liao\textsuperscript{1}},
    \textbf{Sinan Wang\textsuperscript{1},\\
    Bart van Bloemen Waanders\textsuperscript{2},
     Bo Zhu\textsuperscript{1}}\\[4pt]
    \textsuperscript{1}Georgia Institute of Technology
    \quad
    \textsuperscript{2}Sandia National Laboratories
}%
}
\iclrfinalcopy 
\begin{document}
\etocdepthtag.toc{main}

\maketitle

\begin{abstract}
Effective wildfire monitoring requires relating visual evidence to physical fire dynamics, yet real videos with synchronized physical annotations are scarce and high-fidelity 3D simulation is costly. We present a simulation-grounded vision-language model (VLM) framework that automatically converts 2D wildfire simulations into labeled video episodes. A fixed Blender mapping produces low-detail 3D proxies aligned with simulator terrain, fuel layout, fire activity, and wind cues; controllable video generation supplies richer appearance. The proxies are intermediate representations rather than finely rendered final scenes. Generated videos and simulator labels form reusable multimodal memory for a training-free multi-agent VLM system that retrieves reference episodes, reconciles visual and memory-based predictions, and produces structured wildfire reports. On held-out generated episodes, video memory achieves $51.5\%$ exact four-tag accuracy, compared with $22.6\%$ for direct VLM querying and $16$--$17\%$ for text-only memory; the complete system achieves $77.3\%$ accuracy on six simulator-derived report fields. Component ablations, cross-generator tests, and three real-UAV evaluations assess retrieval, reporting, generator changes, and observable monitoring tasks. The framework connects automatic simulation-to-proxy conversion with memory-based VLM reasoning under scarce real-world physical annotations.
\end{abstract}

\section{Introduction}

Wildfire monitoring and reporting are fundamentally about connecting
\emph{what is seen} to \emph{what is happening physically} in the
wildland. Fire-spread simulators (e.g.,
\cite{mandel2011coupled,finney1998farsite,mallet2009modeling,Doyle_SimFire_2024})
encode how fuel, slope, moisture, and wind shape propagation, and
satellite, aerial, and UAV measurements provide partial visual
observations of active incidents. Modern vision-language models
(VLMs) need a framework to align two forms of evidence in a shared
representation. Real wildfire videos are visually rich, yet labels
for wind, moisture, slope, fuel condition, burn extent, and severe-fire
behavior are usually absent. Simulators expose these variables
directly, yet their native outputs are grid states, masks, and fire
fronts, which differ from the aerial-view videos that VLMs are
expected to interpret.

Though existing wildfire VLM work has advanced fire and smoke
recognition, UAV monitoring, satellite risk analysis, and VLM-guided
planning~\cite{du2025wildfire,wu2025fireclip,seidel2025advancing,
ayanzadeh2026wildfirevlm,webb2026fire,xie2025marsha,sandeep2026context},
temporally rich videos paired with physical fire-state labels remain
scarce. Collecting synchronized wind, moisture, slope, fuel, and
fire-behavior annotations is hazardous, expensive, and hard to
control. Available UAV datasets provide valuable detection and
segmentation annotations~\cite{shamsoshoara2021aerial,
kularatne2024fireman,pesonen2025boreal}, but the annotations used here
do not jointly describe these drivers. Fire-spread simulators expose
such conditions without providing camera-view observations, while
full 3D fire simulation~\cite{kokosza2024scintilla,hadrich2021fire}
incurs costs that limit coverage across fuels, terrains, weather,
and cameras. We therefore study a scalable middle ground: efficient
2D simulations over real terrain and fuel maps are automatically
lifted into lightweight Blender proxies, translated into controlled
videos, and stored as physics-labeled memory for VLM reasoning.

\begin{figure*}
    \centering
    \includegraphics[width=0.7\linewidth]{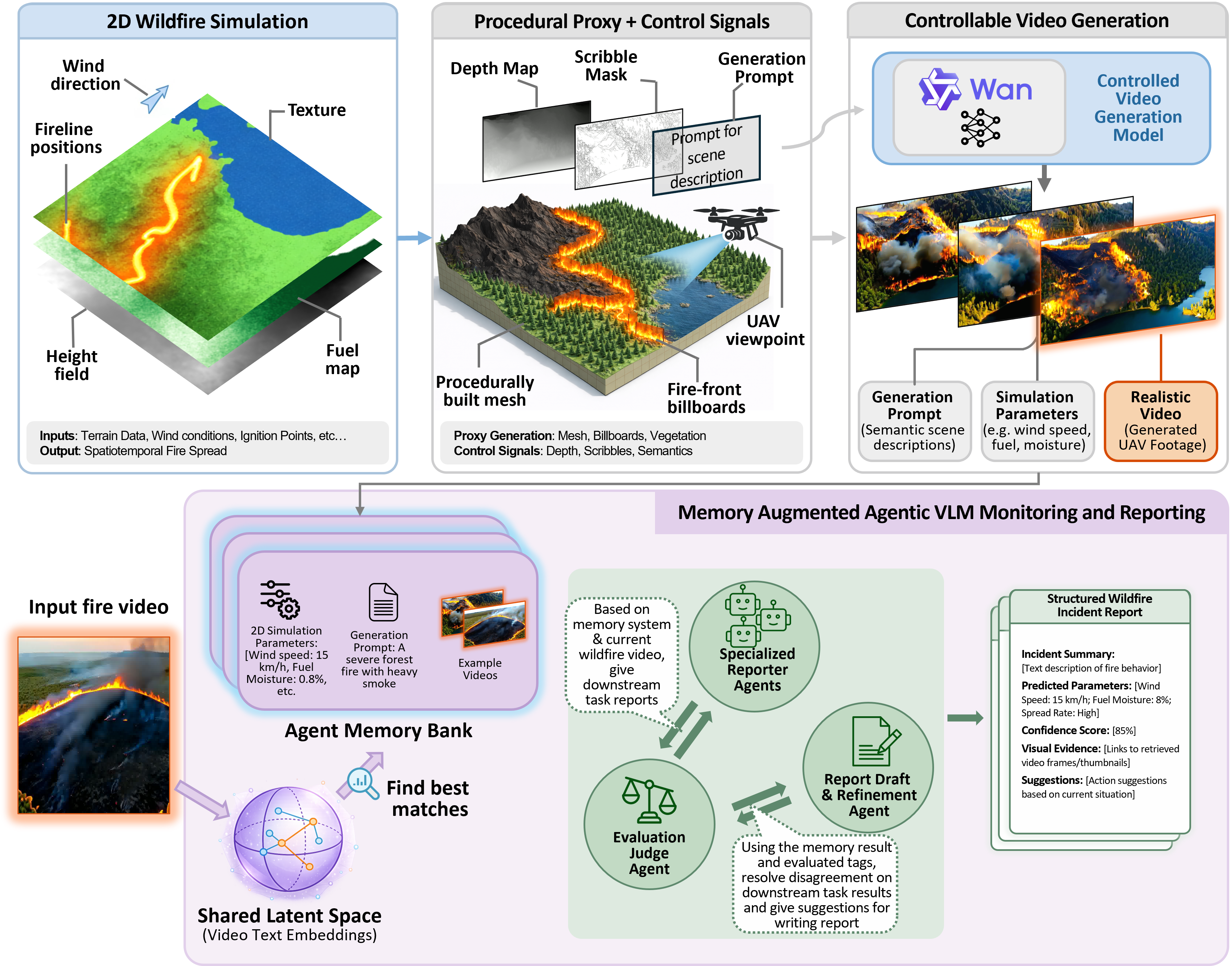}
    \caption{Overall simulation-grounded VLM framework. Top: 2D simulation and procedural 3D proxies provide depth, scribble, and prompt controls for synthetic aerial-view episodes with simulator-derived labels. Bottom: a memory-augmented multi-agent VLM reporter combines query frames, retrieved labeled references, and disagreement adjudication for wildfire monitoring and reporting.}
    \label{fig:placeholder}
    \vspace{-15pt}
\end{figure*}

Recent developments such as Code World
Model~\citep{chen2026codeworldmodel} and Blender scene authoring with
GPT-6 Astra highlight the potential of
executable representations for visual world construction. We explore
this direction in a simulation-grounded VLM framework for wildfire
monitoring and reporting (Figure~\ref{fig:placeholder}). Each episode
starts with a 2D simulation over real LANDFIRE terrain and fuel data
spanning approximately \(3\,\mathrm{km}\). A fully automatic Blender
procedure maps evolving simulator fields into terrain, vegetation,
flame, and wind-driven smoke cues. This lightweight proxy is an
\emph{intermediate representation}: fixed conversion rules maintain
simulator-to-scene correspondence, while depth, scribble, and prompt
controls guide richer video appearance. Generated clips retain their
prompts, simulator parameters, environmental tags, and fire-behavior
labels as reusable multimodal memory. Retrieved episodes then
complement direct visual reasoning for interpreting new wildfire
videos without updating VLM weights. Section~\ref{sec:discussion}
further discusses the connections to code-based world construction
and general-purpose scene authoring.

Generated videos with known simulator labels provide controlled
evaluation targets. We use 500 geo-disjoint locations: 400 episodes
populate memory and 100 are held out for evaluation
(Appendix~\ref{sec:downstream_tasks}). For each query, retrieved
references complement direct visual reasoning, and a judge resolves
reporter disagreements. Video memory achieves 51.5\% exact four-tag
accuracy versus 16--17\% for text-only memory and 22.6\% for direct
VLM querying; reporting reaches 77.3\% on six simulator-derived
fields. Structured records retain predictions, evidence, and
reasoning. Cross-generator and real-UAV tests additionally assess
generator changes, smoke/fire detection, and visible fire-area
assessment.

Our contributions are:
\begin{itemize}
    \item \textbf{Automatic simulation-grounded visual evidence.}
    We connect efficient 2D wildfire simulation to controllable
    video generation through a fully automatic Blender proxy.
    A fixed mapping expresses terrain, fuel layout, fire activity,
    and wind cues as lightweight 3D controls, constructing labeled
    visual episodes under scarce real-world physical annotations.

    \item \textbf{Memory-augmented agentic reporting.}
    We organize generated videos and simulator context as reusable
    multimodal memory, and combine retrieval and reranking with
    complementary VLM reporters and selective disagreement
    adjudication. The training-free system produces
    environmental-driver estimates and structured wildfire reports
    without updating model weights.

    \item \textbf{Controlled evaluation with complementary real-UAV
    tests.}
    Video memory achieves 51.5\% exact four-tag accuracy versus
    22.6\% for direct VLM querying and 16--17\% for text-only memory;
    the full reporter reaches 77.3\% on six simulator-derived
    fields. Ablations, cross-generator tests, and real-UAV
    perceptual evaluations examine the framework's components and
    transfer beyond the primary generated-video setting.
\end{itemize}

\section{Related Work}

\paragraph{Wildfire modeling and observation.}
Wildfire prediction combines spread models with sensing. Rothermel~\cite{rothermel1972mathematical} underlies FARSITE~\cite{finney1998farsite}; level-set methods handle fire-front topology changes~\cite{mallet2009modeling}; WRF-SFIRE couples spread with atmospheric dynamics~\cite{mandel2011coupled}. MODIS and VIIRS provide broad-area fire observations~\cite{justice2002modis,giglio2016collection,schroeder2014new}, complemented by UAV smoke/fire detection and front tracking~\cite{bailon2022real,raoufi2025fire,fan2025real}. Rather than replacing simulators or sensors, we automatically translate simulator states into visual reference episodes for VLM monitoring and reporting.

\paragraph{VLMs for scientific and wildfire understanding.}
Scientific VLM benchmarks show that general-purpose multimodal models can be useful but struggle with domain-specific visual structure, terminology, and evaluation protocols~\cite{esparza2025automated, roberts2024scifibench,wu2024scimmir,kuckreja2024geochat,li2023llava}. Wildfire VLM systems have been studied for scene recognition~\cite{du2025wildfire}, smoke/fire detection~\cite{wu2025fireclip,qi2026smokebench}, UAV situational awareness~\cite{seidel2025advancing}, satellite-image risk reasoning~\cite{ayanzadeh2026wildfirevlm}, and VLM-guided reinforcement learning in wildfire digital twins~\cite{webb2026fire}. These works motivate VLM-based decision support, but they also highlight the need for physically grounded video data and task-specific evaluation.

\vspace{-10pt}
\paragraph{Code world models and simulation-conditioned video.}
WorldCoder~\cite{tang2024worldcoder} constructs executable world models through interaction, while Code World Model~\citep{chen2026codeworldmodel} connects coding-agent-driven state evolution to proxy-conditioned video generation. PhysGen~\cite{liu2024physgen} combines rigid-body simulation with generative video refinement, and Cosmos-Transfer1~\cite{nvidia2025cosmostransfer1} supports spatially controlled generation. Together, these approaches motivate separating structured dynamics and spatial controls from visual appearance. We examine this separation for simulation-grounded wildfire memory; Section~\ref{sec:discussion} discusses its relationship to coding-agent world models and scene authoring.

\vspace{-10pt}
\paragraph{Memory-augmented multimodal reasoning.}
Retrieval-augmented generation supplies external evidence without updating model weights~\cite{lewis2020retrieval}. Agent memories support long-horizon reasoning~\cite{packer2023memgpt,park2023generative}, while multimodal RAG and video retrieval access visual, textual, and scene-level evidence~\cite{jeong2025videorag,luo2024videorag,gia2025vrag,zeng2026scenerag,fu2026videostir} using video--text embeddings~\cite{xu2021videoclip,luo2022clip4clip,wang2022internvideo,ma2022xclip}. WorldMem~\cite{xiao2025worldmem} and PhysStream~\cite{chen2026physstream} use memory for consistent visual generation. Our memory instead pairs reference videos with simulator-controlled conditions and fire-behavior labels for inference across episodes, complementing direct observations and historical-data or context-aware wildfire systems~\cite{xie2025marsha,sandeep2026context}.

\vspace{-3.5mm}
\section{Method}
\vspace{-3.5mm}

Figure~\ref{fig:placeholder} gives the overall workflow. The framework has two main components: a simulation-grounded visual evidence construction method and a memory-augmented multi-agent VLM system for monitoring and reporting. The subsections below follow this order.

\subsection{Simulation-Grounded Visual Evidence Construction}
\label{sec:sim_to_video}

The first stage constructs labeled visual episodes to reduce reliance on scarce real footage with synchronized physical annotations. We retain the scalability and annotations of 2D wildfire simulation while making its outputs visually accessible to VLMs: diverse episodes over real terrain and fuel maps are lifted into procedural 3D proxies and translated into aerial-view clips. Here, \emph{simulation-grounded} refers to the simulator-derived labels and controls linking each episode to its originating scenario; visual and temporal cue alignment is evaluated separately in Section~\ref{sec:simulation_alignment}.

\subsubsection{Physics-Grounded 2D Wildfire Simulation}
\label{sec:2d_simulation}
To balance physical grounding and generation throughput, we use SimFire~\cite{Doyle_SimFire_2024} as our 2D wildfire simulation platform. SimFire follows the classical Rothermel fire-spread model~\cite{rothermel1972mathematical} detailed in Appendix~\ref{sec:rothermel_app}, allowing us to generate large numbers of wildfire episodes with meaningful macroscopic spread behavior. For each episode, we randomly sample a geographic location specified by latitude and longitude, and query the LANDFIRE database to obtain real U.S. fuel and terrain information. Fuel types are represented using the Anderson 13 fuel model \cite{anderson1982aids}, while terrain elevation is used to derive slope and support later 3D proxy construction. Wind direction, wind-related parameters, and fuel-moisture settings are randomized to increase scenario diversity. Each simulation covers a region of approximately \(3\,\mathrm{km}\) in spatial extent. We refer readers to Appendix~\ref{sec:firemodel_app} and \ref{sec:implementation_details} for implementation details.

Each episode is summarized using four physically meaningful factors: wind speed, moisture, slope angle, and vegetation density. Each factor is discretized into low and high levels, yielding a compact set of simulator-derived tags. These tags provide coarse environmental regimes for retrieval-based state estimation and agent reasoning, rather than continuous measurements of a query incident. Tag thresholds are specified in Appendix~\ref{sec:implementation_details}; the six downstream report-field definitions appear in Appendix~\ref{sec:downstream_tasks}.

\vspace{-1.5mm}
\subsubsection{Procedural 3D Proxy}
\label{sec:procedural_proxy}
\vspace{-1.5mm}
As shown in Figure~\ref{fig:placeholder}, the 2D simulator produces fire masks, fuel maps, terrain fields, and fire-front states, but these outputs are simple 2D projections onto the terrain topography and therefore do not provide a 3D visual image similar to UAV video footage. We therefore design a Blender pipeline that automatically construct a lightweight 3D proxy scene for each simulation. Elevation is converted into a height-field terrain mesh; Anderson 13 fuel categories are mapped to procedural vegetation and terrain textures; burning cells are augmented with flame billboards; and smoke billboards are oriented and animated according to wind speed and direction. The proxy is not intended to be photorealistic. Its role is to encode simulator-derived spatial structure---terrain shape, vegetation distribution, active fire location, and wind-conditioned smoke direction---in a form that can guide video generation. For details of fuel models mapping with billboard assets, we refer readers to Appendix~\ref{sec:anderson_app}.

\subsubsection{Controllable Video Generation}
\label{sec:2d_3d}

\begin{figure*}[t]
    \centering
    \includegraphics[width=1.0\linewidth]{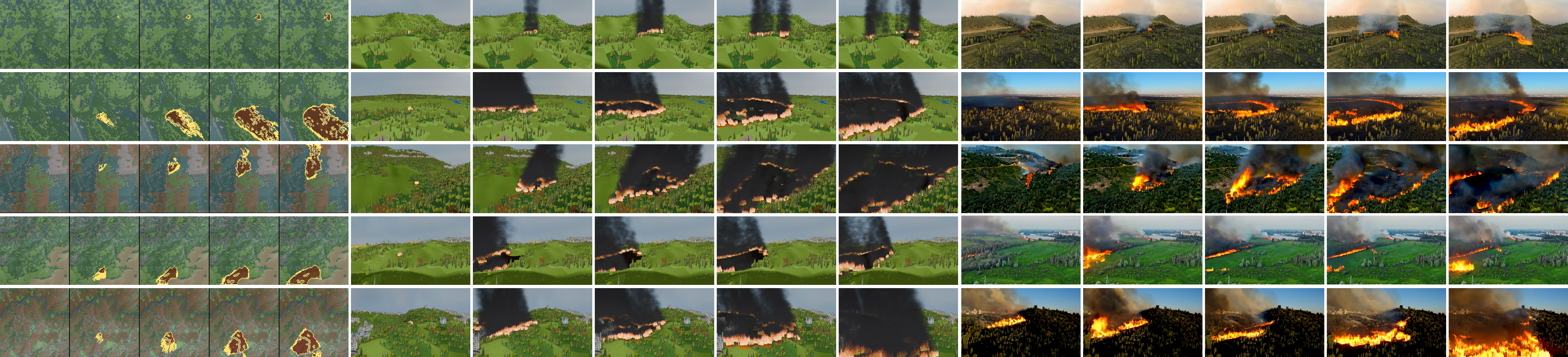}
    \caption{Simulation-grounded synthetic wildfire episodes. Left: SimFire propagation under different settings and locations. Middle: procedural 3D billboard rendering from an aerial viewpoint. Right: generated videos conditioned on the proxy renderings. Episode labels come from the originating simulations.}
    \label{fig:2d_to_3d}
\end{figure*}

The proxy rendering encodes simulator structure but remains visually simple. We use controllable video generation to translate it into a visually richer wildfire clip. Following VACE~\cite{jiang2025vace}, the generator is conditioned on a depth sequence~\cite{ranftl2020towards, yang2024depth}, a scribble sequence~\cite{chan2022learning}, and a text prompt. 
The depth and scribble controls guide geometry, vegetation layout, fire-front location, and camera-consistent structure, while the prompt specifies appearance-level information such as smoke density, flame visibility, vegetation type, and wind-driven motion. The prompt is synthesized from simulation metadata and then refined by a VLM (see Appendix~\ref{sec:video_generation_prompts} for details). We also use parameter-conditioned negative prompts to suppress conflicting attributes, such as strong wind for low-wind cases or still smoke for high-wind cases. Each generated video retains its originating simulator labels. Its observable cues reflect both the proxy controls and metadata-conditioned prompts; cue agreement is assessed rather than assumed.
\vspace{-1.5mm}
\subsection{Memory-Augmented Agentic VLM Monitoring and Reporting}
\vspace{-1.5mm}

\begin{figure*}[t]
    \centering
    \includegraphics[width=1.0\linewidth]{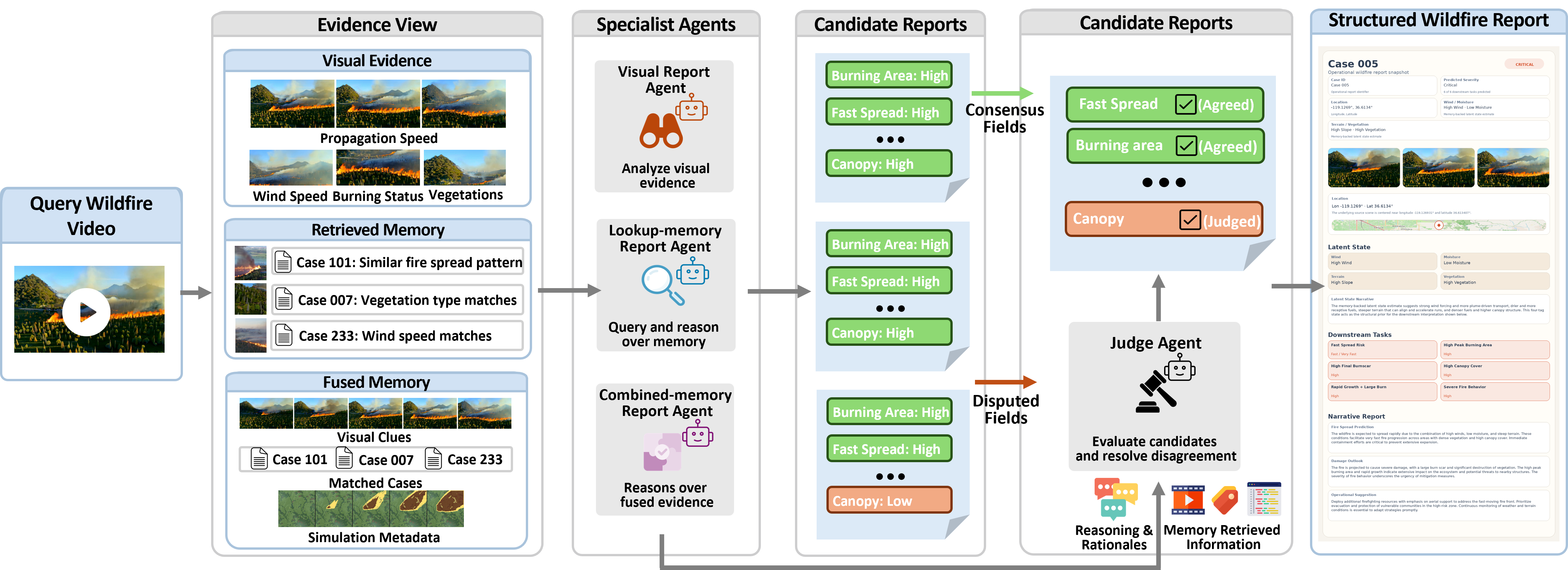}
    \caption{Agentic VLM reporting pipeline. A query wildfire video is analyzed using direct visual evidence and retrieved physics-labeled memory evidence. Agreement across reporters is written to the final report, while disagreement is routed to a judge. The resolved predictions are rendered as a structured HTML report.}
    \label{fig:multiagent_reporting}
    \vspace{-10pt}
\end{figure*}

The second stage uses the generated video episodes as physics-labeled visual memory for wildfire state estimation and reporting. The design goal is to keep the system training-free: new simulator episodes update the memory bank, not the VLM weights. As shown in Figure~\ref{fig:multiagent_reporting}, given a query wildfire video, the system retrieves visually and semantically similar memory episodes, uses their simulator labels as reference priors, and combines these priors with direct visual reasoning in a multi-agent report.

\vspace{-1.5mm}
\subsubsection{Physics-Labeled Multimodal Memory}
\label{sec:agentic_memory}
\vspace{-1.5mm}
Each memory entry is a key--value item. The key is a multimodal embedding of the generated wildfire video and its simulation-derived text description. The value stores the generated video, the prompt used to produce it, the 2D simulation parameters, the environmental tags, and downstream fire-behavior labels. This structure makes the memory both searchable and interpretable: retrieval returns not only similar videos, but also the simulator conditions associated with those reference episodes, not ground-truth measurements of the query.

At query time, the same multimodal embedder maps the input video into the shared embedding space. We compute cosine similarity between the query embedding and each memory key and retrieve the top-\(k\) candidate episodes. The retrieved examples are then used as in-context evidence for estimating the query's wind, moisture, slope, and vegetation tags. In practice (see Section~\ref{sec:ablation_agent}), neither the agent alone nor the embedder alone yields sufficiently accurate classification across these attributes. We therefore complement embedding-based retrieval with candidate reranking before using retrieved labels for query estimation.

\subsubsection{Memory-Augmented Multi-Agent Reporting}
\label{sec:method_multiagent_reporting}

We formulate downstream wildfire understanding as an \emph{agentic VLM reporting} problem combining query observations with retrieved physics-labeled references. Given a query video, the system produces a structured report over fire-behavior fields such as spread risk, burn extent, vegetation condition, and severe-fire indicators. Rather than relying on a single model, we separate report generation into four stages: memory retrieval, multi-agent reporting, disagreement adjudication, and report rendering \cite{lewis2020retrieval, wu2024autogen, chan2023chateval}.

For a query video $q$, we retrieve the top-$k$ most relevant episodes from the memory split. Retrieved episodes supply environmental driver tags, physical summaries, and downstream labels. For each binary report field $t$, the retrieved examples define a score-weighted memory consensus for adjudication:
$
C_t(y \mid q)
=
\frac{\sum_{i=1}^{k} w_i \mathbf{1}[\ell_t(r_i)=y]}
     {\sum_{i=1}^{k} w_i},
$
where $r_i$ is a retrieved episode, $w_i$ is its retrieval score, and $\ell_t(r_i)$ is its label for field $t$. This consensus summarizes the label favored by similar memory episodes and their score-weighted agreement; it is a reference prior rather than a calibrated probability for the query.

The reporting stage uses three agents with complementary evidence. The visual reporter receives only sampled query frames and focuses on directly observable cues. The lookup reporter receives the query frames and a train-derived lookup prior from the top retrieved tag bundle. The combined-memory reporter additionally receives the retrieved episode's simulator summaries. This decomposition keeps disagreements across evidence sources explicit across the reporting pipeline, in line with multi-agent LLM/VLM workflow designs \cite{wu2024autogen,chan2023chateval}.

Reporter agreement determines which fields require adjudication. If all reporters agree on a field, the shared prediction is copied into the final report. If they disagree, only that disputed field is routed to a judge agent~\cite{zheng2023judging, wang2022self}, which sees the query frames, reporter predictions, rationales, top-$k$ memory consensus, and retrieved metadata summaries. 
The inputs and routing protocol are detailed in Appendix~\ref{sec:impl_multiagent_reporting}.

Finally, the resolved predictions are compiled into an HTML report. The structured output records sampled visual evidence, predicted labels, routing status, reporter rationales, and retrieved memory summaries. A separate narrative stage summarizes the predicted tags and report fields. These records support inspection of the inference process; narrative response suggestions are illustrative outputs, not validated operational guidance.

\begin{figure}[t]
\centering

\begin{minipage}[t]{0.43\textwidth}
    \centering
    \vspace{0pt}
    \includegraphics[width=\linewidth]{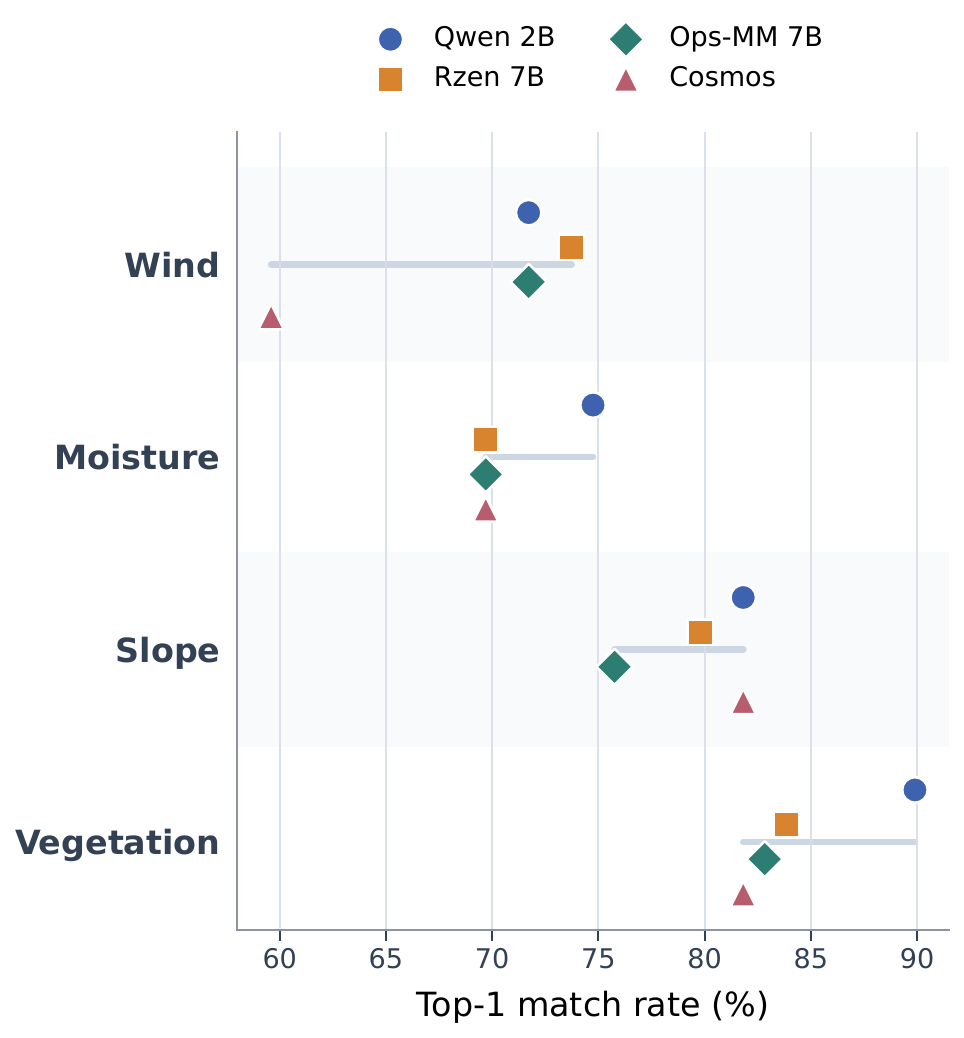}
    \caption{We show the effects of different multimodal embedding models on tag retrieval accuracy. From experiments, Qwen3-2B-VL gives the best retrieval accuracy in most cases which leads to our choice of it being our embedding model.}
    \label{fig:embed_models}
\end{minipage}
\hfill
\begin{minipage}[t]{0.53\textwidth}
    \centering
    \vspace{0pt}
    \captionof{table}{Reporting-component ablation on six simulator-derived fields. Oracle Upper Bound counts a field as correct when any constituent reporter predicts its label correctly; it is not an executable selection rule.}
    \label{tab:method_accuracy}
    \small
    \setlength{\tabcolsep}{4pt}
    \renewcommand{\arraystretch}{1.05}
    \begin{tabular}{@{}>{\raggedright\arraybackslash}p{0.68\linewidth}c@{}}
    \toprule
    \textbf{Method} & \makecell{\textbf{Downstream}\\\textbf{Task Accuracy}} \\
    \midrule
    No Judge Agent (Majority Vote) & 72.6\% \\
    No Lookup Memory Agent & 74.3\% \\
    No Combined Memory Agent & 74.1\% \\
    No Direct Visual Agent & 73.6\% \\
    Only Direct Visual Agent & 66.8\% \\
    Only Combined Memory Agent & 72.2\% \\
    Only Lookup Memory Agent & 67.8\% \\
    Full System & 77.3\% \\
    Oracle Upper Bound & 82.3\% \\
    \bottomrule
    \end{tabular}

    \vspace{8pt}
    \captionof{table}{Multimodal versus text-only memory on held-out generated episodes. 4-Tags requires all four environmental driver tags to match.}
    \label{tab:video_memory_ablation}
    \scriptsize
    \setlength{\tabcolsep}{2.5pt}
    \renewcommand{\arraystretch}{1.08}
    \resizebox{\linewidth}{!}{%
    \begin{tabular}{@{}lccccc@{}}
    \toprule
    \textbf{Method} 
    & \textbf{4-Tags} 
    & \textbf{Wind} 
    & \textbf{Moist.} 
    & \textbf{Slope} 
    & \textbf{Veg.} \\
    \midrule
    Caption video then query memory 
    & 16.2\% & 62.6\% & 51.5\% & 65.7\% & 68.7\% \\

    Video generation prompt as text memory 
    & 17.2\% & 68.7\% & 60.6\% & 66.7\% & 62.6\% \\

    Full Memory 
    & \textbf{51.5\%} 
    & \textbf{74.7\%} 
    & \textbf{83.8\%} 
    & \textbf{82.8\%} 
    & \textbf{89.9\%} \\
    \bottomrule
    \end{tabular}%
    }

\end{minipage}

\end{figure}
\vspace{-2.5mm}
\section{Ablation Study}
\vspace{-2.5mm}
\begin{wrapfigure}{r}{0.6\textwidth}
    \centering
    \vspace{-8pt}
    \includegraphics[width=0.6\textwidth]{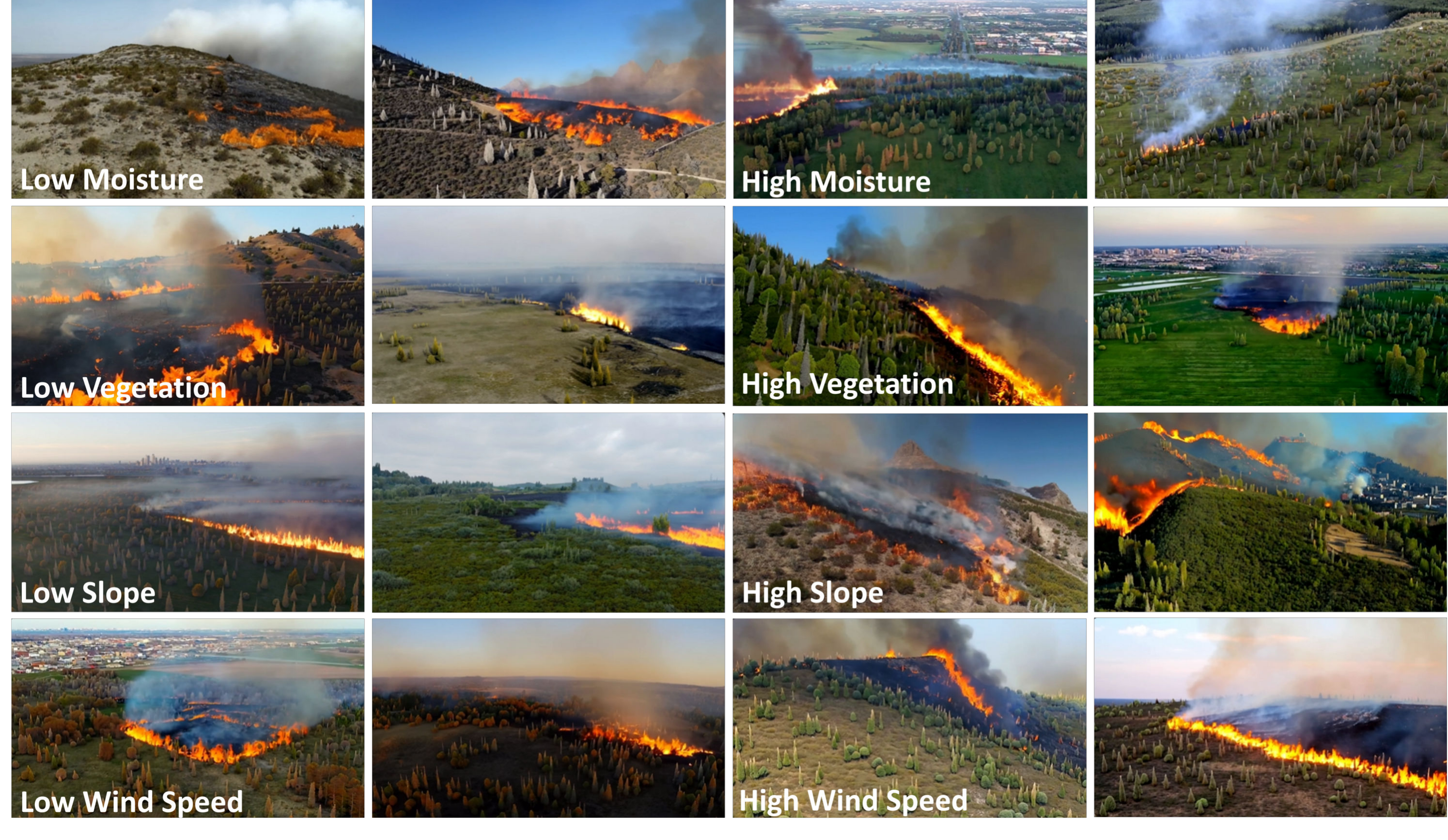}
    \vspace{-20pt}
    \caption{We show generated video frames with different simulation 
tags (2 video clips for each tag).}
    \label{fig:vis_tags}
    \vspace{-10pt}
\end{wrapfigure}
We analyze the main design choices behind the simulation-grounded VLM framework: the embedding model used for retrieval, the reranking step, the need for video memory rather than text-only memory, and the contribution of each reporting agent. Experiments in this section and Sec.~\ref{sec:validation} use 500 disjoint locations: 400 videos populate the memory bank and 100 videos are held out for evaluation. The distribution of downstream tasks and physics-labels are provided in Appendix~\ref{sec:downstream_tasks}. We also refer readers to provided supplementary materials for more visual examples. Local experiments use an NVIDIA H200 GPU; hosted VLM evaluations use model APIs.

\paragraph{Multimodal embedding model.}
\label{sec:embedding}
We compare Qwen3-2B-VL~\cite{qwen3vlembedding} with Cosmos-Embed1~\cite{nvidia2025cosmosembed1}, Rzen-v2-7B~\cite{jian2025rzenembed}, and Ops-MM-7B~\cite{opensearch_ai_ops_mm_embedding_v1_7b}. Figure~\ref{fig:embed_models} shows that Qwen3-2B-VL gives the strongest tag-retrieval accuracy in most cases, so we use it for the final memory bank.

\begin{figure*}[t]
    \centering

    \begin{minipage}{0.95\textwidth}
        \centering
        \captionof{table}{Driver-tag accuracy on held-out generated episodes. Rows require all four, at least three, or at least two correct tags. Bold and underlined values mark the best and second-best results within each group.}
        \label{tab:tag_accuracy}
        \setlength{\tabcolsep}{4pt}
        \renewcommand{\arraystretch}{1.1}
        \begin{tabular}{lcccc|cc}
            \toprule
            & \multicolumn{4}{c|}{\textbf{VLM Direct Query}} & \multicolumn{2}{c}{\textbf{Memory Retrieval}} \\
            \cmidrule(lr){2-5} \cmidrule(lr){6-7}
            \textbf{} &
            \textbf{Gemini 2.5} &
            \textbf{Claude 4} &
            \textbf{GPT-5} &
            \textbf{GPT-4.1} &
            \textbf{w/o GPT-5 Rerank} &
            \textbf{w/ GPT-5 Rerank} \\
            \midrule
            \textbf{All Tags} &
            \textbf{22.6\%} &
            \underline{22.2\%} &
            21.2\% &
            16.2\% &
            \underline{45.5\%} &
            \textbf{51.5\%} \\
            
            \textbf{Three Tags} &
            56.6\% &
            \underline{63.4\%} &
            \textbf{63.6\%} &
            49.5\% &
            \underline{78.8\%} &
            \textbf{82.8\%} \\
            
            \textbf{Two Tags} &
            \underline{90.9\%} &
            \textbf{91.4\%} &
            \underline{90.9\%} &
            85.9\% &
            \underline{94.9\%} &
            \textbf{99.0\%} \\
            \bottomrule
        \end{tabular}
    \end{minipage}

    \vspace{0.8em}

    \begin{minipage}[t]{0.49\textwidth}
        \centering
        \includegraphics[width=\linewidth]{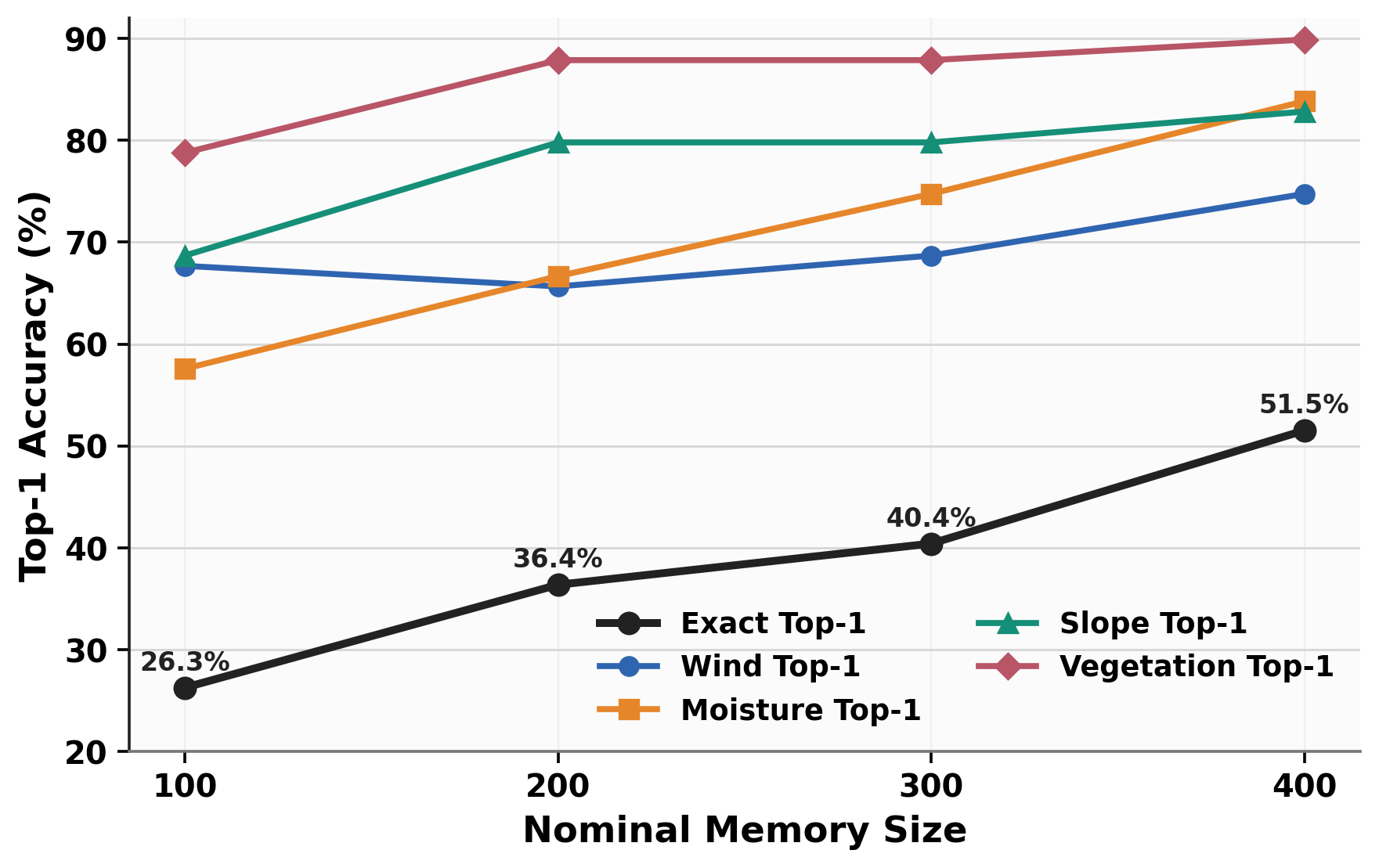}
        \captionof{figure}{Larger memory improves accuracy for individual tags and all four tags.}
        \label{fig:increase_size}
    \end{minipage}
    \hfill
    \begin{minipage}[t]{0.49\textwidth}
        \centering
        \includegraphics[width=\linewidth]{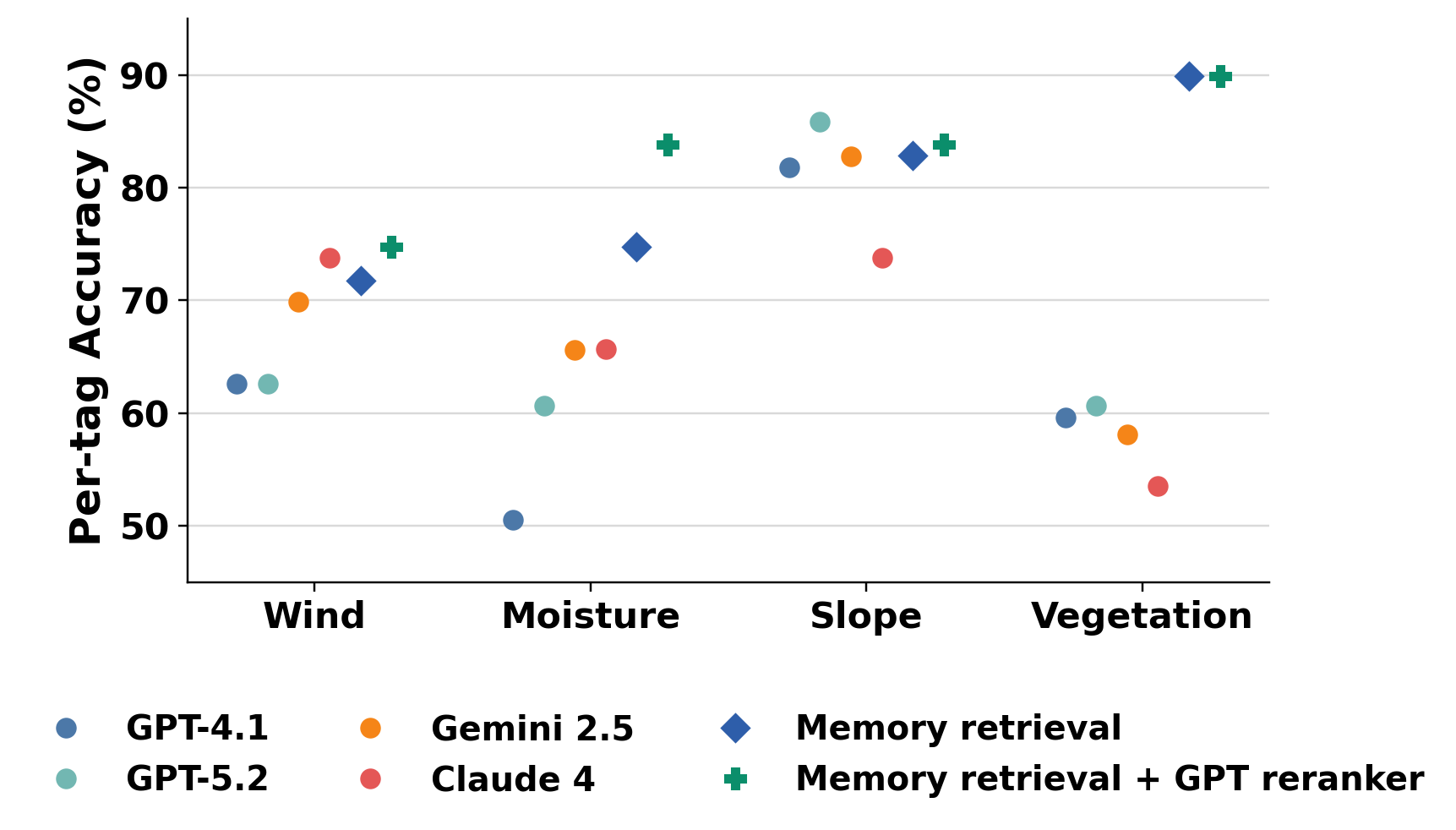}
        \captionof{figure}{Reranked memory leads in most per-tag accuracy comparisons.}
        \label{fig:per_tag_accuracy}
    \end{minipage}

\end{figure*}

\paragraph{Reranking retrieved candidates.}
GPT-5 reranking improves exact four-tag accuracy from 45.5\% to 51.5\% (Table~\ref{tab:tag_accuracy}). Increasing $k$ from 1 to 4 raises four-tag hit rate from 51.52\% to 69.70\%, measuring retrieval coverage rather than final prediction accuracy. Reranker comparisons, latency, and retrieval settings appear in Appendices~\ref{sec:impl_multiagent_reporting}, \ref{app:reranker-ablation}, and~\ref{app:retrieval-size}.
\begin{wrapfigure}{r}{0.5\textwidth}
    \centering
    \includegraphics[width=0.5\textwidth]{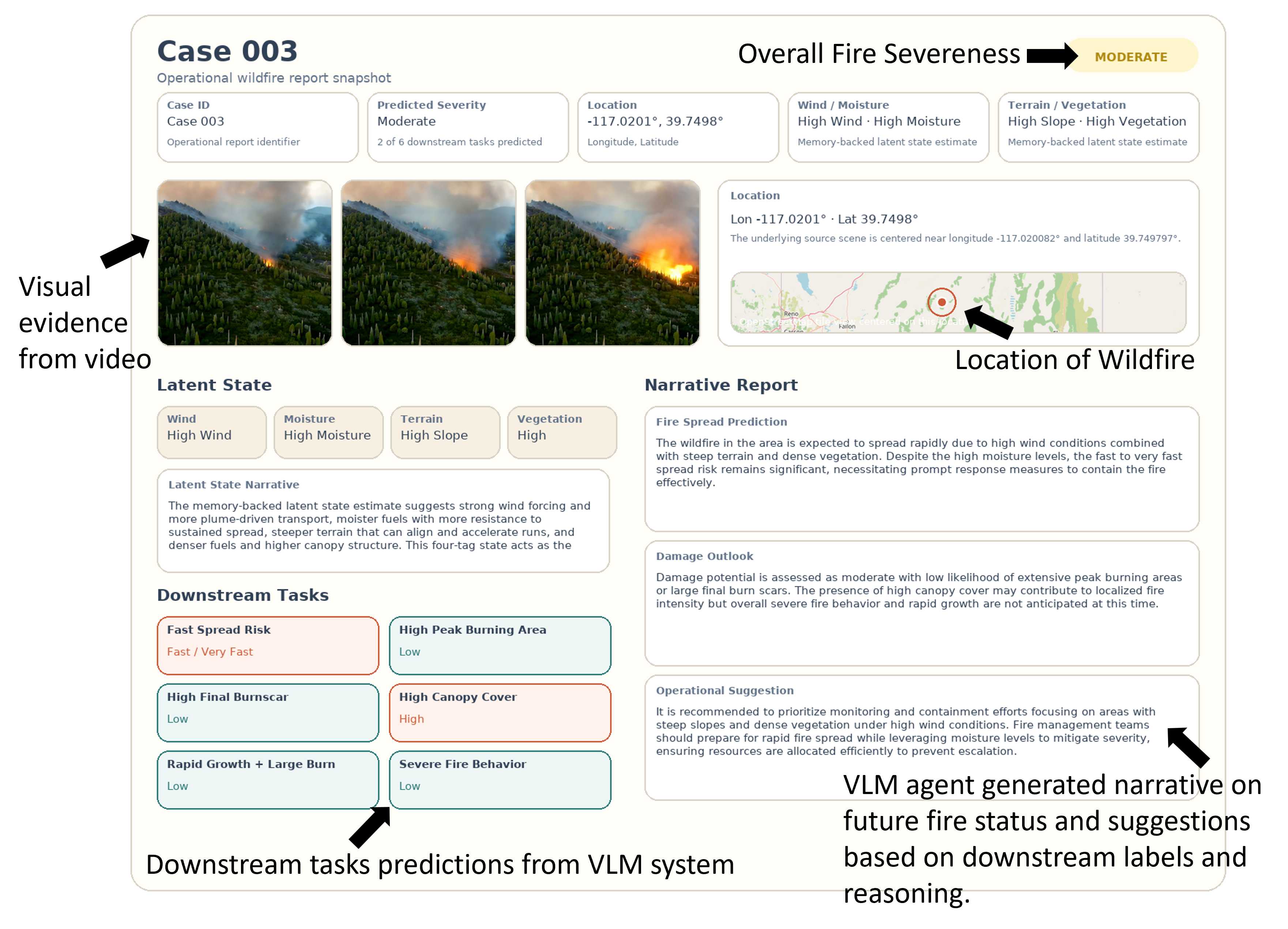}
    \caption{Illustrative reports show predictions, visual evidence, location, and response suggestions; these outputs are not operationally validated.}
    \label{fig:report}
    \vspace{-15pt}
\end{wrapfigure}
\paragraph{Importance of Video Memory.}
Table~\ref{tab:video_memory_ablation} compares full multimodal memory against two text-only variants: direct query-to-text matching and caption-then-text retrieval. Full memory improves exact four-tag accuracy from 16--17\% to 51.5\% and is stronger for every individual tag, supporting the use of stored video exemplars rather than prompt-only memory.

\paragraph{Multi-Agent Effects on Downstream Task Accuracy.}
The full system achieves 77.3\% on six simulator-derived fields, versus 72.2\% for the combined-memory reporter alone and 72.6\% for majority vote (Table~\ref{tab:method_accuracy}). The 82.3\% oracle marks fields correct when any reporter is correct. These ablations compare the specified components, not context- or compute-matched agent configurations.

\vspace{-2.5mm}
\section{Validation and Results}
\label{sec:validation}
\vspace{-2.5mm}

We assess proxy--video alignment, environmental-driver prediction, and six-field reporting on generated episodes with known simulator targets, then test generator changes and real-UAV smoke/fire detection and visible fire-area assessment.
\vspace{-10pt}
\subsection{Simulation-to-Observation Results}
\label{sec:simulation_alignment}

Figures~\ref{fig:2d_to_3d} and~\ref{fig:vis_tags} show simulation-to-video construction and tag-conditioned appearances; additional examples appear in Figure~\ref{fig:more_vlm} and the supplement. Against mismatched proxies, correct-pair SSIM and LPIPS win rates are 76.6\% and 99.1\%, with mean smoke-plume-axis and growth-front-axis errors of $7.6^\circ$ and $35.8^\circ$ (Appendix~\ref{sec:2d_3d_app}). GPT-5.6-luna rates agreement across fire-front location, fuel distribution, burned area, and terrain at 75.00/100 (Appendix~\ref{app:cue-preservation}). These results support partial visual-cue alignment, not preservation of physical spread rates or burned-area trajectories.

\vspace{-5pt}
\subsection{Memory-Based Environmental Driver Prediction}\label{sec:ablation_agent}
\vspace{-5pt}

Table~\ref{tab:tag_accuracy} compares exact four-tag accuracy and the proportions with at least three or two correct tags on held-out generated videos. Reranked memory reaches 51.5\%, versus 22.6\% for the strongest query-only VLM. With 16 labeled reference videos covering all tag combinations, the strongest example-augmented baseline achieves 27.27\%; query simulator metadata remains withheld (Appendix~\ref{app:labeled-examples}).

\subsection{Memory Growth Effects}

\begin{table*}[t]
    \centering
    \vspace{-10pt}
    \caption{Accuracy on six simulator-derived report fields for held-out generated episodes. Higher is better. Bold and underlined values mark the best and second-best results, respectively.}
    \label{tab:downstream_acc}
    \scriptsize
    \setlength{\tabcolsep}{3pt}
    \renewcommand{\arraystretch}{1.05}
    \begin{tabular}{lcccc}
        \toprule
        \textbf{Task} &
        \textbf{Direct Query GPT} &
        \textbf{Multi-Agent + Combined Memory} &
        \textbf{ForestFireVLM} &
        \textbf{Qwen2-Wildfire-2B} \\
        \midrule
        Fast Spread Risk          & \underline{55.7\%} & \textbf{75.3\%} & 58.8\% & 43.3\% \\
        High Peak Burning Area    & \underline{63.9\%} & \textbf{77.3\%} & 54.6\% & 59.8\% \\
        High Final Burnscar       & \underline{64.9\%} & \textbf{80.4\%} & 55.7\% & 60.8\% \\
        High Canopy Cover         & 57.7\% & \textbf{73.2\%} & \underline{67.0\%} & 47.4\% \\
        Rapid Growth + Large Burn & \underline{64.9\%} & \textbf{79.4\%} & 55.7\% & 60.8\% \\
        Severe Fire Behavior      & \underline{66.0\%} & \textbf{78.4\%} & 56.7\% & 59.8\% \\
        \bottomrule
    \end{tabular}
    \vspace{-10pt}
\end{table*}

Increasing the memory bank from 100 to 400 episodes improves exact
four-tag accuracy from 26.3\% to 51.5\%
(Figure~\ref{fig:increase_size}). This supports the training-free
memory design: adding simulator-grounded episodes broadens the
available reference coverage without updating VLM weights.

\begin{wraptable}{r}{0.49\textwidth}
    
    \centering
    \small
    \caption{Generalization tests. 
    (a) Cross-generator accuracy (\%) on generated episodes; LTX denotes LTX-2.3 + AVControl. 
    (b) Real-UAV smoke/fire detection and visible fire-area assessment: Boreal accuracy is a percentage; other scores 
    are on a 0--1 scale. Best baseline is selected separately for each metric, 
    not one common model. Bold includes ties. Full real-UAV results appear in 
    Appendix~\ref{app:generalization-tests}.}
    \label{tab:generalization-compact}

    \vspace{0.7em}

    \resizebox{\linewidth}{!}{%
    \begin{tabular}{@{}lrrrr@{}}
        \toprule
        \multicolumn{5}{@{}l}{\textbf{(a) Cross-generator evaluation}} \\
        Query $\rightarrow$ Memory & Ours & GPT & Claude & Gemini \\
        \midrule
        LTX $\rightarrow$ VACE 
            & \textbf{41.00} & 19.00 & 24.00 & 28.00 \\
        VACE $\rightarrow$ LTX 
            & \textbf{44.44} & 27.27 & 23.23 & 26.26 \\
        \bottomrule
    \end{tabular}%
    }

    \vspace{0.7em}

    \resizebox{\linewidth}{!}{%
    \begin{tabular}{@{}llrr@{}}
        \toprule
        \multicolumn{4}{@{}l}{\textbf{(b) External UAV evaluation}} \\
        Dataset & Metric & Ours & Best baseline \\
        \midrule
        Boreal Forest Fire 
            & Accuracy (\%) 
            & \textbf{100.0} & \textbf{100.0} \\
        FireMan-UAV-RGBT 
            & Balanced accuracy 
            & \textbf{0.9625} & 0.8625 \\
        & Accuracy 
            & \textbf{0.9423} & 0.9231 \\
        FLAME 
            & Mean AP 
            & 0.950 & \textbf{0.952} \\
        & Largest-fire Hit@1 
            & \textbf{0.850} & \textbf{0.850} \\
        & Fire-area NDCG 
            & \textbf{0.984} & 0.973 \\
        \bottomrule
    \end{tabular}%
    }
    \vspace{-30pt}
\end{wraptable}
\subsection{Downstream Fire-Behavior Prediction}

With query simulator quantities withheld, Table~\ref{tab:downstream_acc} compares six simulator-derived report fields (Appendix~\ref{sec:downstream_tasks}) against direct GPT querying, ForestFireVLM~\cite{seidel2025advancing}, and Qwen2-Wildfire-2B~\cite{du2025wildfire}. Our system leads on all six fields, including Fast Spread Risk (75.3\% versus 55.7\% for direct GPT). Predictions, visual evidence, retrieved summaries, tags, and rationales support the structured report and separate narrative (Figure~\ref{fig:report} and supplement).

\subsection{Cross-Generator and Real-UAV Evaluation}
\label{sec:generalization}

\paragraph{Cross-generator evaluation.}
To test robustness to changes in the visual synthesis backbone, we replace VACE with LTX-2.3 + AVControl while retaining the simulation and proxy construction, and test both query--memory directions. Our method achieves 41.00\% for LTX
queries against VACE memory and 44.44\% in the reverse direction,
compared with 28.00\% and 27.27\% for the strongest direct VLM
baselines (Table~\ref{tab:generalization-compact}a). The advantage
therefore persists when queries and memory use different video
generators, although accuracy is below the within-generator result.

\vspace{-10pt}
\paragraph{External UAV footage.}
We evaluate real footage using the annotations available in three public UAV datasets: smoke detection on Boreal Forest Fire~\cite{pesonen2025boreal}, binary fire presence on FireMan-UAV-RGBT~\cite{kularatne2024fireman}, and segmentation-derived visible fire-area assessment on FLAME~\cite{shamsoshoara2021aerial}. These tests complement the controlled physical-label experiments by assessing observable aspects of wildfire monitoring. Table~\ref{tab:generalization-compact}b shows the highest balanced accuracy on FireMan-UAV-RGBT and NDCG on FLAME; our method ties the best FLAME Hit@1 but not mean AP. Several methods reach 100\% on Boreal Forest Fire. The annotations used here do not jointly specify all four environmental drivers, so these results assess real-UAV perception rather than physical-driver recovery. Full results and evaluation descriptions appear in Appendix~\ref{app:generalization-tests}.

\section{Discussion}
\label{sec:discussion}
\paragraph{Relationship to Code World Model.}
Code World Model~\citep{chen2026codeworldmodel} and our framework combine explicit state evolution, spatial proxies, and generative visualization. Code World Model uses a coding agent to maintain persistent state and revise mechanisms for interaction. We instead map prescribed wildfire trajectories to aligned aerial-view episodes through a fixed simulator-to-proxy converter, using these labeled episodes as multimodal memory for interpreting query videos.

\paragraph{Fixed conversion versus scene authoring.}
GPT-6 Astra can author Blender scenes and could help develop our converter. New episodes, however, change inputs rather than code. The proxy prioritizes simulator alignment and low rendering overhead, leaving appearance detail to video generation. Reuse avoids per-episode scene-authoring calls and token costs; other inference costs remain. Like Code World Model, we separate high-level reasoning from repeated execution: the distinction concerns where reasoning is needed, not a measured speed or cost advantage.

\vspace{-1.5mm}
\section{Limitations and Future Work}
\vspace{-1.5mm}
\paragraph{Limitations.}
Real-incident environmental-driver estimation still requires validation against synchronized measurements: our real-UAV tests assess smoke/fire detection and visible fire area, whereas physical-label targets come from simulation. The fixed Blender mapping is domain-specific; new simulator fields or scene types may require extending its rules and assets. Spatial correspondence does not make billboard smoke a 3D physical solution or ensure that generated videos preserve proxy dynamics. Metadata-conditioned prompts may introduce tag-specific cues, and footage may omit sensor noise, camera motion, long-horizon evolution, or rare extremes. The coarse simulator and tags omit coupled fire--atmosphere dynamics, spotting, suppression, and changing weather. Retrieval depends on memory coverage, reporter agreement does not guarantee correctness, and reports require expert validation before operational use.

\paragraph{Future Work.}
Future work should calibrate generated observations against physically annotated real footage, expand scenario coverage, and involve wildfire experts. Further tests should isolate prompt conditioning, compare reporting with matched evidence and compute, and measure stage-wise costs against agent-authored proxy alternatives. Integration with hazard-adaptation and UAV planning frameworks~\cite{xie2025marsha,hyun2025crew,webb2026fire} requires task-specific validation.

\vspace{-2.5mm}
\section{Conclusion}
\vspace{-1.5mm}
We presented a simulation-grounded agentic VLM framework for wildfire monitoring and reporting. A fully automatic Blender conversion expresses 2D wildfire states as lightweight, simulation-aligned proxies; controlled video generation produces labeled episodes for multimodal memory. Retrieved references and complementary reporters support environmental-driver estimation and structured reporting without updating VLM weights. Controlled, cross-generator, and real-UAV perceptual tests support this use of simulation experience under scarce physical annotations. Real-world physical calibration and expert validation remain important next steps.

\bibliography{example_paper}
\bibliographystyle{iclr2027_conference}
\clearpage
\appendix
\etocdepthtag.toc{appendix}
\renewcommand{\contentsname}{Appendix Contents}
\begingroup
\etocsettagdepth{main}{none}
\etocsettagdepth{appendix}{subsection}
\small
\setlength{\parskip}{0pt}
\tableofcontents
\endgroup
\clearpage
\section{Downstream Task and Physics Labels}
\label{sec:downstream_tasks}
\paragraph{Downstream Task and Label Definitions}
We evaluate downstream wildfire understanding using six binary report fields. Each field is defined from simulator-derived quantities associated with an episode. Let $v_{\max}$ denote the maximum radial spread rate in meters per second, $A_{\mathrm{peak}}$ the maximum burning area during the episode in square meters, $A_{\mathrm{final}}$ the final burn-scar area in square meters, and $c$ the mean canopy-cover percentage. For a held-out query, these quantities define evaluation targets and are withheld from inference. Retrieved memory entries may contain the corresponding quantities for their own reference episodes. The fields describe episode-level conditions and outcomes, not validated operational forecasts. Definitions and label construction follow Section~\ref{sec:method_multiagent_reporting}; the four environmental-tag thresholds are given in Appendix~\ref{sec:implementation_details}.

\paragraph{Task and Label Distributions}
We report the class distributions of the four simulation-derived physical labels and the six downstream tasks used in our evaluation. We also include three simple reference baselines: \emph{Majority}, which always predicts the most frequent class in the evaluation set; \emph{Random}, which predicts uniformly at random; and \emph{Stratified}, which samples predictions according to the empirical class prior.

\begingroup
\rowcolors{2}{TableStripe}{white}
\small
\begin{longtable}{@{}L{0.18\linewidth} L{0.37\linewidth} C{0.11\linewidth} C{0.11\linewidth} C{0.11\linewidth}@{}}
\rowcolor{TableHead} \textbf{Task} & \textbf{Class balance} & \textbf{Majority} & \textbf{Random} & \textbf{Stratified} \\
\midrule
\endfirsthead
\rowcolor{TableHead}\textbf{Task} & \textbf{Class balance} & \textbf{Majority} & \textbf{Random} & \textbf{Stratified} \\
\midrule
\endhead
Wind & high: 49.5\%, low: 50.5\% & 50.5\% & 50.0\% & 50.0\% \\
Moisture & high: 49.5\%, low: 50.5\% & 50.5\% & 50.0\% & 50.0\% \\
Slope & high: 44.4\%, low: 55.6\%  & 55.6\% & 50.0\% & 50.6\% \\
Vegetation & high: 47.5\%, low: 52.5\% & 52.5\% & 50.0\% & 50.1\% \\
\bottomrule
\end{longtable}
\endgroup

\begingroup
\rowcolors{2}{TableStripe}{white}
\small
\begin{longtable}{@{}L{0.18\linewidth} L{0.37\linewidth} C{0.11\linewidth} C{0.11\linewidth} C{0.11\linewidth}@{}}
\rowcolor{TableHead} \textbf{Task} & \textbf{Class balance} & \textbf{Majority} & \textbf{Random} & \textbf{Stratified} \\
\midrule
\endfirsthead
\rowcolor{TableHead}\textbf{Task} & \textbf{Class balance} & \textbf{Majority} & \textbf{Random} & \textbf{Stratified} \\
\midrule
\endhead
Fast Spread Risk & fast: 56.7\%, slow/moderate: 43.3\% & 56.7\% & 50.0\% & 50.9\% \\
High Peak Burning Area & high: 40.2\%, low: 59.8\%  & 59.8\% & 50.0\% & 51.9\% \\
High Final Burnscar & high: 39.2\%, low: 60.8\% &  60.8\% & 50.0\% & 52.3\% \\
High Canopy Cover & high: 44.3\%, low: 55.7\% & 55.7\% & 50.0\% & 50.6\% \\
Rapid Growth + Large Burn & high: 39.2\%, low: 60.8\%  & 60.8\% & 50.0\% & 52.3\% \\
Severe Fire Behavior & high: 40.2\%, low: 59.8\% & 59.8\% & 50.0\% & 51.9\% \\
\bottomrule
\end{longtable}
\endgroup

\begin{table*}[t]
\centering
\small
\caption{Binary simulator-derived report-field definitions. Here $v_{\max}$ is the maximum radial spread rate, $A_{\mathrm{peak}}$ is the peak burning area, $A_{\mathrm{final}}$ is the final burn-scar area, and $c$ is the mean canopy-cover percentage.}
\label{tab:downstream_tasks}
\vspace{0.25em}
\rowcolors{2}{TableStripe}{white}
\begin{tabularx}{\textwidth}{
  >{\raggedright\arraybackslash}p{0.20\textwidth}
  >{\raggedright\arraybackslash}X
  >{\raggedright\arraybackslash}X
}
\rowcolor{TableHead}
\textbf{Task} &
\textbf{Meaning} &
\textbf{Positive Label Definition} \\
\midrule

\textbf{Fast Spread Risk}
& The fire spreads at a fast or very fast rate.
& Positive if $v_{\max} \geq 0.30$ m/s. We use four spread classes: slow for $v_{\max}<0.15$, moderate for $0.15 \leq v_{\max}<0.30$, fast for $0.30 \leq v_{\max}<0.45$, and very fast for $v_{\max}\geq0.45$. \\

\textbf{High Peak Burning Area}
& The episode reaches a large active burning area.
& Positive if $A_{\mathrm{peak}} \geq 90{,}000$ m$^2$. \\

\textbf{High Final Burn Scar}
& The episode leaves a large final burned area.
& Positive if $A_{\mathrm{final}} \geq 300{,}000$ m$^2$. \\

\textbf{High Canopy Cover}
& The episode occurs in a relatively high-canopy environment.
& Positive if $c$ is at or above the median canopy cover in the training set. In our data, this threshold is $2.61\%$. \\

\textbf{Rapid Growth + Large Burn}
& The episode combines rapid fire spread with a large final burn scar.
& Positive if $v_{\max} \geq 0.30$ m/s and $A_{\mathrm{final}} \geq 300{,}000$ m$^2$. \\

\textbf{Severe Fire Behavior}
& The episode satisfies multiple indicators of severe fire behavior.
& Positive if at least two of the following three conditions hold: $v_{\max} \geq 0.30$ m/s, $A_{\mathrm{peak}} \geq 90{,}000$ m$^2$, and $A_{\mathrm{final}} \geq 300{,}000$ m$^2$. \\

\bottomrule
\end{tabularx}
\end{table*}

\section{Wildfire Models}
\label{sec:firemodel_app}
For completeness, in this section, we list the fire simulation model we used for generating 2D fire simulations together with the definition of the Anderson 13 Fuel Profile with corresponding billboard assets.
\subsection{Rothermel Surface Fire Spread Model}
\label{sec:rothermel_app}
For completeness, we provide the Rothermel surface fire spread model used for 2D simulation which is the backend model that SimFire uses. It is a semi-empirical model that predicts the forward rate of spread of a surface fire from local fuel, moisture, wind, and terrain conditions~\cite{rothermel1972mathematical}. In our simulations, this model provides a directional rate of spread from an actively burning grid cell to each candidate neighboring cell. All quantities follow the SimFire implementation.

Let cell $i$ be an actively burning cell and cell $j$ be a candidate neighboring cell. The directional rate of spread from $i$ to $j$ is
\begin{equation}
R_{i\rightarrow j}
=
\max\left(
0,\,
\frac{I_R \xi \left(1+\phi_w+\phi_s\right)}
{\rho_b \epsilon Q_{\mathrm{ig}}}
\right),
\label{eq:rothermel_ros}
\end{equation}
where $R_{i\rightarrow j}$ is the directional rate of spread from cell $i$ to cell $j$, $I_R$ is the reaction intensity, $\xi$ is the propagating flux ratio, $\phi_w$ and $\phi_s$ are the wind and slope correction factors, $\rho_b$ is the oven-dry bulk density, $\epsilon$ is the effective heating number, and $Q_{\mathrm{ig}}$ is the heat of preignition.

For each target cell, It first computes the fuel packing and damping terms. Let $w_0$ denote oven-dry fuel load, $\delta$ fuel-bed depth, $M_x$ moisture of extinction, $\sigma$ surface-area-to-volume ratio, $h$ low heat content, $S_T$ total mineral content, $S_e$ effective mineral content, $\rho_p$ oven-dry particle density, and $M_f$ environmental fuel moisture. The net fuel load, bulk density, packing ratio, and optimal packing ratio are
\begin{equation}
w_n = w_0(1-S_T), \qquad
\rho_b = \frac{w_0}{\delta}, \qquad
\beta = \frac{\rho_b}{\rho_p}, \qquad
\beta_{\mathrm{op}} = 3.348\sigma^{-0.8189}.
\end{equation}
Here, $w_n$ is the net fuel load after discounting the total mineral content $S_T$, $\rho_b$ is the oven-dry bulk density of the fuel bed, $\beta$ is the fuel packing ratio, and $\beta_{\mathrm{op}}$ is the optimum packing ratio determined by the surface-area-to-volume ratio $\sigma$.

The moisture and mineral damping coefficients are
\begin{equation}
r_M = \min\left(\frac{M_f}{M_x},1\right), \qquad
\eta_M = 1 - 2.59r_M + 5.11r_M^2 - 3.52r_M^3,
\end{equation}
\begin{equation}
\eta_S = \min\left(0.174S_e^{-0.19},1\right).
\end{equation}
Here, $r_M$ is the moisture ratio between the environmental fuel moisture $M_f$ and the moisture of extinction $M_x$, $\eta_M$ is the moisture damping coefficient, and $\eta_S$ is the mineral damping coefficient determined by the effective mineral content $S_e$.

The maximum reaction velocity and reaction velocity are
\begin{equation}
\Gamma'_{\max}
=
\frac{\sigma^{1.5}}
{495 + 0.0594\sigma^{1.5}},
\qquad
A = 133\sigma^{-0.7913},
\end{equation}
\begin{equation}
\Gamma'
=
\Gamma'_{\max}
\left(\frac{\beta}{\beta_{\mathrm{op}}}\right)^A
\exp\left[
A\left(1-\frac{\beta}{\beta_{\mathrm{op}}}\right)
\right].
\end{equation}
Here, $\Gamma'_{\max}$ is the maximum reaction velocity, $A$ is an empirical exponent controlled by the surface-area-to-volume ratio $\sigma$, and $\Gamma'$ is the reaction velocity after accounting for the ratio between the actual packing ratio $\beta$ and the optimum packing ratio $\beta_{\mathrm{op}}$.

The reaction intensity is then
\begin{equation}
I_R = \Gamma' w_n h \eta_M \eta_S.
\end{equation}
Here, the reaction intensity $I_R$ combines the reaction velocity $\Gamma'$, net fuel load $w_n$, low heat content $h$, moisture damping coefficient $\eta_M$, and mineral damping coefficient $\eta_S$.

The propagating flux ratio is
\begin{equation}
\xi =
\frac{
\exp\left[(0.792+0.681\sigma^{1/2})(\beta+0.1)\right]
}
{192+0.2595\sigma}.
\end{equation}
Here, $\xi$ is the propagating flux ratio, which depends on the surface-area-to-volume ratio $\sigma$ and the packing ratio $\beta$.

Wind is incorporated by projecting the wind velocity onto the direction from the burning cell to the candidate cell. Let $\theta_{ij}$ be the travel direction from cell $i$ to cell $j$, and let $\theta_U$ be the wind direction in mathematical coordinates. SimFire converts wind direction from degrees clockwise from north and uses
\begin{equation}
U_{\parallel}
=
\max\left(0,\, U\cos(\theta_U-\theta_{ij})\right),
\end{equation}
where $U$ is the wind speed and $U_{\parallel}$ is its nonnegative component projected onto the candidate spread direction. Thus, only the wind component aligned with the candidate spread direction increases the spread rate. The wind factor is
\begin{equation}
\phi_w
=
c U_{\parallel}^{b}
\left(\frac{\beta}{\beta_{\mathrm{op}}}\right)^{-e},
\end{equation}
where $\phi_w$ is the wind correction factor, and $c$, $b$, and $e$ are empirical coefficients determined by the fuel surface-area-to-volume ratio and packing properties, with
\begin{equation}
c = 7.47\exp(-0.133\sigma^{0.55}), \qquad
b = 0.02526\sigma^{0.54}, \qquad
e = 0.715\exp(-3.59\times 10^{-4}\sigma).
\end{equation}
Here, $c$, $b$, and $e$ are functions of $\sigma$, and they control how strongly the projected wind speed $U_{\parallel}$ modifies the spread rate.

Terrain slope is also made directional by projecting the local slope onto the travel direction. Let $s_{\mathrm{mag}}$ be the slope magnitude and $\theta_s$ the direction of steepest ascent. SimFire computes
\begin{equation}
s_{\parallel}
=
-s_{\mathrm{mag}}\cos(\theta_s+\theta_{ij}),
\end{equation}
where $s_{\parallel}$ is the signed slope component along the candidate spread direction. It then uses a signed slope correction
\begin{equation}
\phi_s
=
5.275\beta^{-0.3}\operatorname{sgn}(s_{\parallel})s_{\parallel}^2,
\end{equation}
where $\phi_s$ is the slope correction factor, $\operatorname{sgn}(s_{\parallel})=1$ for uphill spread and $-1$ otherwise. This makes uphill spread faster and downhill spread slower in the directional grid update.

Finally, the effective heating number and heat of preignition are
\begin{equation}
\epsilon = \exp\left(-\frac{138}{\sigma}\right),
\qquad
Q_{\mathrm{ig}} = 250 + 1116M_f.
\end{equation}
Here, $\epsilon$ is the effective heating number, which depends on the surface-area-to-volume ratio $\sigma$, and $Q_{\mathrm{ig}}$ is the heat of preignition, which increases with environmental fuel moisture $M_f$. Substituting these terms into Eq.~\eqref{eq:rothermel_ros} gives the local rate of spread $R_{i\rightarrow j}$.

In SimFire, this continuous rate of spread is applied on a discrete grid. For each active burning cell, the simulator enumerates neighboring candidate cells, using eight-connected neighborhoods. For each candidate cell, SimFire evaluates Eq.~\eqref{eq:rothermel_ros}, multiplies the resulting rate by the simulation time step $\Delta t$, and accumulates a burn distance:
\begin{equation}
B_j^{t+\Delta t}
=
B_j^t + R_{i\rightarrow j}\Delta t.
\end{equation}
Here, $B_j^t$ is the accumulated burn distance at candidate cell $j$ at time $t$, and $\Delta t$ is the simulation time step. A candidate cell ignites once its accumulated burn amount exceeds the physical cell width $\Delta x$:
\begin{equation}
B_j^{t+\Delta t} > \Delta x.
\end{equation}
Here, $\Delta x$ is the physical width of one grid cell. Cells with zero fuel load are assigned zero spread rate, and negative rates are clipped to zero. It can additionally attenuate spread through mitigation cells such as firelines, scratchlines, and wetlines by reducing or zeroing the computed rate of spread before the burn amount is accumulated.

\subsection{Anderson 13 Fuel Profiles}
\label{sec:anderson_app}
In this section, we show the the billboards we used for different Anderson 13 fuel code. We also summarizes the runtime assets based on different fuel families. 
\begingroup
\rowcolors{2}{TableStripe}{white}
\small
\begin{longtable}{C{0.08\linewidth} L{0.25\linewidth} L{0.58\linewidth}}

\rowcolor{TableHead}\textbf{Code} & \textbf{Name} & \textbf{Base billboard mix in renderer} \\
\midrule
\endfirsthead
\rowcolor{TableHead}\textbf{Code} & \textbf{Name} & \textbf{Base billboard mix in renderer} \\
\midrule
\endhead
1 & Short grass & \filepath{grass_short} 94\%, \filepath{shrub_green} 6\% \\
2 & Timber (grass + understory) & \filepath{grass_short} 62\%, \filepath{shrub_green} 26\%, \filepath{tree_conifer} 12\% \\
3 & Tall grass & \filepath{grass_tall} 94\%, \filepath{tree_hardwood} 6\% \\
4 & Chaparral & \filepath{shrub_green} 65\%, \filepath{shrub_dry} 20\%, \filepath{grass_tall} 15\% \\
5 & Brush & \filepath{shrub_green} 42\%, \filepath{shrub_dry} 33\%, \filepath{grass_short} 25\% \\
6 & Dormant brush / hardwood slash & \filepath{shrub_dry} 55\%, \filepath{grass_short} 20\%, \filepath{slash} 25\% \\
7 & Southern rough & \filepath{shrub_dry} 40\%, \filepath{shrub_green} 28\%, \filepath{grass_tall} 20\%, \filepath{tree_hardwood} 12\% \\
8 & Closed timber litter & \filepath{tree_conifer} 58\%, \filepath{shrub_green} 22\%, \filepath{grass_short} 20\% \\
9 & Hardwood litter & \filepath{tree_hardwood} 58\%, \filepath{shrub_green} 22\%, \filepath{grass_tall} 20\% \\
10 & Timber (litter + understory) & \filepath{tree_conifer} 32\%, \filepath{tree_hardwood} 12\%, \filepath{grass_short} 40\%, \filepath{grass_tall} 16\% \\
11 & Light logging slash & \filepath{slash} 38\%, \filepath{tree_conifer} 17\%, \filepath{grass_short} 32\%, \filepath{shrub_dry} 13\% \\
12 & Medium logging slash & \filepath{slash} 42\%, \filepath{shrub_dry} 35\%, \filepath{grass_short} 23\% \\
13 & Heavy logging slash & \filepath{slash} 48\%, \filepath{shrub_dry} 26\%, \filepath{grass_tall} 26\% \\
\bottomrule
\end{longtable}
\endgroup

\begingroup
\rowcolors{2}{TableStripe}{white}
\small
\begin{longtable}{@{}L{0.17\linewidth} L{0.36\linewidth} L{0.16\linewidth} L{0.22\linewidth}@{}}
\rowcolor{TableHead}\textbf{Family / effect} & \textbf{Preview variants} & \textbf{Anderson 13 codes} & \textbf{Notes} \\
\midrule
\endfirsthead
\rowcolor{TableHead}\textbf{Family / effect} & \textbf{Preview variants} & \textbf{Anderson 13 codes} & \textbf{Notes} \\
\midrule
\endhead
Short grass &
\begin{minipage}[t]{0.62in}\centering\includegraphics[width=0.50in,height=0.50in,keepaspectratio]{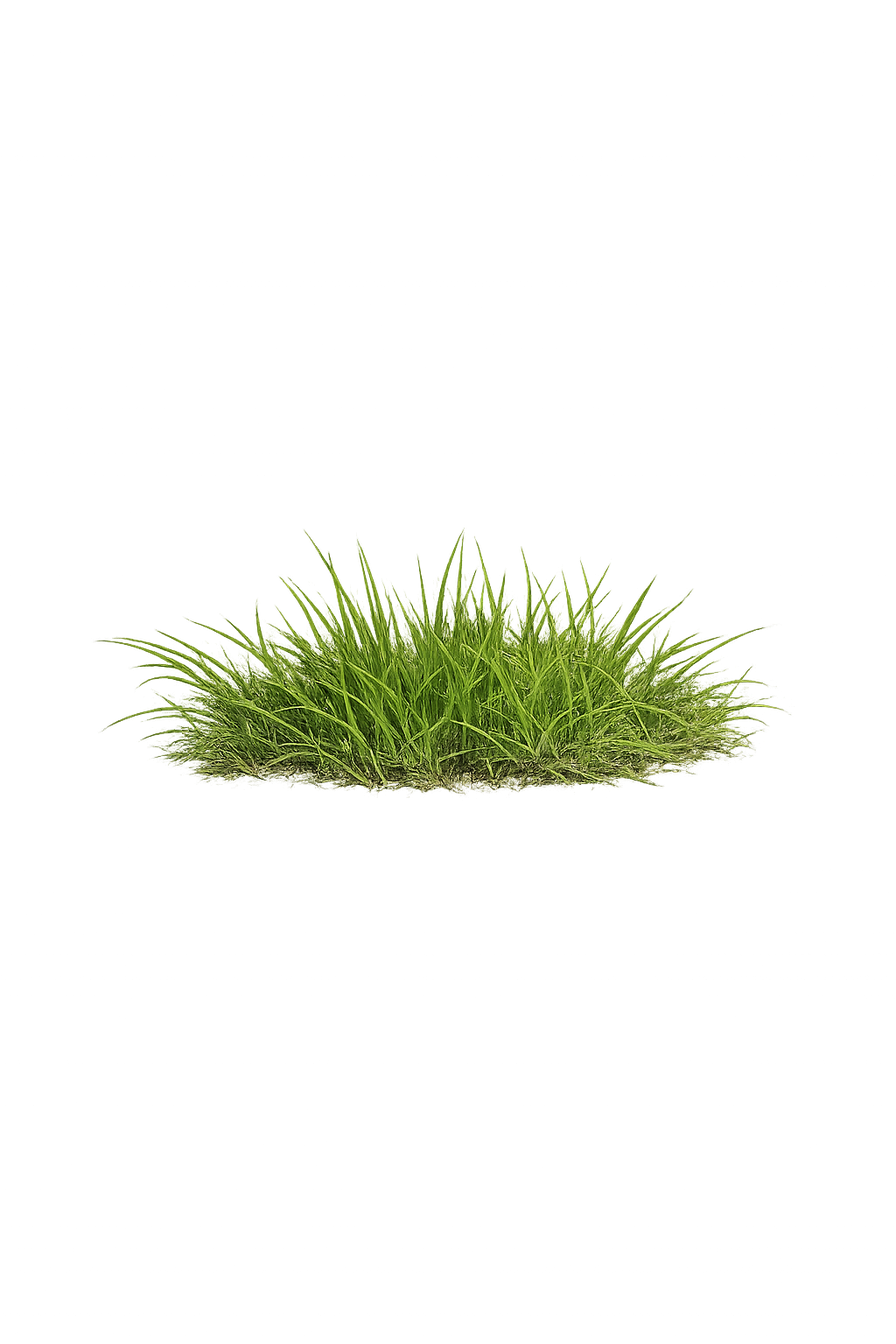}\\[-0.15em]{\tiny\sffamily base}\end{minipage}\hfill
\begin{minipage}[t]{0.62in}\centering\includegraphics[width=0.50in,height=0.50in,keepaspectratio]{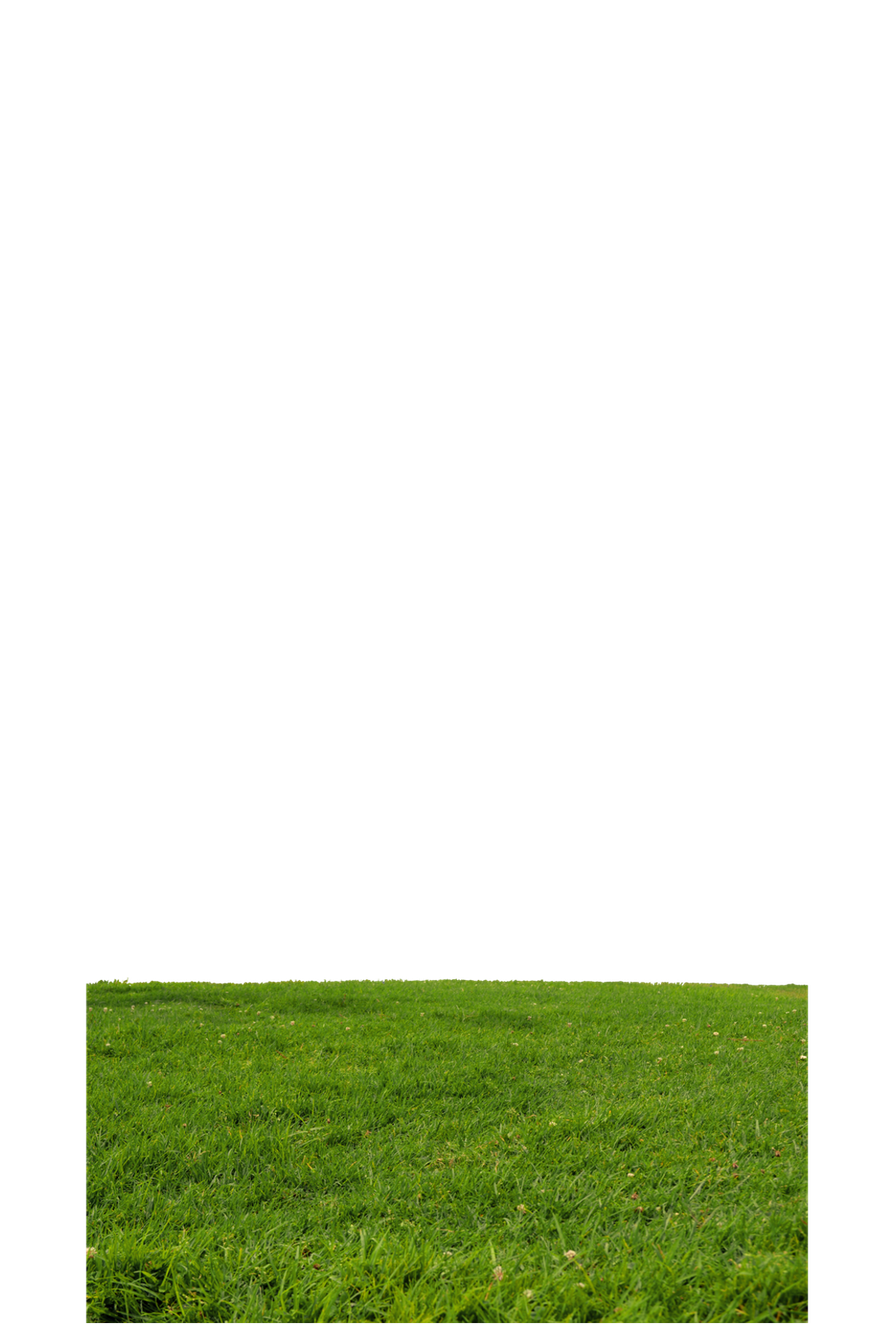}\\[-0.15em]{\tiny\sffamily dry edge}\end{minipage}\hfill
\begin{minipage}[t]{0.62in}\centering\includegraphics[width=0.50in,height=0.50in,keepaspectratio]{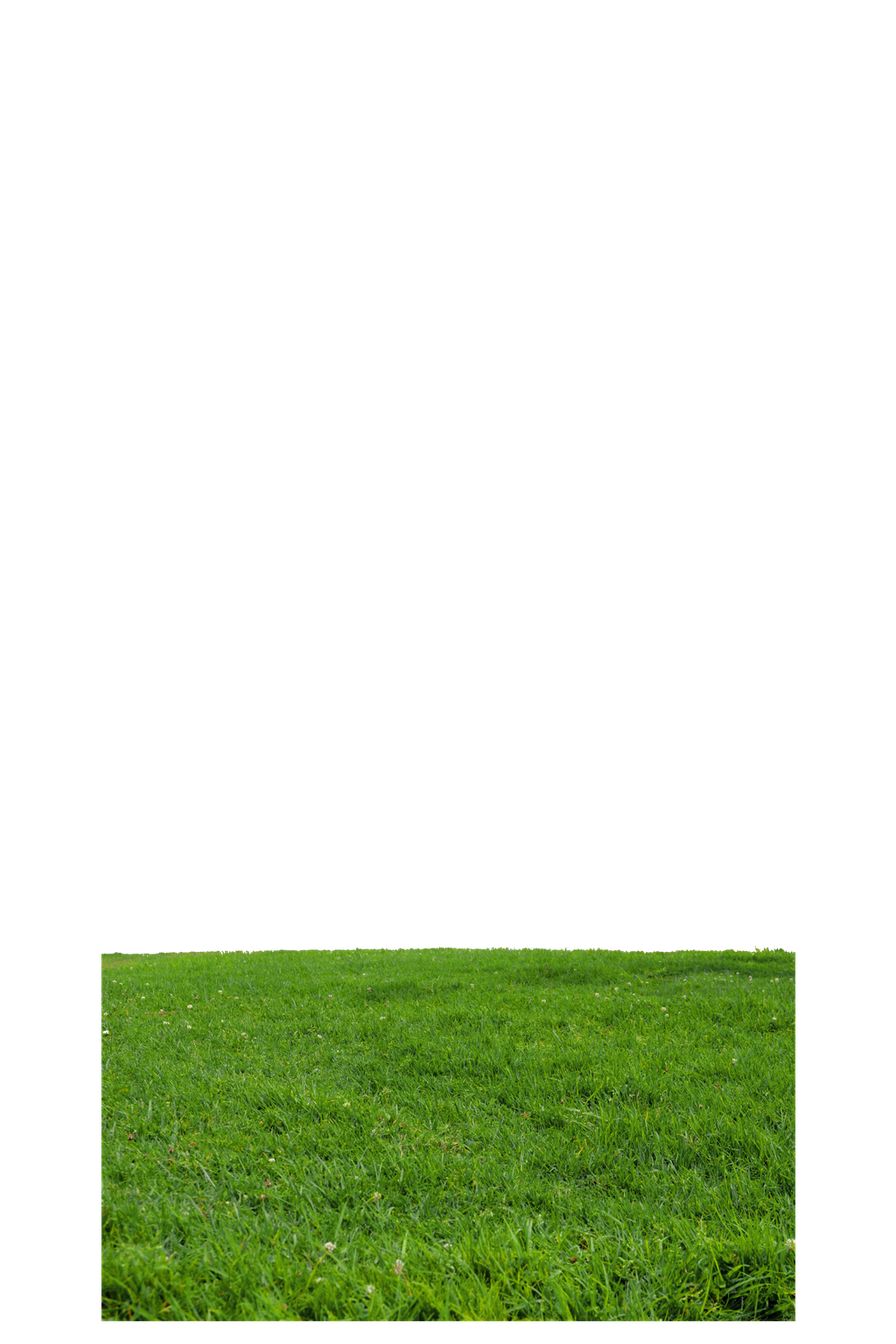}\\[-0.15em]{\tiny\sffamily meadow}\end{minipage}
& \codes{1, 2, 5, 6, 8, 10, 11, 12} & Expanded from the short-grass foliage archetype. \\

Tall grass &
\begin{minipage}[t]{0.62in}\centering\includegraphics[width=0.50in,height=0.50in,keepaspectratio]{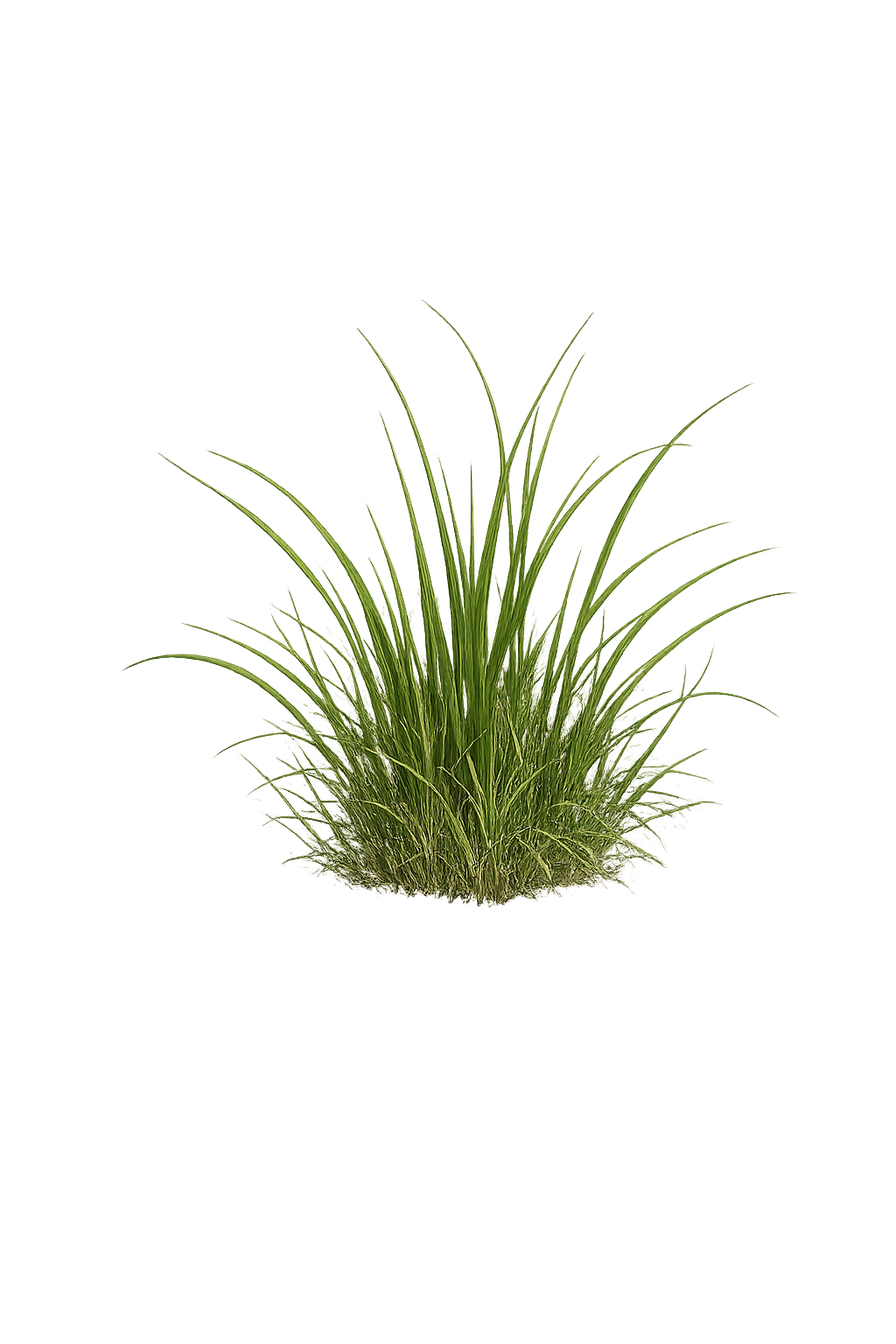}\\[-0.15em]{\tiny\sffamily base}\end{minipage}\hfill
\begin{minipage}[t]{0.62in}\centering\includegraphics[width=0.50in,height=0.50in,keepaspectratio]{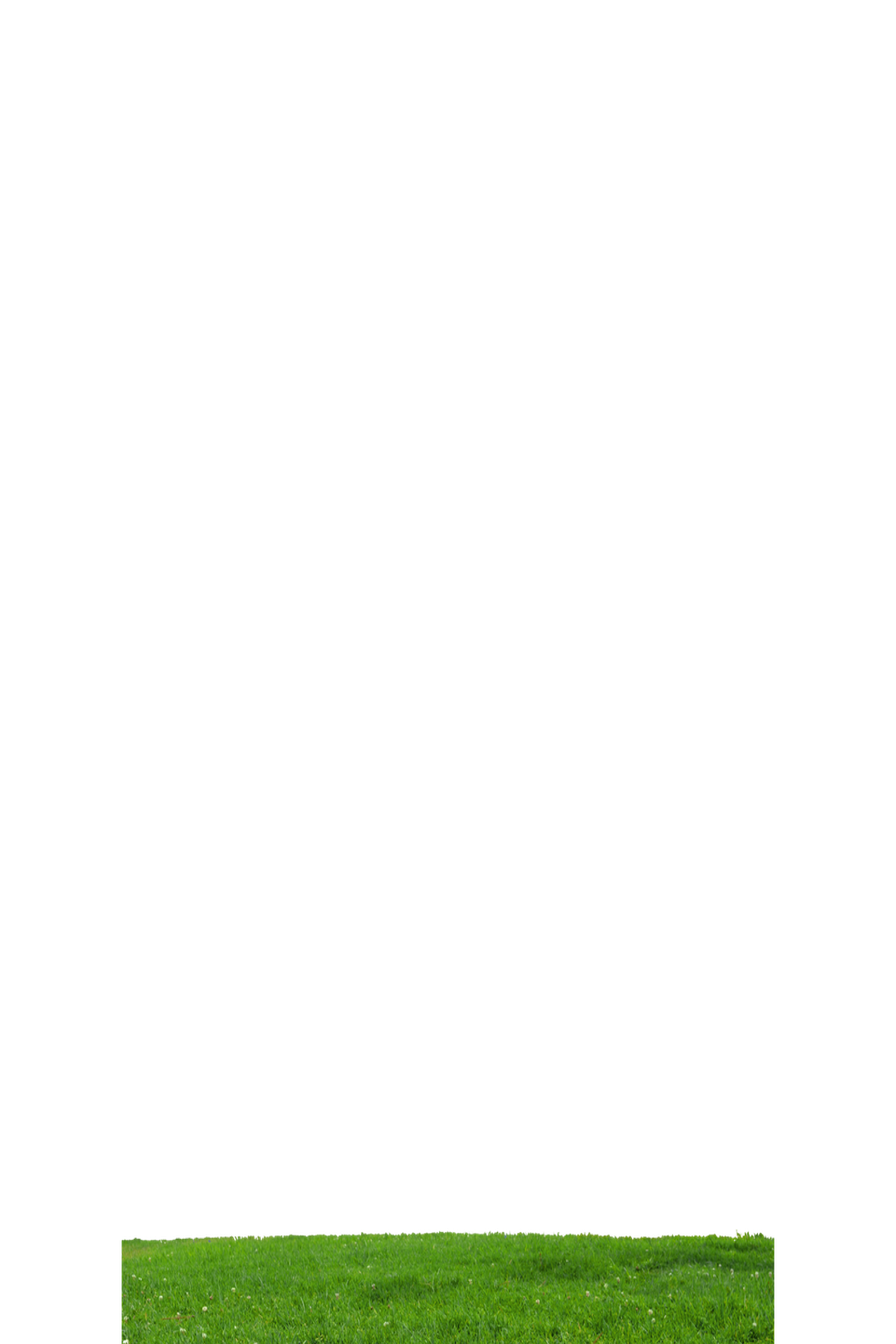}\\[-0.15em]{\tiny\sffamily reedy}\end{minipage}\hfill
\begin{minipage}[t]{0.62in}\centering\includegraphics[width=0.50in,height=0.50in,keepaspectratio]{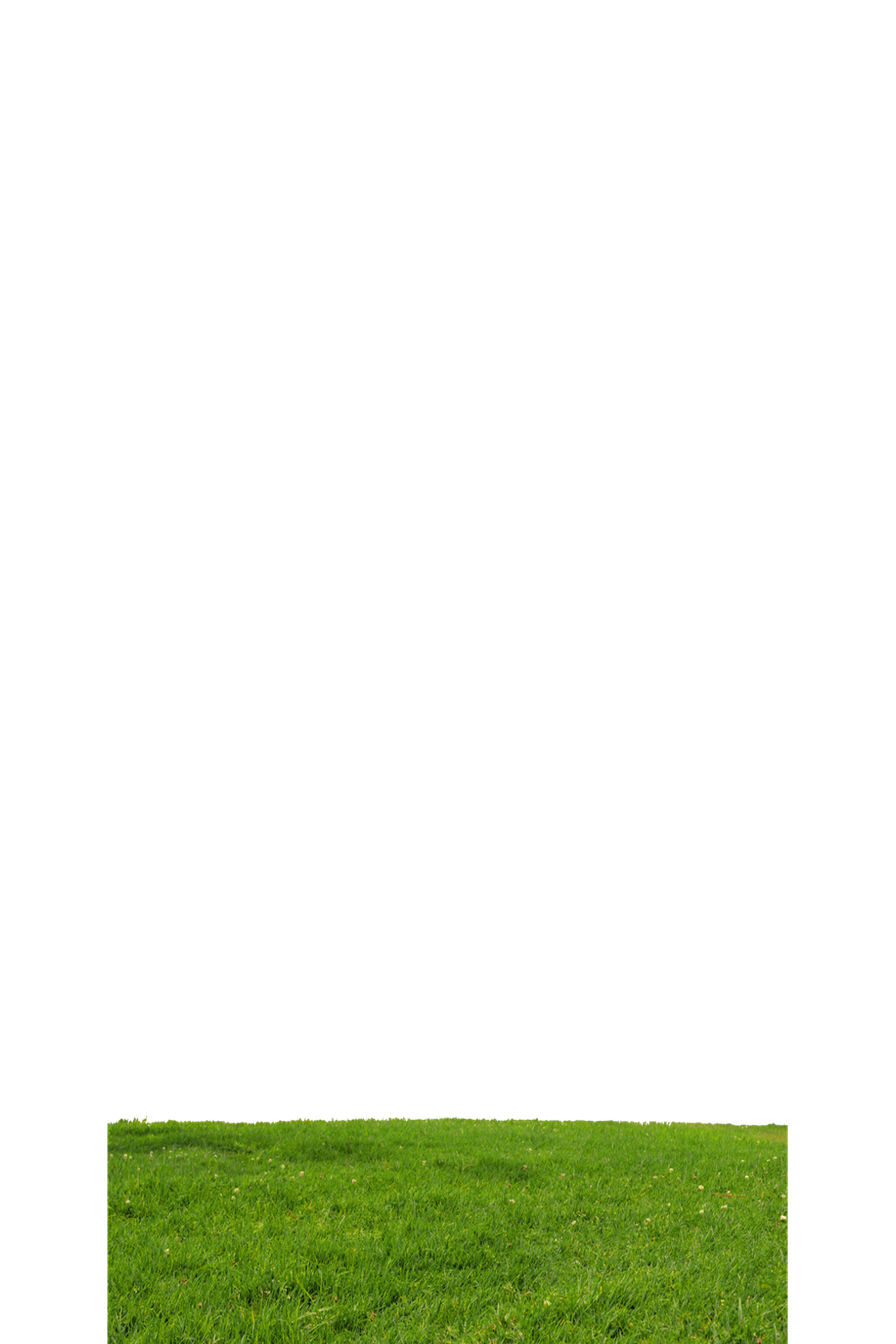}\\[-0.15em]{\tiny\sffamily sunbleached}\end{minipage}
& \codes{3, 4, 7, 9, 10, 13} & Expanded from the tall-grass foliage archetype. \\

Green shrub &
\begin{minipage}[t]{0.62in}\centering\includegraphics[width=0.50in,height=0.50in,keepaspectratio]{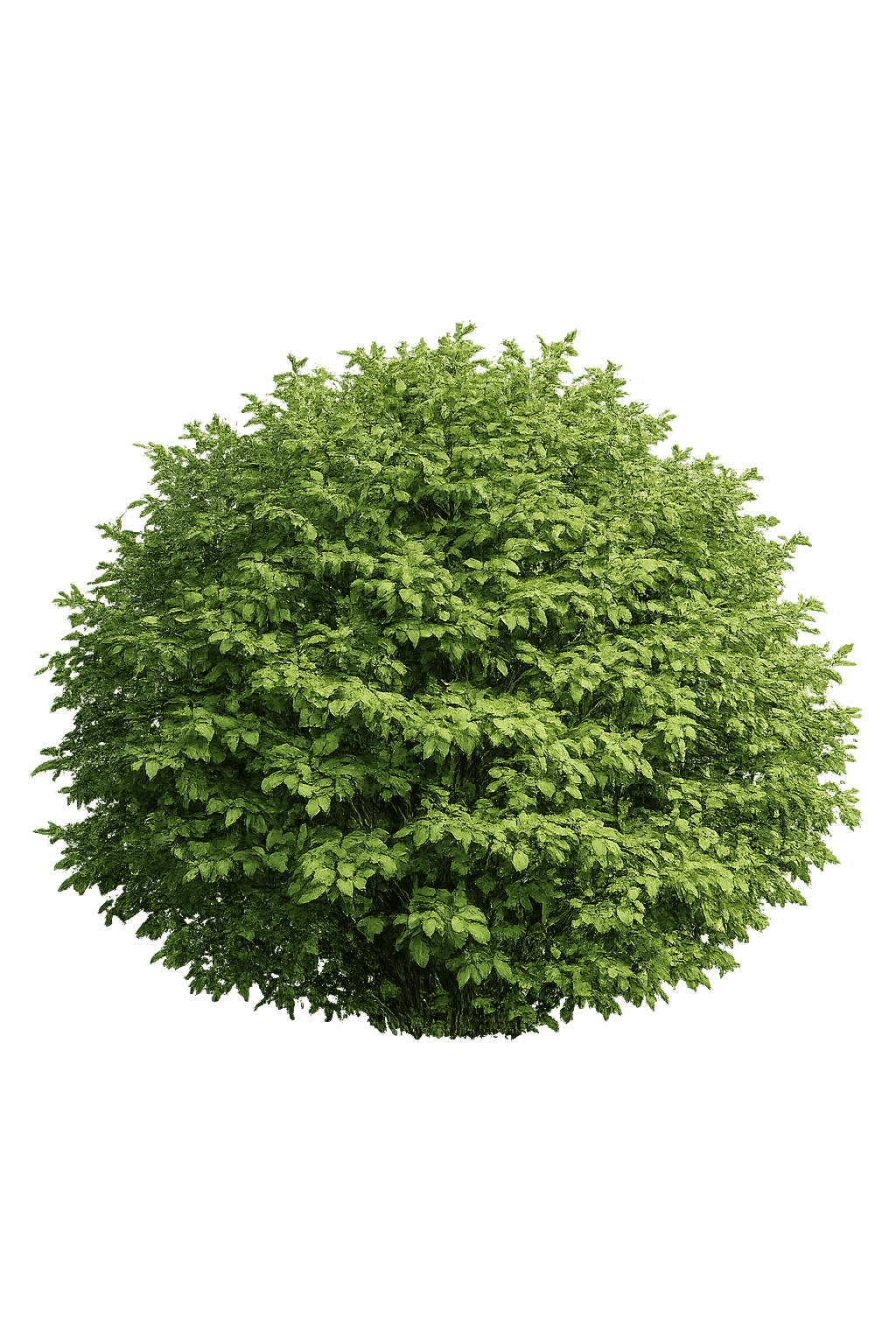}\\[-0.15em]{\tiny\sffamily base}\end{minipage}\hfill
\begin{minipage}[t]{0.62in}\centering\includegraphics[width=0.50in,height=0.50in,keepaspectratio]{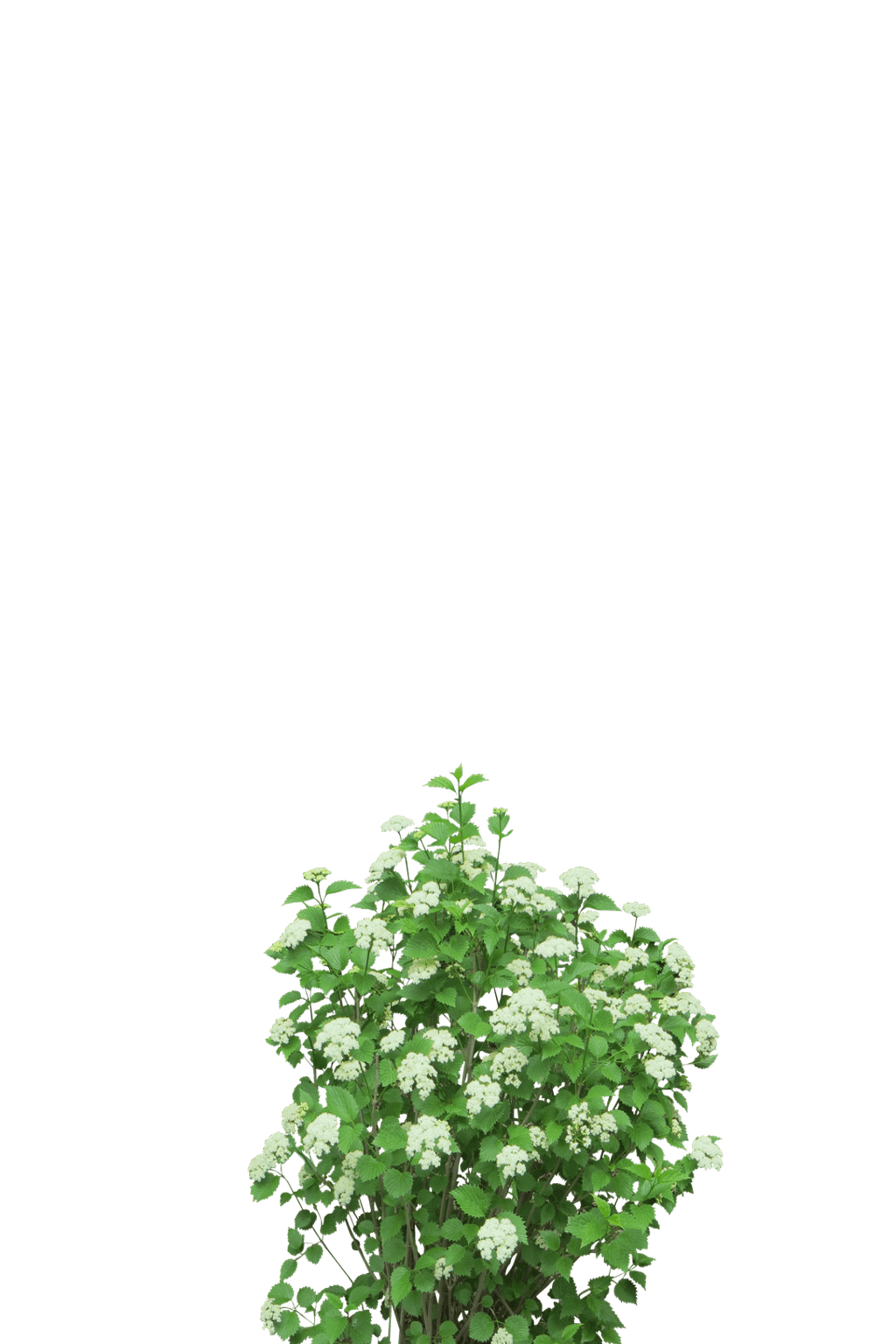}\\[-0.15em]{\tiny\sffamily dense}\end{minipage}\hfill
\begin{minipage}[t]{0.62in}\centering\includegraphics[width=0.50in,height=0.50in,keepaspectratio]{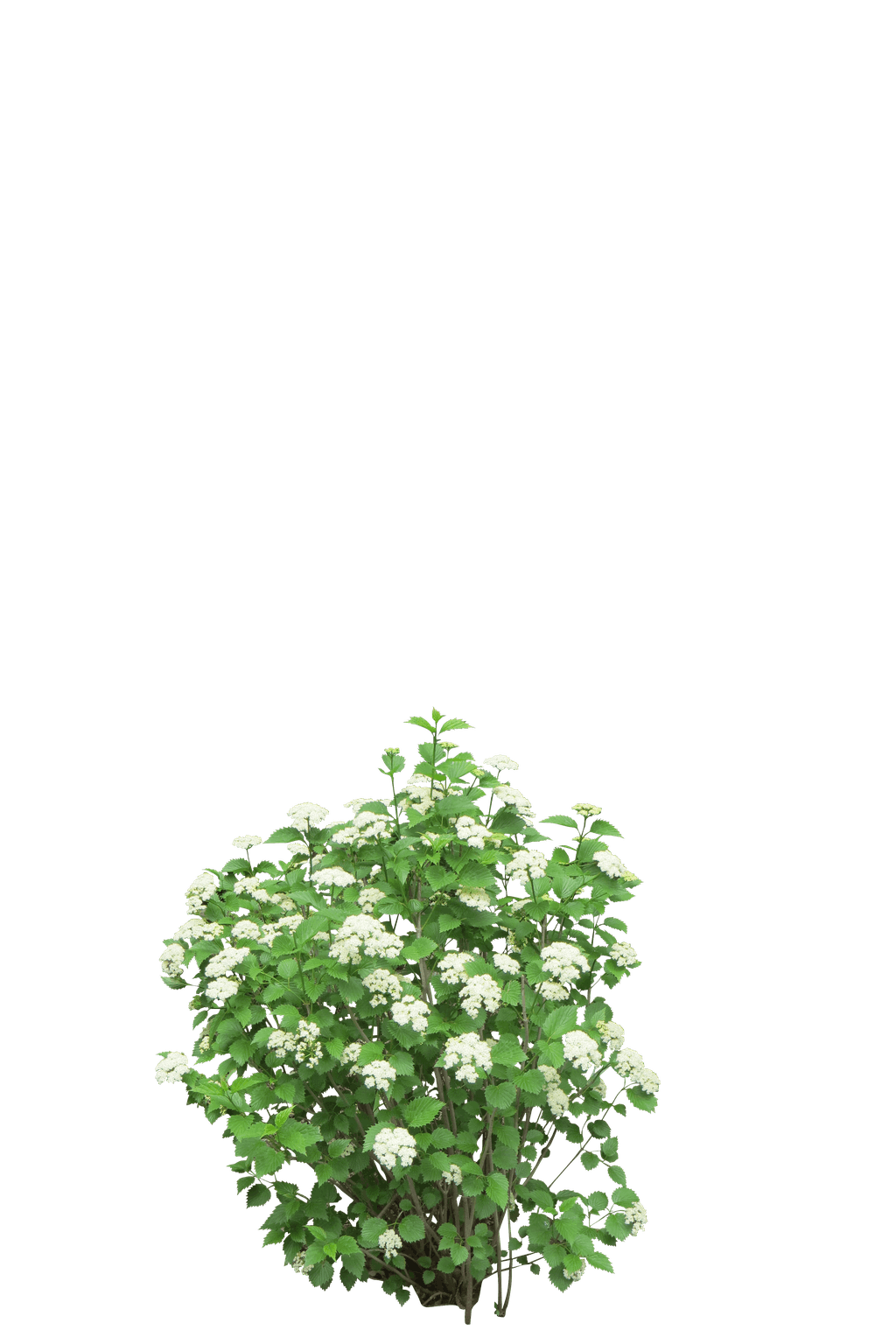}\\[-0.15em]{\tiny\sffamily wide}\end{minipage}
& \codes{1, 2, 4, 5, 7, 8, 9} & Expanded from the green-shrub foliage archetype. \\

Dry shrub &
\begin{minipage}[t]{0.62in}\centering\includegraphics[width=0.50in,height=0.50in,keepaspectratio]{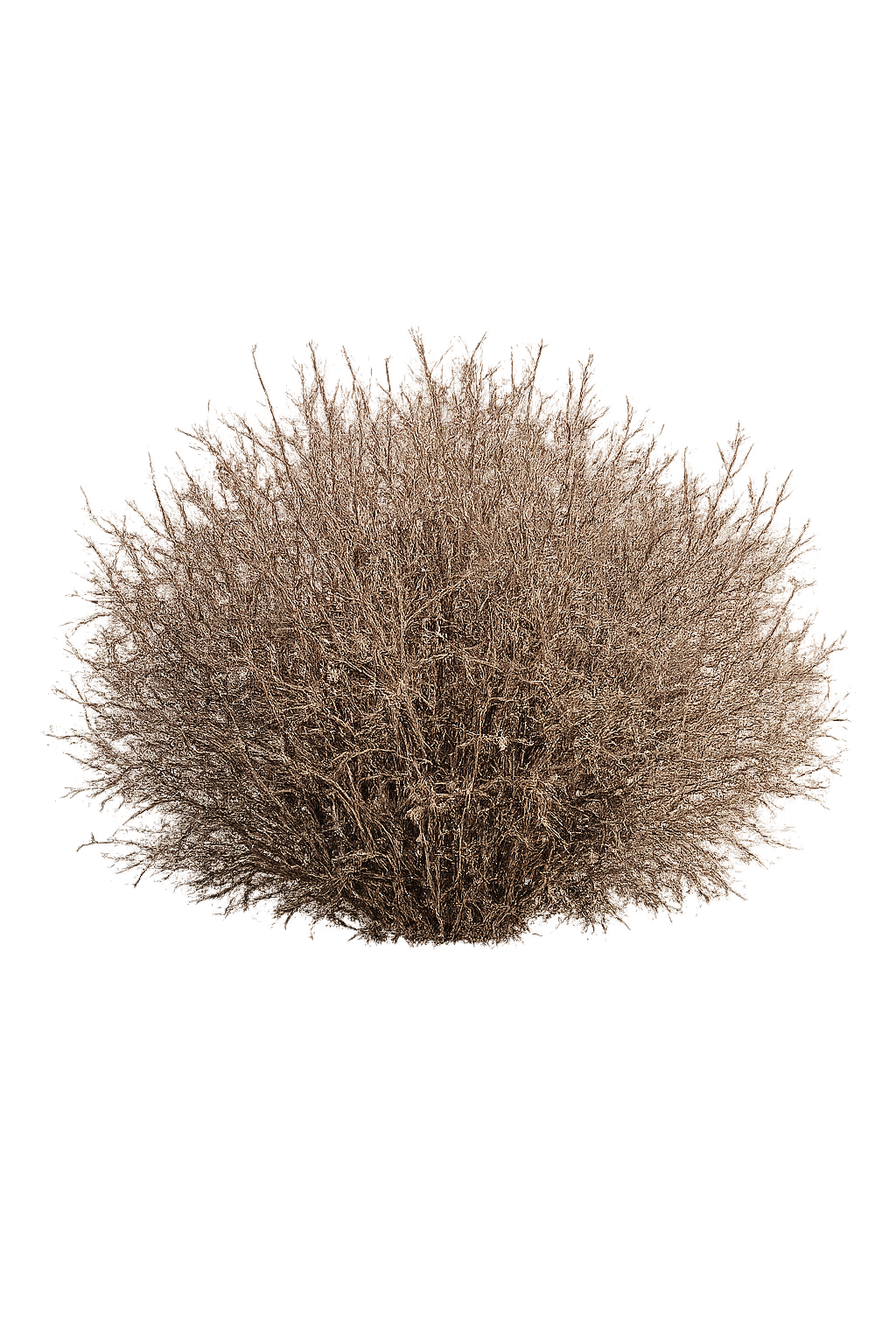}\\[-0.15em]{\tiny\sffamily base}\end{minipage}\hfill
\begin{minipage}[t]{0.62in}\centering\includegraphics[width=0.50in,height=0.50in,keepaspectratio]{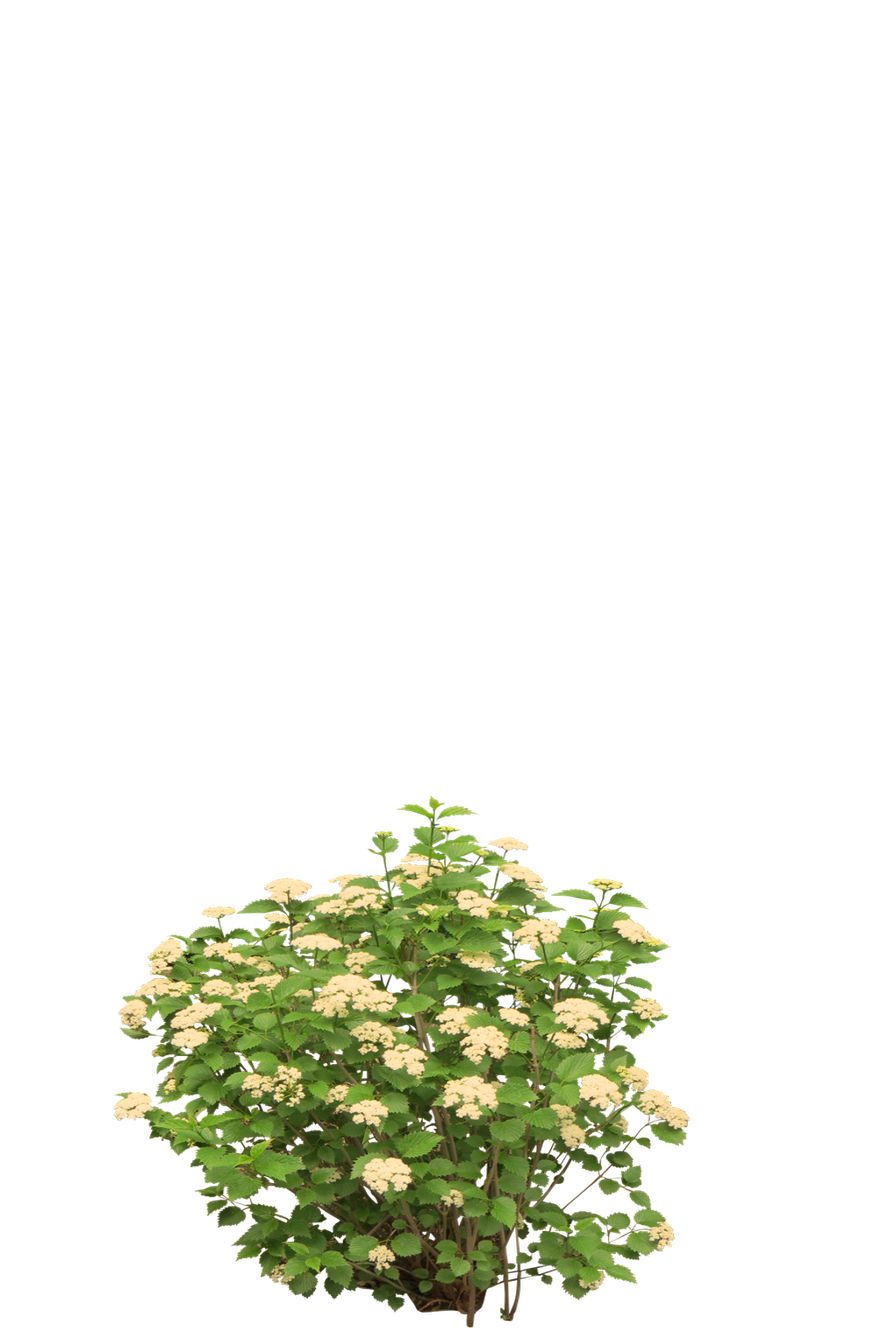}\\[-0.15em]{\tiny\sffamily low}\end{minipage}\hfill
\begin{minipage}[t]{0.62in}\centering\includegraphics[width=0.50in,height=0.50in,keepaspectratio]{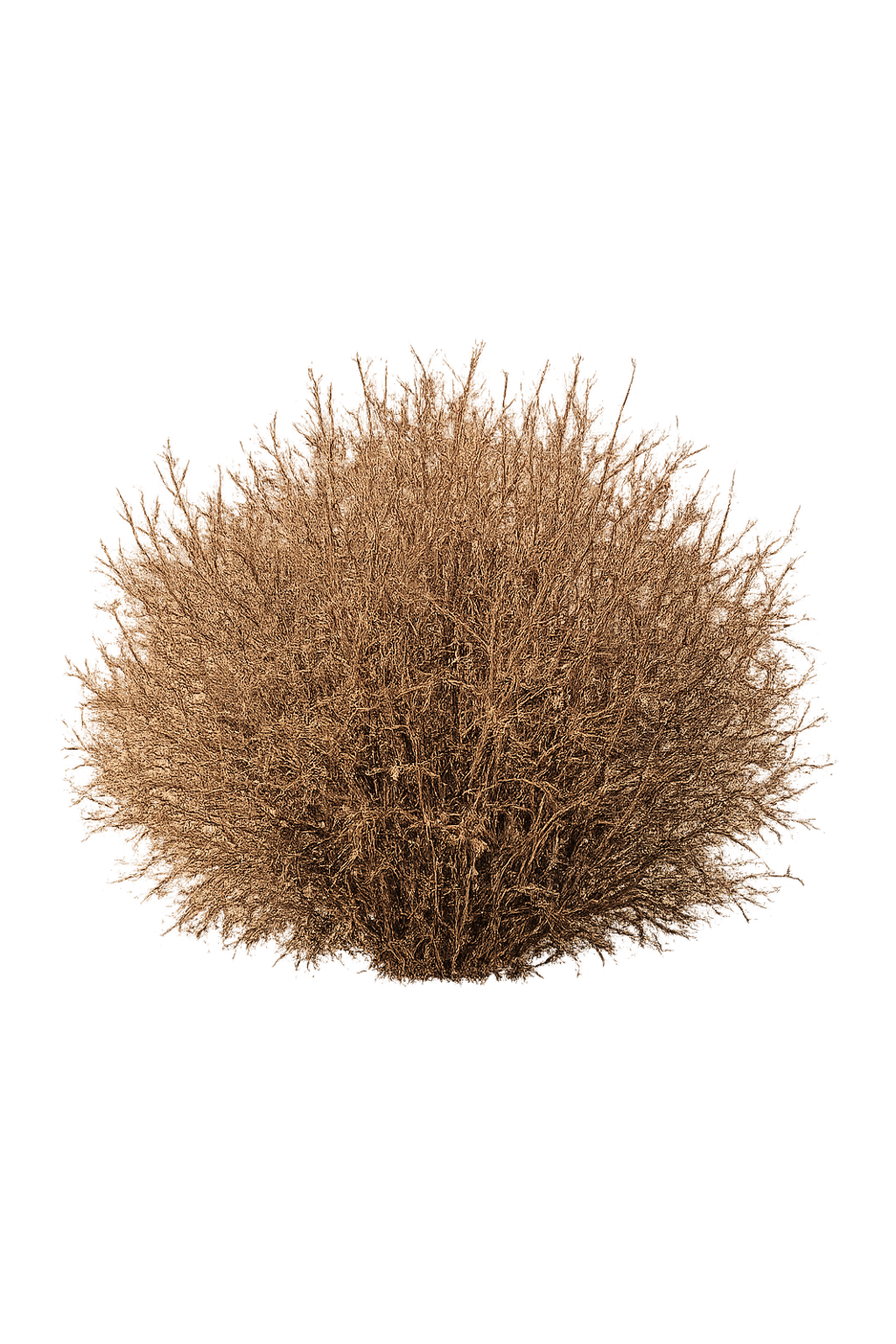}\\[-0.15em]{\tiny\sffamily tangled}\end{minipage}
& \codes{4, 5, 6, 7, 11, 12, 13} & Expanded from the dry-shrub foliage archetype. \\

Shared tree pool &
\begin{minipage}[t]{0.62in}\centering\includegraphics[width=0.50in,height=0.50in,keepaspectratio]{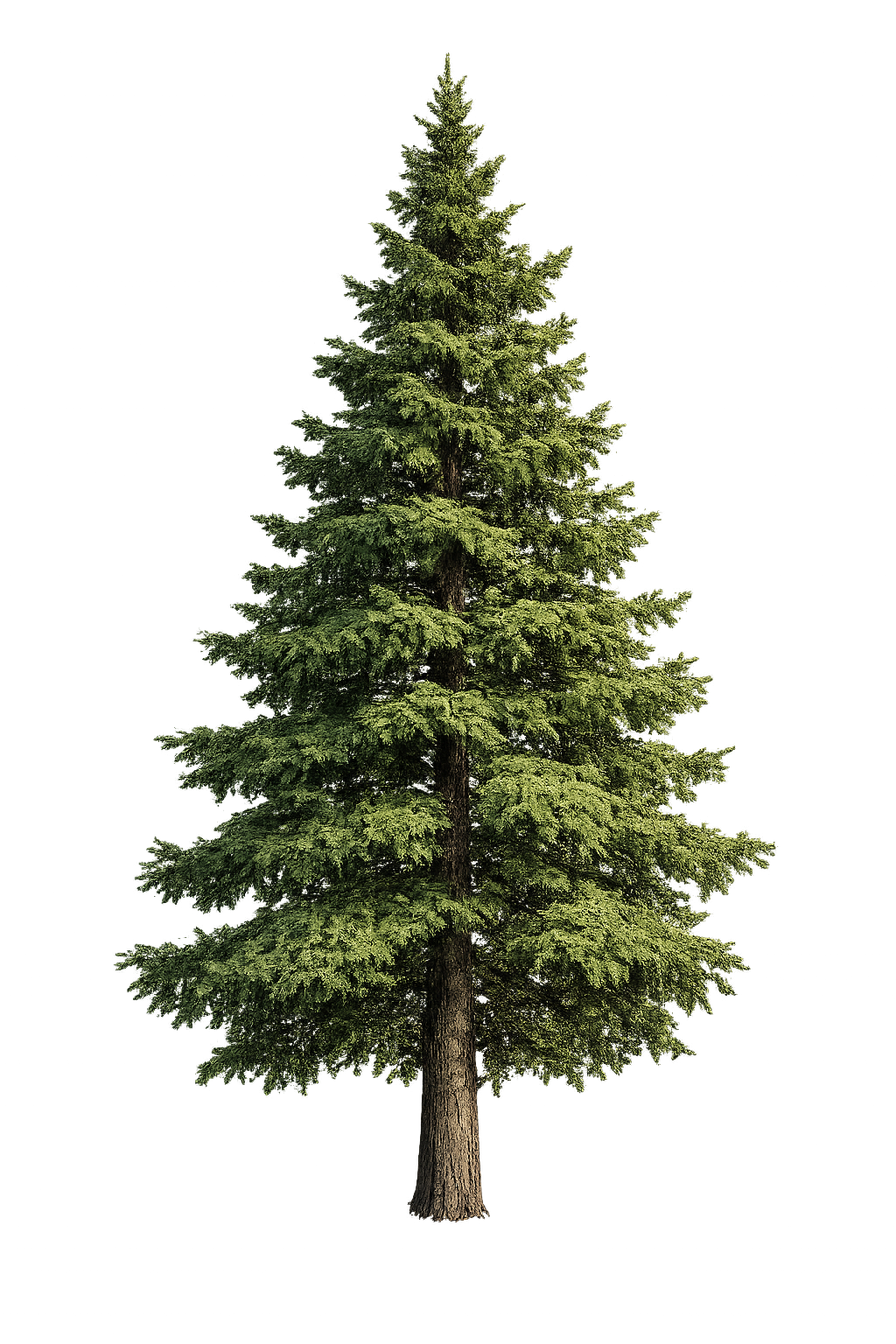}\\[-0.15em]{\tiny\sffamily conifer}\end{minipage}\hfill
\begin{minipage}[t]{0.62in}\centering\includegraphics[width=0.50in,height=0.50in,keepaspectratio]{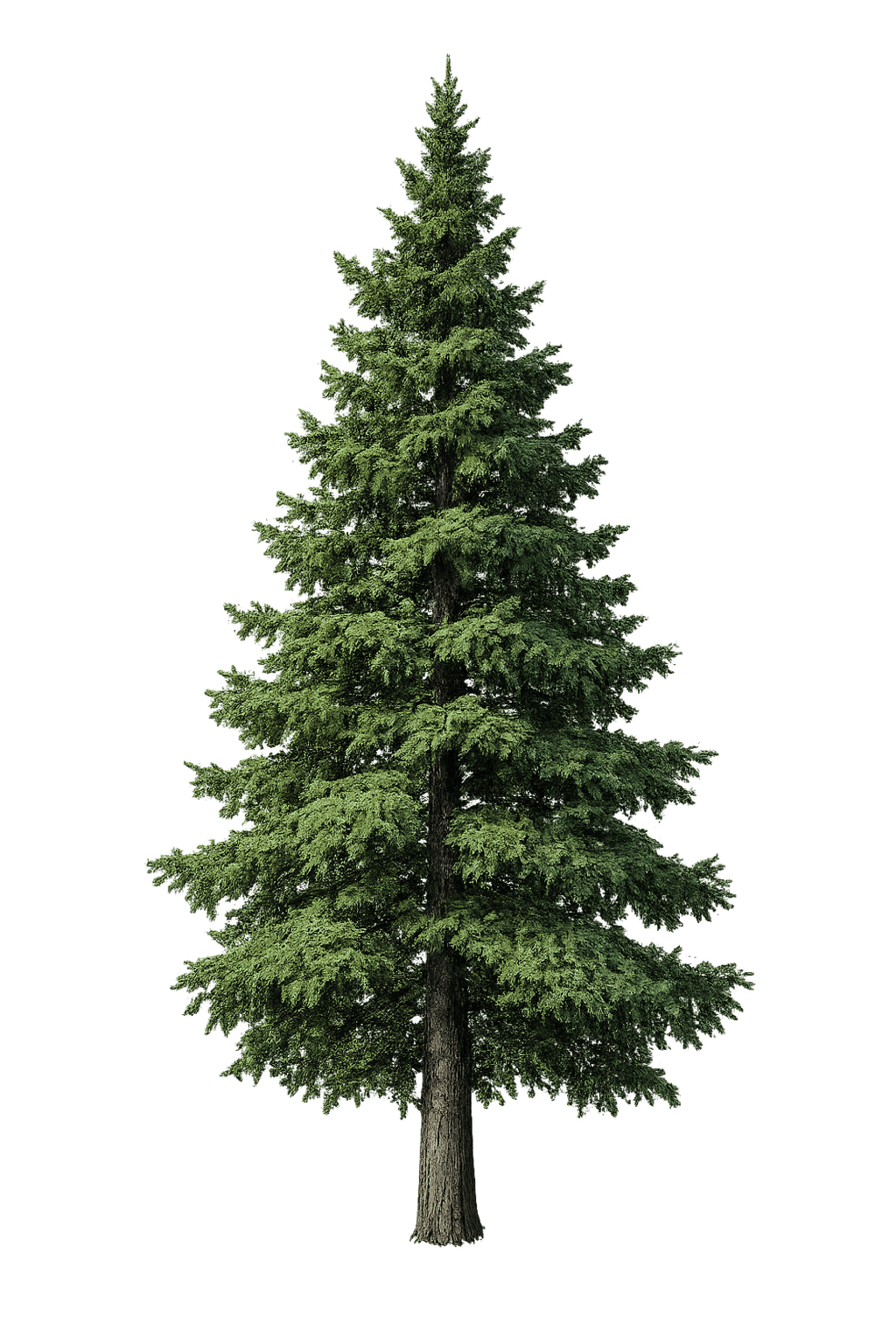}\\[-0.15em]{\tiny\sffamily blue}\end{minipage}\hfill
\begin{minipage}[t]{0.62in}\centering\includegraphics[width=0.50in,height=0.50in,keepaspectratio]{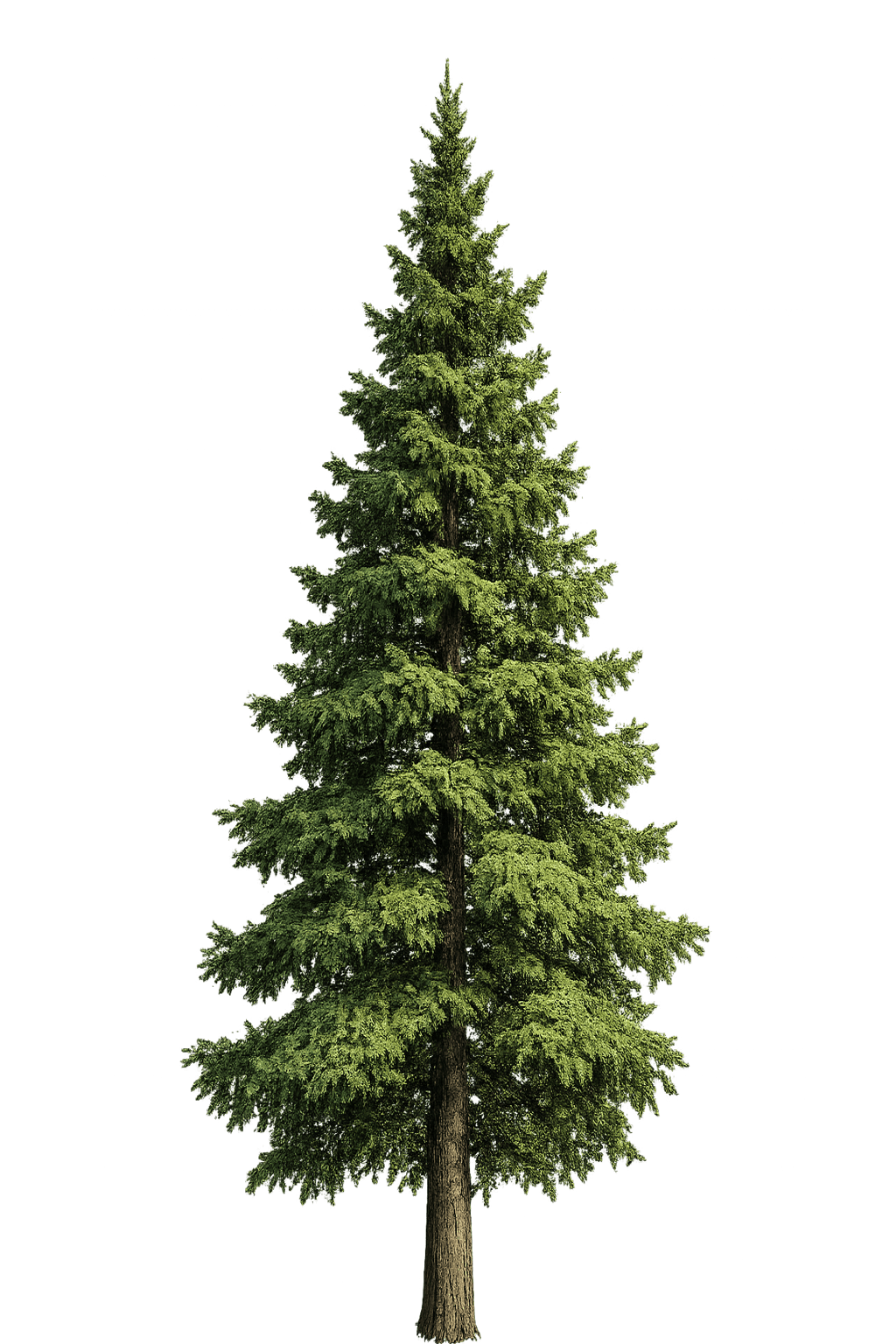}\\[-0.15em]{\tiny\sffamily narrow}\end{minipage}\hfill
\begin{minipage}[t]{0.62in}\centering\includegraphics[width=0.50in,height=0.50in,keepaspectratio]{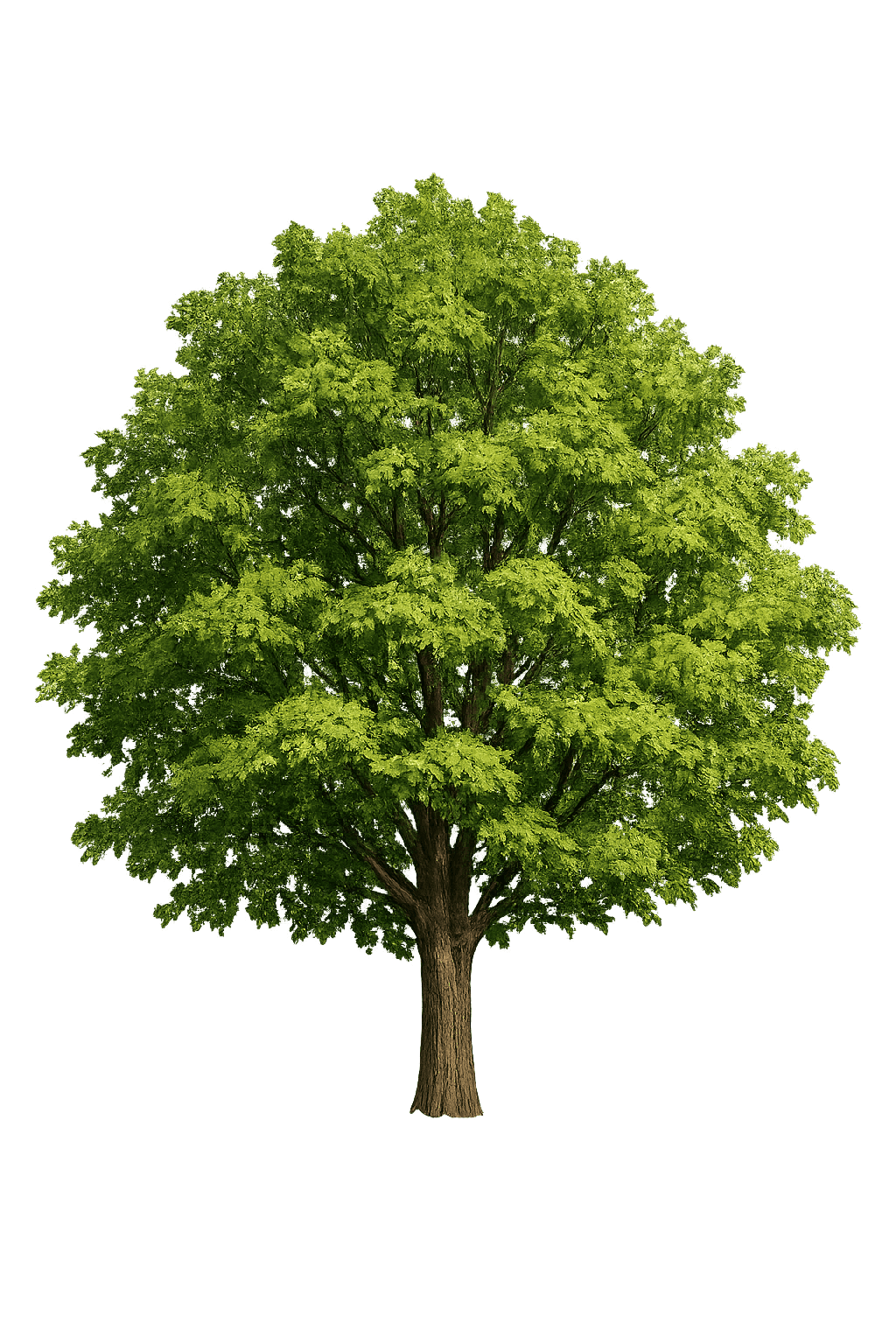}\\[-0.15em]{\tiny\sffamily hardwood}\end{minipage}
& \codes{2, 3, 7, 8, 9, 10, 11} & Current student pipeline expands tree references into this shared active pool. \\

Slash &
\begin{minipage}[t]{0.62in}\centering\includegraphics[width=0.50in,height=0.50in,keepaspectratio]{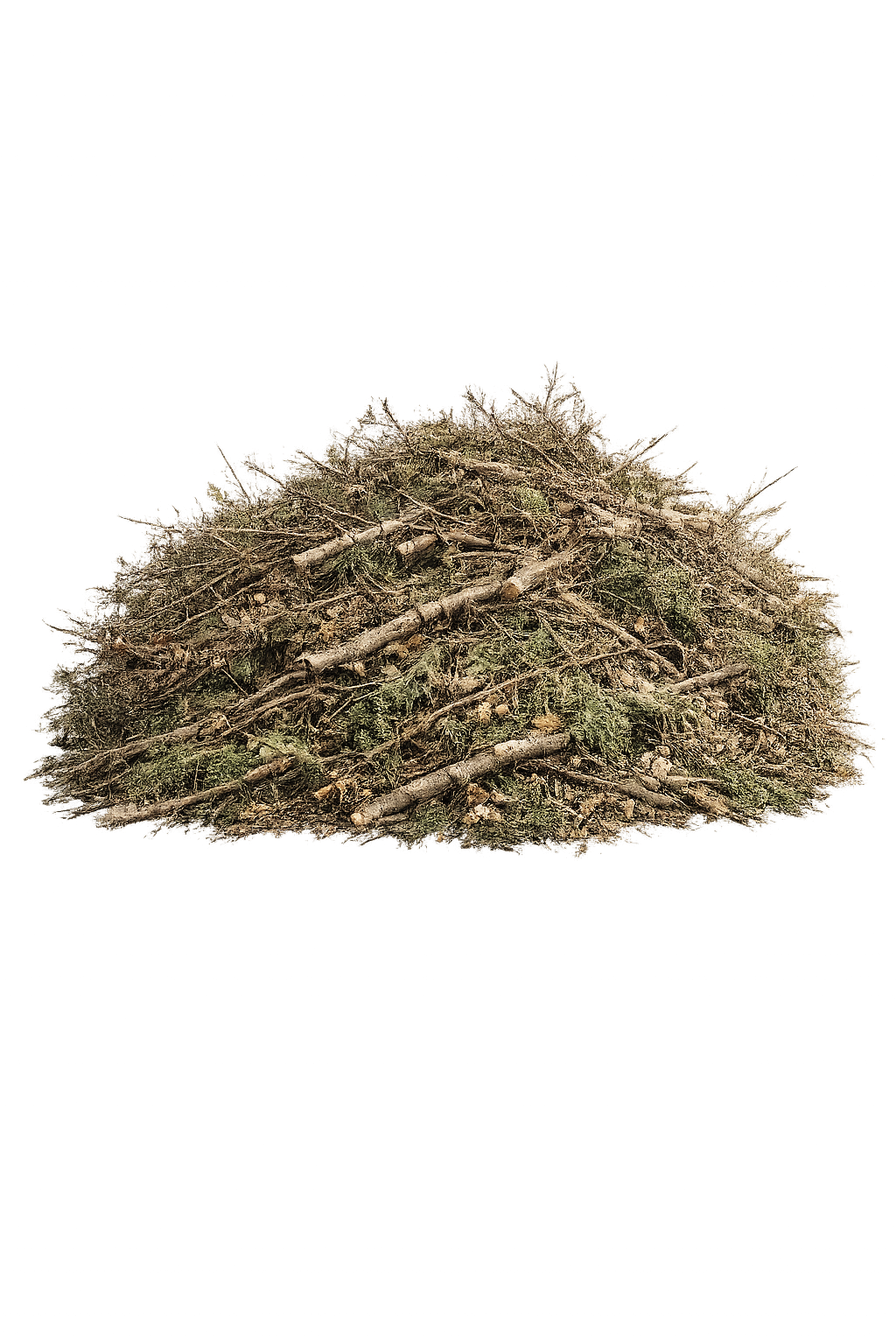}\\[-0.15em]{\tiny\sffamily base}\end{minipage}\hfill
\begin{minipage}[t]{0.62in}\centering\includegraphics[width=0.50in,height=0.50in,keepaspectratio]{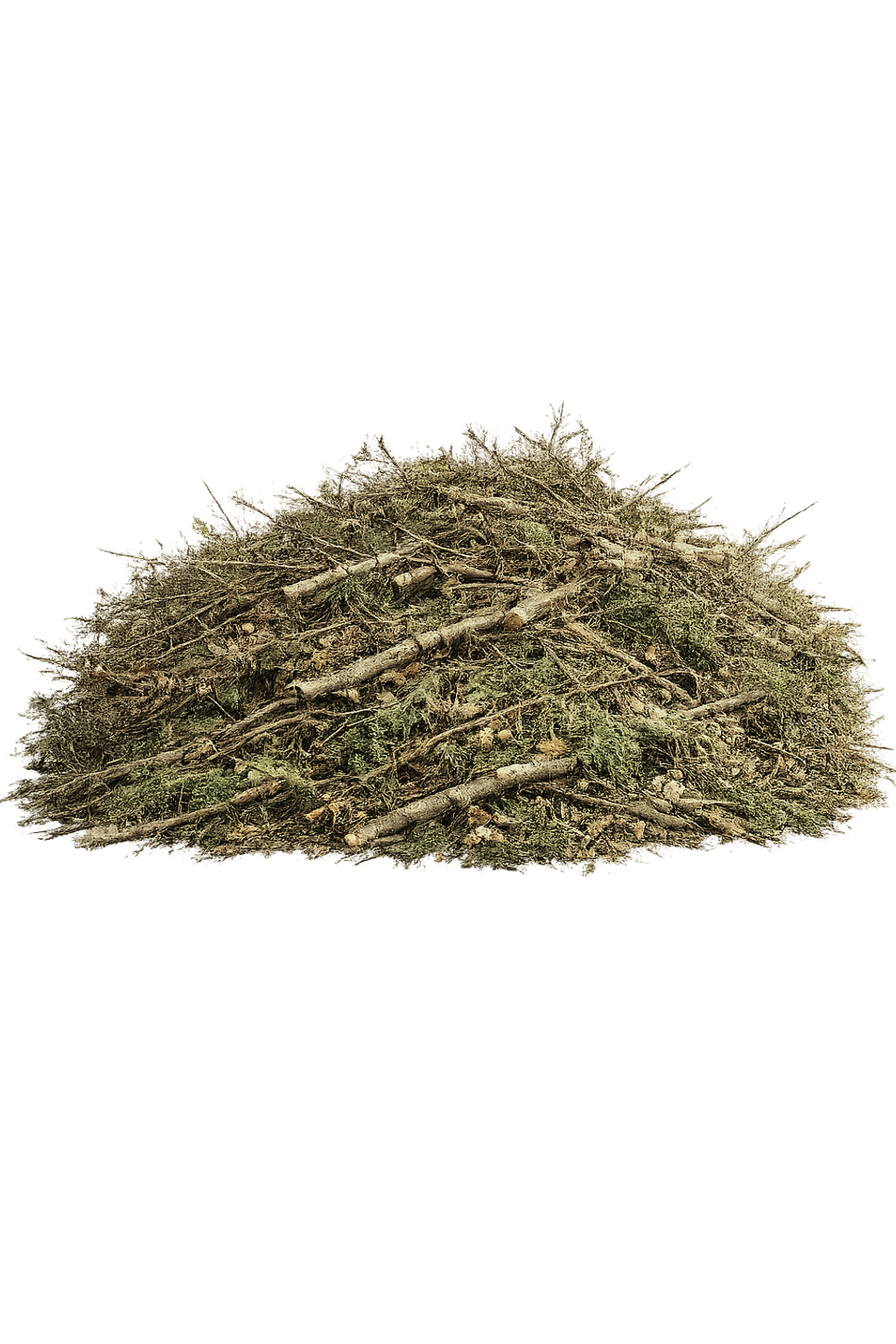}\\[-0.15em]{\tiny\sffamily fresh}\end{minipage}\hfill
\begin{minipage}[t]{0.62in}\centering\includegraphics[width=0.50in,height=0.50in,keepaspectratio]{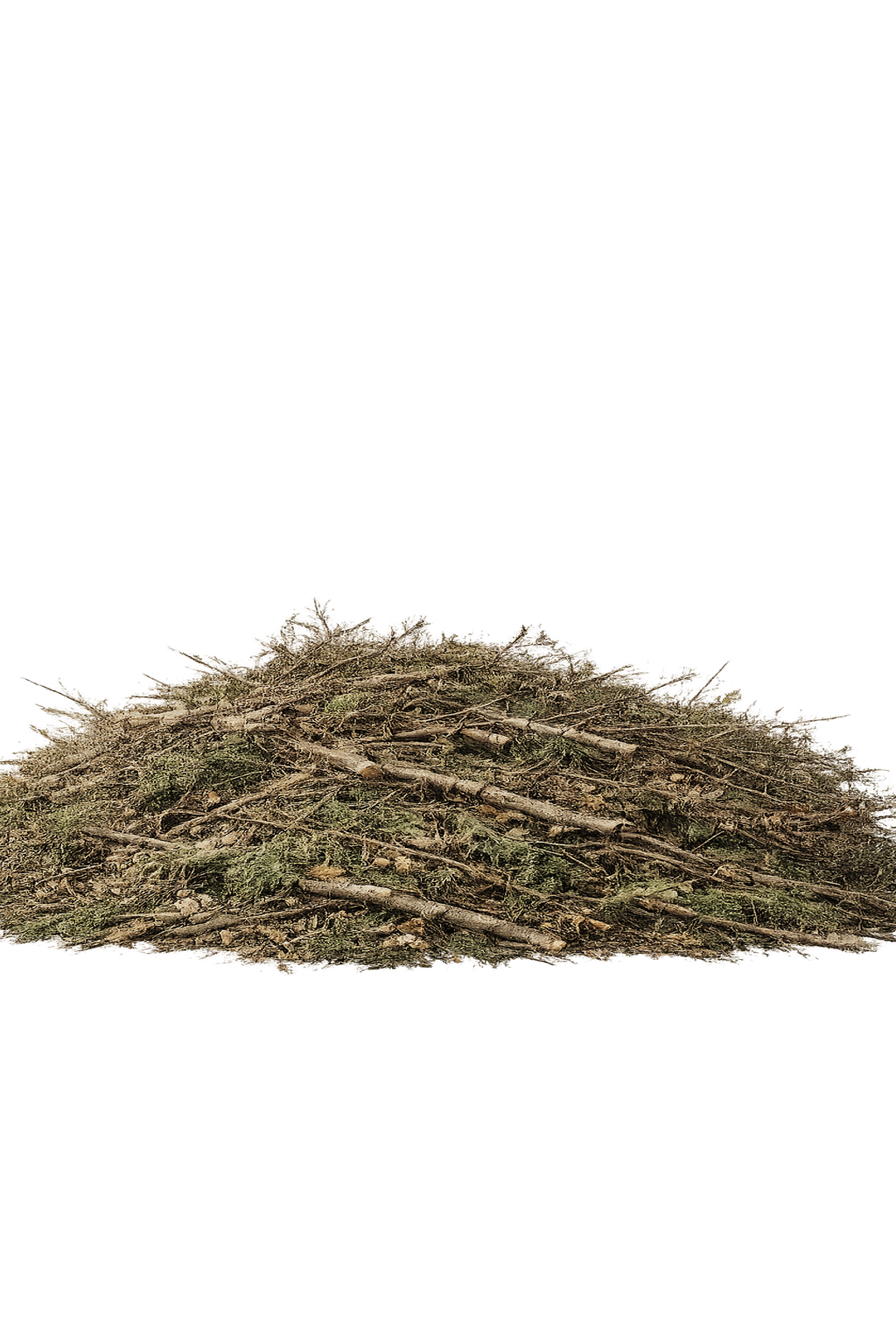}\\[-0.15em]{\tiny\sffamily scattered}\end{minipage}
& \codes{6, 11, 12, 13} & Expanded from the slash foliage archetype. \\

Fire and smoke effects &
\begin{minipage}[t]{0.62in}\centering\includegraphics[width=0.50in,height=0.50in,keepaspectratio]{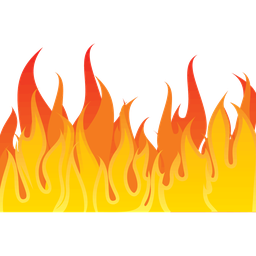}\\[-0.15em]{\tiny\sffamily flame}\end{minipage}\hfill
\begin{minipage}[t]{0.62in}\centering\includegraphics[width=0.50in,height=0.50in,keepaspectratio]{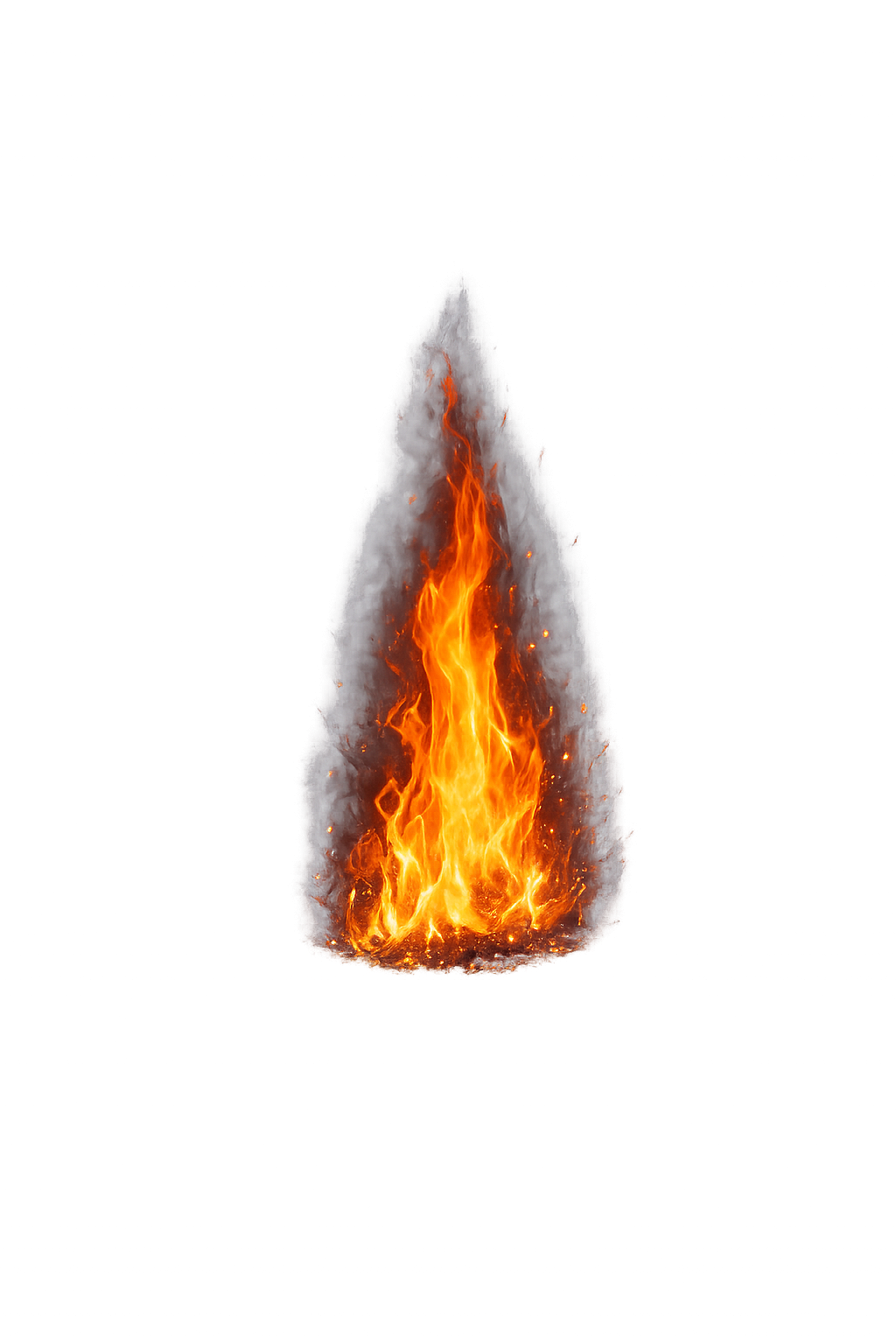}\\[-0.15em]{\tiny\sffamily ember}\end{minipage}\hfill
\begin{minipage}[t]{0.62in}\centering\includegraphics[width=0.50in,height=0.50in,keepaspectratio]{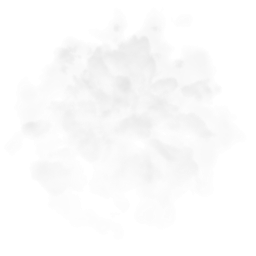}\\[-0.15em]{\tiny\sffamily smoke}\end{minipage}
& \codes{--} & Effect billboards for fire or smoke rendering; not tied to Anderson fuel codes. \\

\bottomrule
\end{longtable}
\endgroup

\section{Implementation Details}
\label{sec:implementation_details}
In this section, we provide additional implementation details of the proposed framework. 

\begin{figure}[htbp]
    \centering
    
    \includegraphics[width=0.7\textwidth]{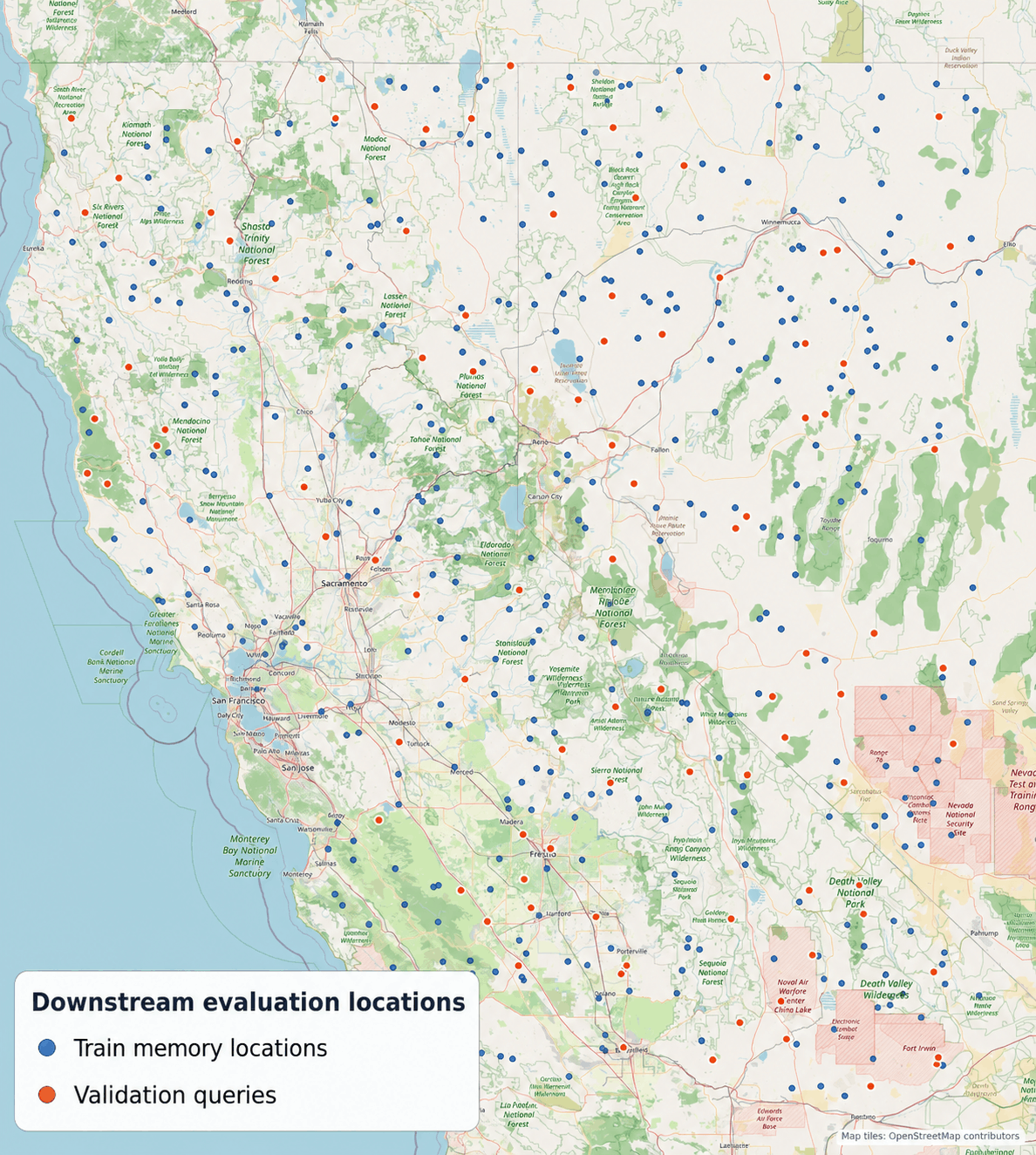}
    \caption{Geographic locations for memory construction and held-out evaluation; no VLM weights are updated.}
    \label{fig:locations_app}
    
\end{figure}

\paragraph{2D Simulation.}
For the 2D simulation stage, we use SimFire \cite{Doyle_SimFire_2024}. All simulations are conducted on a \(3\,\mathrm{km}\) domain, with each grid cell corresponding to approximately \(30\,\mathrm{m}\). The simulation horizon is set to represent one hour of wildfire propagation. For the four major factors affecting the simulations, we define low wind speed as below \(3\,\mathrm{mph}\) and high wind speed as above \(3\,\mathrm{mph}\). Low moisture is defined in the range \([0.0001, 0.005)\), while high moisture is defined in the range \([0.005, 0.01]\). For vegetation coverage and slope, we first query real-world fuel maps and terrain data for each location, and then perform post hoc categorization based on the resulting measurements. Under this procedure, the dataset is approximately divided into low vegetation coverage for values below \(6.96\%\), and high vegetation coverage for values between \(6.96\%\) and \(70.77\%\). Similarly, low slope corresponds to values between \(0^\circ\) and \(9.10^\circ\), while high slope corresponds to values between \(9.10^\circ\) and \(34.94^\circ\). We additionally show the queried 500 physical locations in Figure~\ref{fig:locations_app}.

\paragraph{3D Rendering.}
We use Blender to perform proxy 2D-to-3D rendering. For geometry, the elevation fields queried from LANDFIRE are converted into 3D terrain surfaces. For vegetation, following the Anderson 13 fuel model, we collect multiple vegetation and terrain textures and represent different fuel types using billboards. If a grid cell is labeled as burning in the 2D simulation, we randomly sample fire billboards within that cell and additionally render tilted smoke billboards according to the wind speed and wind direction. For camera placement, we choose aerial viewpoints that can observe the initial ignition region while resembling footage captured by a UAV. Finally, all videos are rendered in Blender using the Eevee renderer.

\paragraph{3D Generation.}
We use VACE~\cite{jiang2025vace} as the video generation backbone. Following the preprocessing strategy of~\cite{chen2025fluid}, we compute both depth and scribble control videos from proxy videos rendered in Blender. To better align the controls with the domain expected by VACE, we estimate depth using~\cite{yang2024depth} and extract scribble signals using~\cite{chan2022learning}. To screen depth alignment, we measure the Spearman correlation between the generated video depth and the proxy-rendered depth. If the score is below 0.35, we retry generation up to three times; if all retries fail, we resample a new location and repeat the full generation pipeline.

\paragraph{Memory System.}
We use Qwen3-2B-VL as the embedding model. As shown in Sec.~\ref{sec:embedding}, this model provides more accurate results than the other embedding models considered in our setting. We use GPT-5 as the reranker; the corresponding prompt is provided in Appendix~\ref{sec:candidate-reranking}.

\paragraph{Multi-Agent Reporting}
\label{sec:impl_multiagent_reporting}
The memory split supplies reference episodes and train-derived lookup priors without updating VLM weights. The top $k=5$ retrieved episodes are aggregated to compute task-wise consensus scores and metadata summaries for adjudication, whereas four-tag prediction in Table~\ref{tab:tag_accuracy} uses the top $k=1$ retrieved episode. The three reporters correspond to different prompt contexts: the visual reporter receives only sampled query frames; the lookup reporter receives query frames and the top retrieved tag-bundle prior; and the combined-memory reporter receives query frames, the lookup prior, and retrieved simulator-derived summaries. For each report field, unanimous reporter outputs are copied directly into the final prediction vector, while disputed fields are sent to the judge. The judge prompt includes the sampled frames, reporter predictions, short rationales, locked unanimous fields, top-$k$ memory consensus, and retrieved metadata summaries, and is constrained to revise only disputed fields. The resolved predictions, routing decisions, rationales, and retrieval summaries are serialized and rendered into an HTML report for each query.

\section{Simulation to Observation Alignment Numerical Metrics}
\label{sec:2d_3d_app}
Here, we quantitatively assess whether the VACE generation pipeline follows the provided proxy render. We evaluate this by testing whether the generated video is more consistent with its paired render than with a mismatched render. The evaluation metrics are summarized in Table~\ref{tab:metrics_vace}, and the corresponding results are reported in Table~\ref{tab:vace_result}. Higher agreement with paired than mismatched renders indicates sensitivity to the intended controls. These relative image-alignment measurements do not establish exact physical equivalence between generated videos and simulator trajectories.

\begingroup
\rowcolors{2}{TableStripe}{white}
\small
\begin{longtable}{L{0.2\linewidth} L{0.1\linewidth} L{0.6\linewidth}}
\caption{Proxy--generated alignment metrics.}\label{tab:metrics_vace}\\
\rowcolor{TableHead}\textbf{Metric} & \textbf{Directions} & \textbf{Meaning}\\
\midrule
\endfirsthead
\rowcolor{TableHead}\textbf{Metric} & \textbf{Directions} & \textbf{Meaning}\\
\midrule
\endhead
SSIM & $\uparrow$  & Grayscale structural similarity between aligned VACE and render frames.\\
Edge F1 / Edge-control F1 & $\uparrow$  & Canny-edge overlap between VACE and render frames with a small dilation tolerance. \\
RGB L1 & $\downarrow$  & Mean absolute RGB difference over aligned frames. \\
LPIPS & $\downarrow$ & AlexNet LPIPS perceptual distance averaged over aligned frames.\\
Correct-pair win rate & $\uparrow$  & Fraction of samples for which the paired render is closer to the VACE output than a mismatched render. \\
\bottomrule
\end{longtable}
\endgroup

\begingroup
\rowcolors{2}{TableStripe}{white}
\small
\begin{longtable}{@{}L{0.09\linewidth} L{0.09\linewidth} L{0.09\linewidth} L{0.09\linewidth} L{0.09\linewidth} L{0.09\linewidth} L{0.09\linewidth} L{0.09\linewidth}@{}}
\caption{Proxy--generated alignment and correct-pair win rates.}\label{tab:vace_result}\\
\rowcolor{TableHead}\textbf{SSIM} & \textbf{SSIM Win Rate} & \textbf{Edge F1} & \textbf{Edge F1 Win Rate} & \textbf{RGB L1} & \textbf{RGB L1 Win Rate} & \textbf{LPIPS} & \textbf{LPIPS Win Rate}\\
\midrule
\endfirsthead
\rowcolor{TableHead}\textbf{SSIM} & \textbf{SSIM Win Rate} & \textbf{Edge F1} & \textbf{Edge F1 Win Rate} & \textbf{RGB L1} & \textbf{RGB L1 Win Rate} & \textbf{LPIPS} & \textbf{LPIPS Win Rate}\\
\midrule
\endhead
0.334 & 76.6\% & 0.612 & 81.3\% & 0.189 & 72.0\% & 0.516 & 99.1\% \\
\bottomrule
\end{longtable}
\endgroup

We also compare two image-space directional cues between generated and proxy videos. \textbf{Smoke plume axis} Smoke-colored pixels are extracted in
each sampled frame. We then use PCA to give the dominant plume axis. Axes are averaged over frames and compared as undirected axes. \textbf{Growth-front fire-spread axis} Flame pixels from the
first third of the clip are unioned into an early-fire mask. Flame pixels from
the last third are unioned into a late-fire mask. Pixels already near the early
mask are removed from the late mask, giving a late-growth region. The vector
from the early-fire centroid to the late-growth centroid estimates spread
direction. We then compare the generated and rendered videos fire spread axis. 
\begingroup
\rowcolors{2}{TableStripe}{white}
\small
\begin{longtable}{L{0.3\linewidth} L{0.1\linewidth} L{0.1\linewidth} L{0.1\linewidth} L{0.1\linewidth} L{0.1\linewidth}}
\rowcolor{TableHead} \textbf{Metric} & \(\le 30^\circ\) & \(\le 45^\circ\) & \(\le 60^\circ\) & Median & Mean \\
\midrule
\endfirsthead
\rowcolor{TableHead}\textbf{Metric} & \(\le 30^\circ\) & \(\le 45^\circ\) & \(\le 60^\circ\) & Median & Mean \\
\midrule
\endhead
Smoke plume axis & 95.8\% & 97.4\% & 99.2\% & 3.7 & 7.6 \\
Growth-front fire-spread axis & 47.1\% & 67.3\% & 79.0\% & 32.7 & 35.8 \\
\bottomrule
\end{longtable}
\endgroup

The mean errors are $7.6^\circ$ for the smoke-plume axis and $35.8^\circ$ for the growth-front fire-spread axis. The larger growth-direction deviation limits a claim of full dynamic preservation. These image-space cues do not measure physical spread rate or burned-area trajectories.

\subsection{VLM-Based Observable Cue Agreement}
\label{app:cue-preservation}

We use GPT-5.6-luna to assess agreement between sampled frames from a
proxy video and its corresponding conditioned generated video. Each
pair is scored from 0 to 100 for fire-front location, fuel distribution,
burned area, and terrain structure. Table~\ref{tab:cue-preservation}
reports the average scores. Identical proxy--proxy pairs score
approximately 95, while identical generated--generated pairs score
approximately 90. These identity comparisons contextualize the
ratings; the scores are not calibrated percentages of physical
fidelity. This assessment complements the numerical alignment metrics
above rather than establishing exact physical equivalence.

\begin{table}[h]
\centering
\small
\caption{VLM-assessed agreement on a 0--100 scale. Identity-pair values
are approximate overall averages, not per-cue scores.}
\label{tab:cue-preservation}
\begin{tabular}{@{}lr@{}}
\toprule
Cue or comparison & Score \\
\midrule
Fire-front location & 79.17 \\
Fuel distribution & 68.75 \\
Burned area & 70.83 \\
Terrain structure & 81.25 \\
Overall & 75.00 \\
\midrule
Identical proxy--proxy pairs & $\approx 95$ \\
Identical generated--generated pairs & $\approx 90$ \\
\bottomrule
\end{tabular}
\end{table}

\section{Model-to-Prompt Mapping}
\label{sec:prompts_agent}
\textbf{Reading guide.} Blue boxes denote system prompts, green boxes user prompts, orange boxes schemas and output contracts, purple or teal boxes task and wrapper prompts, red boxes negative prompts, and gray boxes assembly or payload snippets.
\begingroup
\rowcolors{2}{AccentSoft}{white}
\renewcommand{\arraystretch}{1.12}
\begin{longtable}{@{}L{0.28\linewidth} L{0.28\linewidth} L{0.36\linewidth}@{}}
\rowcolor{Accent!12}
\textbf{Model family} & \textbf{Evaluation role} & \textbf{Prompt family} \\
\midrule
\endfirsthead
\rowcolor{Accent!12}
\textbf{Model family} & \textbf{Evaluation role} & \textbf{Prompt family} \\
\midrule
\endhead
GPT-4.1 / GPT-5.2 & direct query-only four-tag inference & shared query-only tag prompt in Section~\ref{sec:query-only-openai} \\
Gemini 2.5 Pro & direct query-only four-tag inference & shared query-only prompt plus Gemini/Claude JSON wrapper in Section~\ref{sec:query-only-external} \\
Claude Sonnet 4 & direct query-only four-tag inference & shared query-only prompt plus Gemini/Claude JSON wrapper in Section~\ref{sec:query-only-external} \\
Qwen2-Wildfire-2B & local VLM four-tag inference; six-task downstream conditioned on its own predicted tags & local VLM binary-tag prompt in Section~\ref{sec:local-vlm}; downstream VLM prediction in Section~\ref{sec:vlm-downstream-conditioned} \\
ForestFireVLM-7B & local VLM four-tag inference; six-task downstream conditioned on its own predicted tags & local VLM binary-tag prompt in Section~\ref{sec:local-vlm}; downstream VLM prediction in Section~\ref{sec:vlm-downstream-conditioned} \\
GPT-5-mini & downstream specialist inference & downstream specialist system + context blocks in Section~\ref{sec:downstream-specialists} \\
GPT-5-mini family & multi-agent disputed-task resolution & multi-agent judge prompt in Section~\ref{sec:multiagent-judge} \\
GPT-5-mini & final HTML report narrative generation & report-generation prompt in Section~\ref{sec:html-report} \\
GPT-5-mini & video-generation prompt rewriting & prompt-refinement template in Section~\ref{sec:prompt-refine} \\
Qwen3-VL embedding / reranker & hybrid retrieval and optional reranking & Qwen retrieval/rerank formatting in Section~\ref{sec:qwen-rerank} \\
\bottomrule
\end{longtable}
\endgroup

\section{Video Generation Prompt}
\label{sec:video_generation_prompts}
The following templates document the controls used to construct the reference episodes. Environmental tags influence both positive and negative prompts, so generated appearance reflects the prompt design as well as the simulated state and proxy geometry. The templates are reproduced as experimental records; requested realism and motion characteristics are generation objectives, not guarantees of physical fidelity.
\subsection{Video Generation Prompt From Simulation Data}
\paragraph{Raw Prompt Template}

The raw prompt uses a fixed sentence template with simulation-conditioned slots:
\begin{PromptBlock}[prompt/system]
Photorealistic aerial drone video from a distant, stabilized stand-off viewpoint
over a broad natural landscape, showing <duration> of wildfire propagation with
the terrain, smoke plume, and fire perimeter kept readable in a wide frame while
the active flames remain visible. Surface cover should read as <burnable cover>,
with <barriers> acting as fire breaks; fuel continuity is <continuity> and
barrier influence is <barrier influence>. <vegetation density text>
<built environment text> Use <wind strength> winds moving toward <wind direction>,
<moisture>, and <terrain>. <flame visibility>. Depict a <run size> fire with
<fire motion>, <smoke motion>, realistic flank slowdowns at fragmented fuels and
barrier edges and <spotting>. Natural
daylight, cinematic realism, and physically grounded motion.
\end{PromptBlock}

\paragraph{Fuel and Barrier Description}

The fuel map is summarized by pixel count and percentage for each fuel code.
Anderson-13 burnable fuel codes are mapped to natural-language land-cover names,
and common non-burnable codes are mapped to barrier names.
\begingroup
\rowcolors{2}{TableStripe}{white}
\small
\begin{longtable}{L{0.2\linewidth} L{0.4\linewidth} L{0.3\linewidth}}
\rowcolor{TableHead}\textbf{Input} & \textbf{Decision rule} & \textbf{Prompt effect} \\
\midrule
\endfirsthead
\rowcolor{TableHead}\textbf{Input} & \textbf{Decision rule} & \textbf{Prompt effect} \\
\midrule
\endhead
Fuel percentage by code &
The top four burnable land-cover groups are selected after grouping by semantic cover type. &
Forms the phrase ``Surface cover should read as ...'' \\

Non-burnable percentage by code &
The top three non-burnable groups are selected after grouping by semantic barrier type. &
Forms the phrase ``with ... acting as fire breaks.'' \\

Cover percentage $p$ &
$p \ge 60$: dominant; $30 \le p < 60$: major; $10 \le p < 30$: moderate;
$3 \le p < 10$: minor; $p < 3$: trace. &
Adds qualitative size words such as ``major brushy shrubland'' or
``minor open water.'' \\

Burnable percentage $b$ &
$b < 60$: fragmented; $60 \le b < 85$: partially interrupted;
$b \ge 85$: mostly continuous. &
Sets the fuel-continuity phrase. \\

Non-burnable percentage $n$ &
$n \ge 25$: substantial; $10 \le n < 25$: noticeable; $n < 10$: limited. &
Sets the barrier-influence phrase. \\
\bottomrule
\end{longtable}
\endgroup

\paragraph{Urban Visibility}

The city overlay count controls how strongly the
prompt asks for visible urban context. Urban pixels are not duplicated in the
generic barrier list when the city is already visible, clear, or prominent.

\paragraph{Wind, Smoke, and Spotting}

Wind speed and direction are read from simulation metadata. Direction in
degrees is converted to one of sixteen compass directions. Wind speed controls
the wind-strength phrase, smoke behavior, fire-motion wording, flame-visibility
wording, spotting behavior, and the negative prompt. These are appearance controls: in particular, spotting language does not mean that ember trajectories or spotting dynamics are simulated by the current surface-spread model.

Fire-motion text combines wind speed with spread intensity. For very light wind,
fast or very fast spread is described as ``steady to locally fast flame-front
growth driven mainly by fuels and slope rather than wind-whipped runs.'' For
moderate wind, the wording becomes ``noticeable downwind push.'' For strong or
very strong wind, the wording becomes ``wind-driven flame-front growth with
pronounced downwind push.''

\begingroup
\rowcolors{2}{TableStripe}{white}
\small
\begin{longtable}{@{}L{0.15\linewidth} L{0.15\linewidth} L{0.3\linewidth} L{0.3\linewidth}@{}}
\rowcolor{TableHead}\textbf{Wind speed} & \textbf{Wind phrase} & \textbf{Smoke phrase} & \textbf{Spotting phrase} \\
\midrule
\endfirsthead
\rowcolor{TableHead}\textbf{Wind speed} & \textbf{Wind phrase} & \textbf{Smoke phrase} & \textbf{Spotting phrase} \\
\midrule
\endhead
$<0.75$ m/s & very light &
Mostly vertical smoke column with gentle drift and limited sideways shear. &
Only sparse short-range ember lofting where continuity supports it. \\
\addlinespace
$0.75$--$<2.0$ m/s & light &
Mostly upright smoke plume with mild downwind drift and restrained lateral shear. &
Limited ember lofting and short-range spotting where continuity supports it. \\
\addlinespace
$2.0$--$<5.0$ m/s & moderate &
Clear downwind smoke advection with moderate plume tilt and lateral drift. &
Occasional ember spotting where continuity supports it. \\
\addlinespace
$5.0$--$<8.0$ m/s & strong &
Strong downwind smoke transport with a clearly bent plume and sustained lateral advection. &
Frequent downwind ember transport and intermittent spotting where continuity supports it. \\
\addlinespace
$\ge 8.0$ m/s & very strong &
Strong downwind smoke transport with a clearly bent plume and sustained lateral advection. &
Frequent downwind ember transport and intermittent spotting where continuity supports it. \\
\bottomrule
\end{longtable}
\endgroup

\paragraph{Moisture, Terrain, and Vegetation}

\begingroup
\rowcolors{2}{TableStripe}{white}
\small
\begin{longtable}{L{0.2\linewidth} L{0.4\linewidth} L{0.3\linewidth}}
\rowcolor{TableHead}\textbf{Feature} & \textbf{Decision rule} & \textbf{Prompt effect} \\
\midrule
\endfirsthead
\rowcolor{TableHead}\textbf{Feature} & \textbf{Decision rule} & \textbf{Prompt effect} \\
\midrule
\endhead
Moisture tag &
\texttt{low\_moisture}: very dry surface fuels;
\texttt{high\_moisture}: comparatively moister surface fuels. &
Tags override numeric moisture wording. \\
\addlinespace
Numeric moisture $m$ &
$m < 0.03$: very dry; $0.03 \le m < 0.06$: dry;
$0.06 \le m < 0.10$: moderately dry; $m \ge 0.10$: moist. &
Used when no explicit moisture tag is present. \\
\addlinespace
Elevation relief $r$ &
$r < 100$ m: gentle; $100 \le r < 300$ m: rolling;
$300 \le r < 600$ m: rugged; $r \ge 600$ m: very rugged. &
Sets the terrain phrase. Default wording says spread accelerates on favorable
slopes and eases where aspect is less favorable. \\
\addlinespace
Slope tag &
\texttt{high\_slope}: emphasizes steeper local slopes;
\texttt{low\_slope}: emphasizes broader gentler local slopes. &
Modifies terrain wording beyond the numeric relief class. \\
\addlinespace
Vegetation tag &
\texttt{high\_vegetation}: denser vegetation, continuous canopy, fewer gaps;
\texttt{low\_vegetation}: more open, sparse vegetation, patchier canopy. &
Adds a separate vegetation-density sentence. \\
\bottomrule
\end{longtable}
\endgroup

\paragraph{Negative Prompt Generation}

The negative prompt is also generated from the same simulation metadata. It
always includes ``camera shake'' and ``rapid viewpoint changes.'' Wind, moisture,
slope, and vegetation tags then add exclusions intended to discourage contradictory visual
behavior.
\begingroup
\rowcolors{2}{TableStripe}{white}
\small
\begin{longtable}{L{0.3\linewidth} L{0.6\linewidth} }
\rowcolor{TableHead}\textbf{Condition} & \textbf{Additional negative phrases} \\
\midrule
\endfirsthead
\rowcolor{TableHead}\textbf{Condition} & \textbf{Additional negative phrases} \\
\midrule
\endhead
$w < 0.75$ m/s &
strong wind; very strong wind; gale-force smoke; wind-whipped flames; violent
lateral smoke; long horizontal smoke bands; storm-like plume bending. \\
\addlinespace
$0.75 \le w < 2.0$ m/s &
strong wind; gale-force smoke; wind-whipped flames; extreme lateral smoke
advection; fully horizontal smoke plume. \\
\addlinespace
$2.0 \le w < 5.0$ m/s &
hurricane-force wind; perfectly still air; fully vertical plume with no drift. \\
\addlinespace
$w \ge 5.0$ m/s &
still air; calm smoke column; perfectly vertical plume with no downwind drift. \\
\addlinespace
\texttt{low\_moisture} &
rain-soaked fuels; wet ground everywhere. \\
\addlinespace
\texttt{high\_moisture} &
ultra-dry fuels; explosive firestorm behavior from uniformly desiccated
vegetation. \\
\addlinespace
\texttt{low\_slope} &
sheer mountainsides; cliff-dominated terrain. \\
\addlinespace
\texttt{high\_slope} &
perfectly flat plain; table-flat ground. \\
\addlinespace
\texttt{low\_vegetation} &
dense closed-canopy forest everywhere; jungle-thick vegetation. \\
\addlinespace
\texttt{high\_vegetation} &
nearly barren terrain; treeless empty plain. \\
\bottomrule
\end{longtable}
\endgroup

\subsection{Prompt Refinement for Video Generation}

\label{sec:prompt-refine}
As mentioned in \ref{sec:2d_3d}. This stage rewrites an existing wildfire video prompt into a cleaner text-to-video prompt while preserving the intended wildfire scenario.

\subsubsection{System Prompt for Prompt Refinement}
\begin{PromptBlock}[prompt/system]
You are an expert text-to-video prompt writer for realistic environmental simulations.
Rewrite user prompts to improve visual clarity, motion continuity, physical realism, and generation stability.
Keep the environmental meaning faithful.
Preserve explicit cues about visible city buildings or developed structures, aerial drone capture, distant stand-off viewpoint,
elapsed fire duration such as one hour, blackened burned land or burn scars, and a clearly visible active flame front when they are present.
Preserve explicit wind-strength cues. For low-wind or very-light-wind cases, keep smoke mostly upright with only gentle drift and do not rewrite them into wind-whipped or strongly horizontal plumes.
Do not let the rewrite imply that smoke fully hides the flames when the source prompt says the active fireline must stay readable.
For strong-wind cases, preserve clearly downwind smoke transport rather than weakening the wind effects.
Keep the rewrite compact and high-signal, ideally about 90 to 140 words in a single paragraph.
Avoid coordinates, fuel model names, numeric fuel codes, brittle shot-by-shot instructions, and markdown.
Return exactly one concise paragraph.
\end{PromptBlock}

\subsubsection{User Prompt Template for Prompt Refinement}
\begin{PromptBlock}[prompt/user]
Refine this wildfire video generation prompt for a modern text-to-video model.
Preserve the scenario and realism constraints while making it clearer and more producible:

<original prompt text>
\end{PromptBlock}

\paragraph{Message assembly.}
\begin{PromptBlock}[prompt/wrapper]
[system]  <prompt-refinement system prompt above>
[user]    <prompt-refinement user prompt above>
\end{PromptBlock}

\paragraph{Output contract.}
The response is constrained to exactly one concise paragraph, with low verbosity and reasoning effort in the OpenAI Responses API call.

\section{Direct Four-Tag Evaluation Prompts}

\subsection{OpenAI Query-Only Tag Prompt}
\label{sec:query-only-openai}
The VLM direct query baselines in Table~\ref{tab:tag_accuracy} use a shared instruction shown below. The OpenAI request is assembled as one user message containing this instruction block, followed by the query frames.
In later sections, we refer this prompt as \texttt{DEFAULT\_QUERY\_ONLY\_CLASSIFICATION\_INSTRUCTIONS}.
\begin{PromptBlock}[prompt/code]
Classify the query wildfire video into four binary tags using only visible evidence from the frames.
Decide each field independently.
Use smoke and flame drift or lean for wind, vegetation greenness or dryness and burn behavior for moisture, terrain incline and relief for slope, and plant density or fuel coverage for vegetation.
For each field, return a label of low or high and one short reason based on visible evidence.
\end{PromptBlock}

\paragraph{Message assembly.}
\begin{PromptBlock}[prompt/wrapper]
[input_text]  <instruction block above>
[input_text]  Query video to classify.
[input_text]  Query frame 1.
[input_image] <frame 1>
...
[input_text]  Query frame n.
[input_image] <frame n>
\end{PromptBlock}

\subsection{Gemini / Claude External Query-Only Prompt Wrapper}
\label{sec:query-only-external}
Gemini and Claude use the same core direct query instruction, but the text prompt is wrapped with an explicit JSON-format reminder.

\begin{PromptBlock}[prompt/code]
<DEFAULT_QUERY_ONLY_CLASSIFICATION_INSTRUCTIONS>
Return only a JSON object with exactly four keys: wind_regime_tag, moisture_regime_tag, slope_regime_tag, and vegetation_regime_tag. Each key must map to an object with exactly two keys: label and reason. The label must be either low or high. The reason must be one short sentence based only on visible evidence. Do not return markdown.
\end{PromptBlock}

\paragraph{Gemini message assembly.}
\begin{PromptBlock}[prompt/wrapper]
[text]        <wrapped instruction block above>
[text]        Query video to classify.
[inline_data] <frame 1>
...
[inline_data] <frame n>
\end{PromptBlock}

\paragraph{Claude message assembly.}
\begin{PromptBlock}[prompt/wrapper]
[text]   <wrapped instruction block above>
[text]   Query video to classify.
[image]  <frame 1>
...
[image]  <frame n>
\end{PromptBlock}

The default provider models in this evaluator are:
\begin{itemize}[leftmargin=1.5em]
\item Gemini: \texttt{gemini-2.5-pro}
\item Claude: \texttt{claude-sonnet-4-20250514}
\end{itemize}

\subsection{Local VLM Query-Only Prompt for Qwen2-Wildfire-2B and ForestFireVLM}
\label{sec:local-vlm}
For wildfire finetuned VLM baselines shown in Table~\ref{tab:downstream_acc}, we first use the follow prompt to get the 4 tag prediction and then use this together with the prompts in Appendix~\ref{sec:vlm-downstream-conditioned} to perform downstream task prediction. For simulation tag prediction, the prompt used is shown below: 

\begin{PromptBlock}[prompt/code]
Classify this wildfire video into four binary tags from visible evidence only.
For wind use smoke/flame drift, for moisture use dryness and burn behavior,
for slope use terrain incline/relief, for vegetation use fuel density/coverage.
This benchmark is strictly binary. There is no medium or moderate class.
If evidence is mixed, choose the closer of low or high and explain briefly.
Return only JSON with keys wind_regime_tag, moisture_regime_tag, slope_regime_tag, vegetation_regime_tag.
Each key may be either a string label or an object with label and reason.
Every label must be exactly low or high.
\end{PromptBlock}

\paragraph{Message assembly.}
\begin{PromptBlock}[prompt/wrapper]
[text]   <prompt or prompt + retry suffix>
[text]   Frame 1.
[image]  <frame 1>
...
[text]   Frame n.
[image]  <frame n>
\end{PromptBlock}

\subsection{OpenAI Candidate-Reranking Prompt}
\label{sec:candidate-reranking}
Here we show the prompt we used for GPT reranking which correspond to the result shown in Table~\ref{tab:tag_accuracy}.

\begin{PromptBlock}[prompt/code]
Score how well each candidate video's known four-tag bundle matches the query wildfire video.
Use only visible evidence from the query frames and the candidate frames.
The candidate tags are the known labels for that candidate, and the candidate video is a visual anchor for those labels.
Score each candidate independently on a 0 to 100 scale, where 100 means the candidate's full four-tag bundle is a very strong match to the query and 0 means it is a very poor match.
Consider wind, moisture, slope, and vegetation together.
Return one numeric score and one short reason for each candidate.
\end{PromptBlock}

\paragraph{Candidate text template.}
\begin{PromptBlock}[prompt/wrapper]
Candidate video <index>. Return item_id=<candidate_id> exactly in your response.
Known labels: wind_regime_tag=<wind>,
moisture_regime_tag=<moisture>,
slope_regime_tag=<slope>,
vegetation_regime_tag=<vegetation>.
\end{PromptBlock}

\paragraph{Message assembly.}
\begin{PromptBlock}[prompt/wrapper]
[input_text]  <candidate-scoring instruction block above>
[input_text]  Query video.
[input_text]  You must return exactly one scored entry for every candidate item_id shown below. Do not omit any candidate. Use the short item_id tokens exactly as written, such as C01, C02, and C03.
[input_text]  Query frame 1.
[input_image] <query frame 1>
...
[input_text]  Candidate video 1. Return item_id=C01 exactly in your response. Known labels: ...
[input_text]  Candidate 1 frame 1.
[input_image] <candidate 1 frame 1>
...
\end{PromptBlock}

\subsection{Baselines with Labeled Reference Videos}
\label{app:labeled-examples}

We augment direct VLM inference with 16 labeled reference videos per
query, one for each combination of the four binary environmental
driver tags. Reference frames are accompanied by ground-truth tags
and descriptions; the query's simulator metadata is withheld.
Table~\ref{tab:labeled-examples} reports exact four-tag accuracy and
the proportions of queries with at least three or two correct tags.
The strongest exact-match baseline is GPT-5 mini at 27.27\%, compared
with 51.5\% for the full memory system in
Table~\ref{tab:tag_accuracy}. Model versions differ between some query-only and example-augmented baselines, so cross-group comparisons are not a matched-model ablation of adding examples.

\begin{table}[h]
\centering
\small
\caption{Environmental driver prediction with 16 labeled reference
videos per query. All entries are percentages.}
\label{tab:labeled-examples}
\begin{tabular}{@{}lrrr@{}}
\toprule
Model & All four & At least three & At least two \\
\midrule
GPT-4.1 mini & 24.27 & 57.65 & 91.92 \\
GPT-5 mini & 27.27 & 66.67 & 95.96 \\
Claude Sonnet 4.5 & 23.23 & 63.64 & 95.96 \\
Gemini 2.5 Pro & 26.26 & 75.76 & 97.98 \\
\bottomrule
\end{tabular}
\end{table}

\section{Qwen-Embedding Retrieval and Hybrid Rerank Prompts}
\label{sec:qwen-rerank}

\subsection{Our Reranking Method}
The main hybrid retrieval build referenced throughout the downstream pipeline is the Qwen-embedding-backed retrieval setup, followed by GPT candidate scoring over the retrieved shortlist.

\subsection{Optional Qwen3-VL Reranker Instruction}
We also tested an optional Qwen3-VL-2B reranker (instead of using GPT API) with the default instruction:

\begin{PromptBlock}[prompt/task]
Given a search query, retrieve relevant candidates that answer the query.
\end{PromptBlock}

\subsubsection{Qwen3-VL Reranker Message Formatting}
When the Qwen3-VL reranker is enabled, each query-document pair is formatted as follows.

\paragraph{System message.}
\begin{PromptBlock}[prompt/system]
Judge whether the Document meets the requirements based on the Query and the Instruct provided.
Note that the answer can only be "yes" or "no".
\end{PromptBlock}

\paragraph{User message template.}
\begin{PromptBlock}[prompt/user]
<Instruct>: Given a search query, retrieve relevant candidates that answer the query.
<Query>:
<query video and/or query text>

<Document>:
<candidate video and optional candidate prompt text>
\end{PromptBlock}

\paragraph{Practical note.}
The reranker comparison in Appendix~\ref{app:reranker-ablation} reports the added model-choice and timing evaluation.

\subsection{Reranker Accuracy and Latency}
\label{app:reranker-ablation}

Table~\ref{tab:reranker-ablation} compares reranker accuracy and scoring
latency. GPT-5 and Gemini 2.5 Pro both achieve 51.52\% exact four-tag
accuracy, with lower mean scoring latency for GPT-5 in the tested
configurations. Qwen3-VL-Reranker-2B runs locally on an H200 and is
faster but less accurate. API-based and local execution are distinct
deployment conditions, so these measurements are not a
hardware-controlled comparison of model efficiency. Preparation plus
scoring is reported separately from scorer-only latency; neither is
the end-to-end multi-agent pipeline runtime.

\begin{table}[h]
\centering
\small
\setlength{\tabcolsep}{3pt}
\caption{Reranker ablation. Latencies are in seconds and exact four-tag
accuracy is a percentage. Mean and median refer to scorer latency in seconds.}
\label{tab:reranker-ablation}
\begin{tabular}{@{}llrrrr@{}}
\toprule
Reranker & Execution & Mean & Median &
\shortstack{Prep.\\+ scorer} & \shortstack{Four-tag\\accuracy} \\
\midrule
GPT-5 & API & 9.18 & 8.60 & 11.80 & 51.52 \\
Claude Sonnet 4.5 & API & 12.30 & 11.97 & 14.59 & 46.46 \\
Gemini 2.5 Pro & API & 22.24 & 21.59 & 24.53 & 51.52 \\
Qwen3-VL-Reranker-2B & Local H200 & 5.79 & 5.16 & 7.97 & 40.40 \\
\bottomrule
\end{tabular}
\end{table}

\subsection{Retrieval-Size Ablation}
\label{app:retrieval-size}

Table~\ref{tab:retrieval-size} reports the exact four-tag hit rate and
per-attribute Hit@\(k\) as the number of retrieved videos increases
from one to four. Exact four-tag hit rate increases from 51.52\% to
69.70\%. These are retrieval-coverage measurements, not final
predictions from a multi-agent system using larger retrieval sets.

\begin{table}[h]
\centering
\small
\setlength{\tabcolsep}{4pt}
\caption{Retrieval-size ablation. Hit rates are percentages and marginal
gains are percentage-point increases in exact four-tag hit rate.
These are retrieval metrics, not downstream prediction accuracies.}
\label{tab:retrieval-size}
\begin{tabular}{@{}rrrrrrr@{}}
\toprule
$k$ & \shortstack{Exact\\four-tag hit} &
\shortstack{Marginal\\gain} & Wind & Moisture & Slope & Vegetation \\
\midrule
1 & 51.52 & -- & 74.75 & 83.84 & 82.83 & 89.90 \\
2 & 57.58 & +6.06 & 79.80 & 88.89 & 88.89 & 91.92 \\
3 & 64.65 & +7.07 & 81.82 & 94.95 & 92.93 & 92.93 \\
4 & 69.70 & +5.05 & 87.88 & 95.96 & 95.96 & 97.98 \\
\bottomrule
\end{tabular}
\end{table}

\section{Downstream Six-Task Specialist Prompts}
\label{sec:downstream-specialists}
Here we show the system prompts and context blocks used for the downstream
six-task wildfire behavior inference agents in Section~\ref{sec:method_multiagent_reporting}. The fixed multi-agent pipeline
keeps three specialists: a direct visual specialist, a lookup-memory specialist,
and a combined-memory specialist. Unanimous specialist fields pass through
directly, and the GPT judge resolves only disputed fields in the final fixed run.

\subsection{Downstream Task List}
\label{sec:task_list}

\begin{PromptBlock}[prompt/task]
Predict all of these task keys:
- spread_fast_bin: Fast Spread Risk: choose `fast_or_very_fast` if the query likely shows fast or very fast spread. Otherwise choose `slow_or_moderate`.
- peak_high_bin: High Peak Burning Area: choose `high` if peak burning area is likely at least 90,000 m^2, else `low`.
- burnscar_high_bin: High Final Burnscar: choose `high` if final burnscar area is likely at least 300,000 m^2, else `low`.
- vegetation_mean_canopy_cover_pct_median_bin: High Canopy Cover: choose `high` if canopy cover is likely above the train-set median (~2.61%), else `low`.
- rapid_growth_and_large: Rapid Growth + Large Burn: choose `high` only if the query likely has fast or very fast spread and also likely ends with final burnscar area at least 300,000 m^2. Otherwise choose `low`.
- severe_fire_behavior: Severe Fire Behavior: choose `high` if at least 2 of these 3 are likely true: fast/very_fast spread, peak burning area at least 90,000 m^2, final burnscar area at least 300,000 m^2. Otherwise choose `low`.
\end{PromptBlock}

\subsection{Visual Report Agent}
\label{sec:visual-task-rubrics}

\paragraph{Visual Report Agent System Prompt}
\begin{PromptBlock}[prompt/system]
You are the direct visual specialist in a wildfire multi-agent downstream inference pipeline. Use only the query video frames and the task rubric; do not assume the retrieved memory prior is correct, and do not use train labels. Your role is to provide calibrated visual evidence, not to agree with memory specialists. Be conservative for thresholded area tasks: high labels require broad, continuous, visible evidence. Do not infer final burnscar or high canopy from smoke density alone. Return only JSON matching the requested schema.
\end{PromptBlock}

\paragraph{Visual Report Agent Task Rubrics}
\begin{PromptBlock}[prompt/task]
Predict all downstream task keys from visual evidence only:
- spread_fast_bin: Fast Spread Risk: choose `fast_or_very_fast` only when the active flame front visibly expands or translates substantially over the sampled frames. Choose `slow_or_moderate` for compact, slowly changing, or ambiguous fronts.
- peak_high_bin: High Peak Burning Area: choose `high` only when active burning visibly covers a broad continuous landscape-scale area likely >= 90,000 m^2. Choose `low` when the active flaming footprint is localized or uncertain.
- burnscar_high_bin: High Final Burnscar: choose `high` only when later frames show a large continuous blackened/burned footprint likely >= 300,000 m^2. Do not infer high burnscar from smoke alone.
- vegetation_mean_canopy_cover_pct_median_bin: High Canopy Cover: choose `high` only for clear forest/tree canopy or repeated dense tree patches. Choose `low` for mostly grass, shrub, bare, or sparse isolated trees. Avoid overcalling high from small tree clusters.
- rapid_growth_and_large: Rapid Growth + Large Burn: choose `high` only when both visible fast spread and a large final burned footprint are supported. If either condition is uncertain, choose `low`.
- severe_fire_behavior: Severe Fire Behavior: choose `high` only if at least two visual indicators are supported: fast spread, large active burning area, large final burnscar. Choose `low` if only one or none are visually supported.
Output a label and a short visual reason for each task.
\end{PromptBlock}

\paragraph{Message assembly.}
\begin{PromptBlock}[prompt/code]
[input_text]  <visual specialist system prompt>
[input_text]  <visual specialist task rubric>
[input_text]  Query video to evaluate.
[input_text]  Query frame 1.
[input_image] <query frame 1, detail=low>
...
[input_text]  Query frame n.
[input_image] <query frame n, detail=low>
\end{PromptBlock}

\subsection{Lookup- and Combined-Memory Specialist Shared System Prompt}

The lookup-memory and combined-memory specialists use the shared memory-specialist prompt. The memory prior is treated as soft context rather
than ground truth, and query frames remain primary evidence.

\begin{PromptBlock}[prompt/system]
Estimate downstream wildfire behavior attributes for the query video. Use the query frames as primary evidence. You are also given a hybrid-memory prior from a retrieved similar case, including retrieved latent tags and metadata. Treat the memory prior as soft context, not ground truth for the query. If the memory prior conflicts with visible query evidence, trust the query frames. Return only JSON matching the requested schema.
\end{PromptBlock}

\subsection{Lookup-Memory Specialist Context}

\begin{PromptBlock}[prompt/code]
Predict all of these task keys:
- spread_fast_bin: Fast Spread Risk: choose `fast_or_very_fast` if the query likely shows fast or very fast spread. Otherwise choose `slow_or_moderate`.
- peak_high_bin: High Peak Burning Area: choose `high` if peak burning area is likely at least 90,000 m^2, else `low`.
- burnscar_high_bin: High Final Burnscar: choose `high` if final burnscar area is likely at least 300,000 m^2, else `low`.
- vegetation_mean_canopy_cover_pct_median_bin: High Canopy Cover: choose `high` if canopy cover is likely above the train-set median (~2.61%), else `low`.
- rapid_growth_and_large: Rapid Growth + Large Burn: choose `high` only if the query likely has fast or very fast spread and also likely ends with final burnscar area at least 300,000 m^2. Otherwise choose `low`.
- severe_fire_behavior: Severe Fire Behavior: choose `high` if at least 2 of these 3 are likely true: fast/very_fast spread, peak burning area at least 90,000 m^2, final burnscar area at least 300,000 m^2. Otherwise choose `low`.

Hybrid lookup prior from the top-1 retrieved tag bundle follows.
The lookup prior is train-derived from the retrieved tags. Use it as a soft prior only.
If the query frames clearly disagree, trust the query frames over the lookup prior.
{
  "retrieved_item_id": "<top1 item_id>",
  "retrieval_score": <top1 score>,
  "retrieved_tags": {
    "wind_regime_tag": "<low|high>",
    "moisture_regime_tag": "<low|high>",
    "slope_regime_tag": "<low|high>",
    "vegetation_regime_tag": "<low|high>"
  },
  "lookup_prior_from_retrieved_tags": {
    "spread_fast_bin": "<label>",
    "peak_high_bin": "<label>",
    "burnscar_high_bin": "<label>",
    "vegetation_mean_canopy_cover_pct_median_bin": "<label>",
    "rapid_growth_and_large": "<label>",
    "severe_fire_behavior": "<label>"
  }
}
\end{PromptBlock}

\paragraph{Message assembly.}
\begin{PromptBlock}[prompt/wrapper]
[input_text]  <shared memory-specialist system prompt>
[input_text]  <lookup specialist context block>
[input_text]  Query video to evaluate.
[input_text]  Query frame 1.
[input_image] <query frame 1, detail=low>
...
[input_text]  Query frame n.
[input_image] <query frame n, detail=low>
\end{PromptBlock}

\subsection{Combined-Memory Agent Context}
The combined-memory specialist receives the query frames together with the combined-memory context block and the lookup-memory context block. The combined block includes the retrieved prompt, environmental metadata, and fire-behavior summaries shown below.

\begin{PromptBlock}[prompt/code]
Predict all of these task keys:
- spread_fast_bin: Fast Spread Risk: choose `fast_or_very_fast` if the query likely shows fast or very fast spread. Otherwise choose `slow_or_moderate`.
- peak_high_bin: High Peak Burning Area: choose `high` if peak burning area is likely at least 90,000 m^2, else `low`.
- burnscar_high_bin: High Final Burnscar: choose `high` if final burnscar area is likely at least 300,000 m^2, else `low`.
- vegetation_mean_canopy_cover_pct_median_bin: High Canopy Cover: choose `high` if canopy cover is likely above the train-set median (~2.61%), else `low`.
- rapid_growth_and_large: Rapid Growth + Large Burn: choose `high` only if the query likely has fast or very fast spread and also likely ends with final burnscar area at least 300,000 m^2. Otherwise choose `low`.
- severe_fire_behavior: Severe Fire Behavior: choose `high` if at least 2 of these 3 are likely true: fast/very_fast spread, peak burning area at least 90,000 m^2, final burnscar area at least 300,000 m^2. Otherwise choose `low`.

Hybrid memory prior from the top-1 retrieved similar case follows.
Use it as a soft prior only. Do not copy it blindly if the query frames disagree.
{
  "retrieved_item_id": "<top1 item_id>",
  "retrieval_score": <top1 score>,
  "retrieved_tags": {
    "wind_regime_tag": "<low|high>",
    "moisture_regime_tag": "<low|high>",
    "slope_regime_tag": "<low|high>",
    "vegetation_regime_tag": "<low|high>"
  },
  "retrieved_prompt": "<retrieved training prompt>",
  "retrieved_environment_metadata": {
    "wind_speed_mean_mph": <float>,
    "wind_dir_mean_cardinal": "<cardinal direction>",
    "environment_moisture": <float>,
    "slope_mean_deg": <float>,
    "vegetation_mean_canopy_cover_pct": <float>,
    "elevation_relief_m": <float>,
    "burnable_fraction": <float>
  },
  "retrieved_fire_behavior_metadata": {
    "peak_burning_area_m2": <float>,
    "final_burnscar_area_m2": <float>,
    "burnscar_ratio": <float>,
    "max_radial_spread_mps": <float>
  }
}
\end{PromptBlock}
\paragraph{Message assembly.}
\begin{PromptBlock}[prompt/wrapper]
[input_text]  <shared memory-specialist system prompt>
[input_text]  <combined-memory specialist context block>
[input_text]  <lookup specialist context block>
[input_text]  Query video to evaluate.
[input_text]  Query frame 1.
[input_image] <query frame 1, detail=low>
...
[input_text]  Query frame n.
[input_image] <query frame n, detail=low>
\end{PromptBlock}

\subsection{Specialist Response Schema}

All three specialists return a label and a short reason for each downstream task.

\begin{PromptBlock}[prompt/code]
{
  "type": "json_schema",
  "name": "hybrid_downstream_prediction",
  "schema": {
    "type": "object",
    "properties": {
      "spread_fast_bin": {
        "type": "object",
        "properties": {
          "label": {"type": "string", "enum": ["slow_or_moderate", "fast_or_very_fast"]},
          "reason": {"type": "string"}
        },
        "required": ["label", "reason"],
        "additionalProperties": false
      },
      "peak_high_bin": {
        "type": "object",
        "properties": {
          "label": {"type": "string", "enum": ["low", "high"]},
          "reason": {"type": "string"}
        },
        "required": ["label", "reason"],
        "additionalProperties": false
      },
      "burnscar_high_bin": {
        "type": "object",
        "properties": {
          "label": {"type": "string", "enum": ["low", "high"]},
          "reason": {"type": "string"}
        },
        "required": ["label", "reason"],
        "additionalProperties": false
      },
      "vegetation_mean_canopy_cover_pct_median_bin": {
        "type": "object",
        "properties": {
          "label": {"type": "string", "enum": ["low", "high"]},
          "reason": {"type": "string"}
        },
        "required": ["label", "reason"],
        "additionalProperties": false
      },
      "rapid_growth_and_large": {
        "type": "object",
        "properties": {
          "label": {"type": "string", "enum": ["low", "high"]},
          "reason": {"type": "string"}
        },
        "required": ["label", "reason"],
        "additionalProperties": false
      },
      "severe_fire_behavior": {
        "type": "object",
        "properties": {
          "label": {"type": "string", "enum": ["low", "high"]},
          "reason": {"type": "string"}
        },
        "required": ["label", "reason"],
        "additionalProperties": false
      }
    },
    "required": [
      "spread_fast_bin",
      "peak_high_bin",
      "burnscar_high_bin",
      "vegetation_mean_canopy_cover_pct_median_bin",
      "rapid_growth_and_large",
      "severe_fire_behavior"
    ],
    "additionalProperties": false
  }
}
\end{PromptBlock}

\subsection{Downstream Task Prediction Prompt for GPT, Qwen2-Wildfire-2B and ForestFireVLM}
\label{sec:vlm-downstream-conditioned}

This section shows the prompt used in Table~\ref{tab:downstream_acc} for the three VLMs (Direct GPT Query, Qwen2-Wildfire-2B and ForestFireVLM) on six-task downstream prediction when their own predicted four-tag state is provided as soft context together with the video frames. Here, the task list variable is the string from Appendix~\ref{sec:visual-task-rubrics}.

\begin{PromptBlock}[prompt/code]
Estimate downstream wildfire behavior attributes for the query video.
Use the query frames as primary evidence.
You are also given a predicted latent 4-tag wildfire state from an earlier classifier for this same query.
Treat those predicted tags as soft context only, not ground truth.
If the predicted tags conflict with visible evidence, trust the query frames.
Return only JSON with the requested task keys.

Predict all of these task keys:
<task list above>

Allowed labels:
- spread_fast_bin: slow_or_moderate, fast_or_very_fast
- peak_high_bin: low, high
- burnscar_high_bin: low, high
- vegetation_mean_canopy_cover_pct_median_bin: low, high
- rapid_growth_and_large: low, high
- severe_fire_behavior: low, high

Predicted latent 4-tag state for this query (soft context only):
{"wind_regime_tag":"...","moisture_regime_tag":"...","slope_regime_tag":"...","vegetation_regime_tag":"..."}

Return only JSON with exactly these keys and string labels.
{"spread_fast_bin":"slow_or_moderate","peak_high_bin":"low","burnscar_high_bin":"low","vegetation_mean_canopy_cover_pct_median_bin":"low","rapid_growth_and_large":"low","severe_fire_behavior":"low"}
\end{PromptBlock}

\paragraph{Message assembly.}

\begin{PromptBlock}[prompt/wrapper]
[text]   <downstream-conditioned prompt or prompt + retry suffix>
[text]   Frame 1.
[image]  <frame 1>
...
[text]   Frame n.
[image]  <frame n>
\end{PromptBlock}

\subsection{Multi-Agent Judge Prompts}
\label{sec:multiagent-judge}
The templates below mention both disagreement-only and audit-all use. The reported fixed pipeline follows the disagreement-only routing described in Appendix~\ref{sec:impl_multiagent_reporting}: only disputed task keys are requested, while unanimous fields are retained. Terms such as ``calibrated'' in these prompts express requested behavior, not an empirically validated uncertainty calibration. The prompts are retained verbatim to document the evaluated system.
\paragraph{Judge System Prompt}
\begin{PromptBlock}[prompt/system]
You are the judge agent in a multi-agent wildfire downstream inference pipeline. Three specialist agents have already analyzed the query: a direct visual specialist, a lookup-prior specialist, and a combined-memory specialist. Your job is to audit and resolve the requested downstream tasks. These may be only disputed tasks, or an audit-all pass. Use the query frames as primary evidence for directly visible spread pace and obvious footprint extent. Use the top-k hybrid-memory consensus and retrieved metadata as calibrated soft prior evidence, not ground truth. Do not merely vote between specialists. Apply task-specific reliability: the direct visual specialist is useful for spread and for conservative negative evidence, but area-threshold outcomes and canopy cover should lean more on strong top-k memory consensus. You may override unanimous specialist outputs when they conflict with strong memory consensus or clear visual evidence. For high labels on peak area, final burnscar, rapid growth, or severe behavior, require either broad visible evidence or strong memory support. Return only JSON matching the requested schema.
\end{PromptBlock}

\paragraph{Judge Context Block}
The judge received only the task keys where the three specialists disagreed. All
six labels and reasons from every specialist were still included as context.

\begin{PromptBlock}[prompt/wrapper]
Audit and resolve these requested tasks:
- <task_key_1>: <task description from TASK_SPECS>
- <task_key_2>: <task description from TASK_SPECS>
...

Previously agreed tasks, if any. Treat them as context, not immutable labels during audit-all runs:
<json.dumps(agreed_labels, ensure_ascii=True, sort_keys=True)>

Specialist outputs follow. These are inputs, not votes:
{
  "Direct visual specialist": {
    "predictions": {
      "spread_fast_bin": "<label>",
      "peak_high_bin": "<label>",
      "burnscar_high_bin": "<label>",
      "vegetation_mean_canopy_cover_pct_median_bin": "<label>",
      "rapid_growth_and_large": "<label>",
      "severe_fire_behavior": "<label>"
    },
    "reasons": {
      "spread_fast_bin": "<reason>",
      "peak_high_bin": "<reason>",
      "burnscar_high_bin": "<reason>",
      "vegetation_mean_canopy_cover_pct_median_bin": "<reason>",
      "rapid_growth_and_large": "<reason>",
      "severe_fire_behavior": "<reason>"
    }
  },
  "Lookup-prior specialist": {
    "predictions": {...},
    "reasons": {...}
  },
  "Combined-memory specialist": {
    "predictions": {...},
    "reasons": {...}
  }
}

Top-k hybrid-memory consensus follows. This is more reliable than any single retrieved item, but still only a soft prior:
{
  "candidate_count": 5,
  "task_consensus": {
    "spread_fast_bin": {
      "winner": "<label>",
      "confidence": <float>,
      "distribution": {
        "slow_or_moderate": <float>,
        "fast_or_very_fast": <float>
      }
    },
    "peak_high_bin": {
      "winner": "<label>",
      "confidence": <float>,
      "distribution": {
        "low": <float>,
        "high": <float>
      }
    },
    "burnscar_high_bin": {
      "winner": "<label>",
      "confidence": <float>,
      "distribution": {
        "low": <float>,
        "high": <float>
      }
    },
    "vegetation_mean_canopy_cover_pct_median_bin": {
      "winner": "<label>",
      "confidence": <float>,
      "distribution": {
        "low": <float>,
        "high": <float>
      }
    },
    "rapid_growth_and_large": {
      "winner": "<label>",
      "confidence": <float>,
      "distribution": {
        "low": <float>,
        "high": <float>
      }
    },
    "severe_fire_behavior": {
      "winner": "<label>",
      "confidence": <float>,
      "distribution": {
        "low": <float>,
        "high": <float>
      }
    }
  },
  "weighted_metric_means": {
    "peak_burning_area_m2": <float>,
    "final_burnscar_area_m2": <float>,
    "max_radial_spread_mps": <float>,
    "vegetation_mean_canopy_cover_pct": <float>
  },
  "top_candidates": [
    {
      "item_id": "<train/sample_id>",
      "hybrid_score": <float>,
      "retrieved_tags": {...},
      "lookup_prior": {
        "spread_fast_bin": "<label>",
        "peak_high_bin": "<label>",
        "burnscar_high_bin": "<label>",
        "vegetation_mean_canopy_cover_pct_median_bin": "<label>",
        "rapid_growth_and_large": "<label>",
        "severe_fire_behavior": "<label>"
      },
      "environment": {...},
      "fire_behavior": {...}
    }
  ]
}

Decision guidance: prefer the query frames for visible spread evidence; use strong memory consensus for less directly visible outcomes such as canopy and final burnscar; do not default to any single specialist. If the direct visual specialist is conservative but memory strongly supports a high threshold outcome, explain why you accept or reject that memory prior.
\end{PromptBlock}

\paragraph{Message assembly.}
\begin{PromptBlock}[prompt/wrapper]
[input_text]  <judge system prompt>
[input_text]  Audit and resolve these requested tasks:
              - <only task keys where the three specialists disagree>
[input_text]  Previously agreed tasks, if any. Treat them as context, not immutable labels during audit-all runs: <agreed_labels JSON>
[input_text]  Specialist outputs follow. These are inputs, not votes: <specialist_payload JSON>
[input_text]  Top-k hybrid-memory consensus follows. This is more reliable than any single retrieved item, but still only a soft prior: <memory_consensus JSON>
[input_text]  Decision guidance: prefer the query frames for visible spread evidence; use strong memory consensus for less directly visible outcomes such as canopy and final burnscar; do not default to any single specialist. If the direct visual specialist is conservative but memory strongly supports a high threshold outcome, explain why you accept or reject that memory prior.
[input_text]  Query video to adjudicate.
[input_text]  Query frame 1.
[input_image] <query frame 1, detail=low>
...
[input_text]  Query frame n.
[input_image] <query frame n, detail=low>
\end{PromptBlock}

\paragraph{Judge Response Schema}
\begin{PromptBlock}[prompt/code]
{
  "type": "json_schema",
  "name": "multiagent_downstream_resolution",
  "schema": {
    "type": "object",
    "properties": {
      "<requested_task_key>": {
        "type": "object",
        "properties": {
          "label": {
            "type": "string",
            "enum": <TASK_SPECS[task_key]["label_enum"]>
          },
          "reason": {
            "type": "string"
          }
        },
        "required": ["label", "reason"],
        "additionalProperties": false
      }
    },
    "required": ["<all requested task keys>"],
    "additionalProperties": false
  }
}
\end{PromptBlock}

\subsubsection{Leave-One-Out Judge Prompt Variant}
For the ablations studies shown in Table~\ref{tab:method_accuracy}, the judge prompt was changed only in the specialist roster sentence. The remainder of the prompt stayed the same.

\begin{PromptBlock}[prompt/code]
You are the judge agent in a multi-agent wildfire downstream inference pipeline. <Two specialist agents have / Specialist agents have> already analyzed the query: <joined list of the remaining specialist names>. Your job is to audit and resolve the requested downstream tasks. These may be only disputed tasks, or an audit-all pass. Use the query frames as primary evidence for directly visible spread pace and obvious footprint extent. Use the top-k hybrid-memory consensus and retrieved metadata as calibrated soft prior evidence, not ground truth. Do not merely vote between specialists. Apply task-specific reliability: the direct visual specialist is useful for spread and for conservative negative evidence, but area-threshold outcomes and canopy cover should lean more on strong top-k memory consensus. You may override unanimous specialist outputs when they conflict with strong memory consensus or clear visual evidence. For high labels on peak area, final burnscar, rapid growth, or severe behavior, require either broad visible evidence or strong memory support. Return only JSON matching the requested schema.
\end{PromptBlock}

\section{HTML Report Generation Prompt}
\label{sec:html-report}
To generate the narrative in Figure~\ref{fig:report}, we use GPT to summarize the predicted environmental tags and six report-field labels. The structured inference records retain retrieval and routing information, while the narrative prompt below deliberately omits these internal details. Its use of ``operational'' describes the requested writing style, not an established deployment capability. Reports and response suggestions are illustrative research outputs requiring expert validation before operational use.

\subsection{System Prompt for Report Generation}
\begin{PromptBlock}[prompt/system]
You write concise operational wildfire reports. Follow the user's schema exactly and output valid JSON only.
\end{PromptBlock}

\subsection{User Prompt Template for Report Generation}
The user prompt is instantiated per case with the predicted location, severity label, four latent tags, and six downstream task labels.

\begin{PromptBlock}[prompt/user]
Write a concise wildfire operational report for one case.

Use the downstream task predictions and four-tag latent state exactly as the evidence source.
Do not mention ground truth, validation, agents, routing, retrieval, memory, or model internals.
Do not mention confidence scores.
Write in a polished, professional style suitable for a final user-facing wildfire report.

Location: Longitude <lon>, Latitude <lat>
Severity: <severity_label>

Predicted 4-tag state:
- Wind: <tag label> (confidence <c>)
- Moisture: <tag label> (confidence <c>)
- Terrain: <tag label> (confidence <c>)
- Vegetation: <tag label> (confidence <c>)

Downstream task predictions:
- Fast Spread Risk: <prediction text>
- High Peak Burning Area: <prediction text>
- High Final Burnscar: <prediction text>
- High Canopy Cover: <prediction text>
- Rapid Growth + Large Burn: <prediction text>
- Severe Fire Behavior: <prediction text>

Return strict JSON with exactly these keys:
- fire_spread_prediction
- damage_prediction
- suggestion

Each value must be one compact paragraph of 2-4 sentences.
\end{PromptBlock}

\paragraph{Message assembly.}
\begin{PromptBlock}[prompt/wrapper]
[system]  You write concise operational wildfire reports. Follow the user's schema exactly and output valid JSON only.
[user]    <instantiated report-generation prompt above>
\end{PromptBlock}

\paragraph{Output contract.}
\begin{PromptBlock}[prompt/schema]
{
  "fire_spread_prediction": "<compact 2-4 sentence paragraph>",
  "damage_prediction": "<compact 2-4 sentence paragraph>",
  "suggestion": "<compact 2-4 sentence paragraph>"
}
\end{PromptBlock}

\subsection{Structured Report Example}
\label{app:report-examples}
Figure~\ref{fig:report} shows a representative report from the complete pipeline, combining visual evidence, predicted environmental tags, downstream predictions, retrieved metadata, natural-language reasoning, and response suggestions. These reports illustrate the output format rather than establish operational validity.

\section{Structured Output Contracts}
\subsection{Four-Tag Classification Schema}
\begin{PromptBlock}[prompt/schema]
{
  "wind_regime_tag":        {"label": "low|high", "reason": "<short visible-evidence sentence>"},
  "moisture_regime_tag":    {"label": "low|high", "reason": "<short visible-evidence sentence>"},
  "slope_regime_tag":       {"label": "low|high", "reason": "<short visible-evidence sentence>"},
  "vegetation_regime_tag":  {"label": "low|high", "reason": "<short visible-evidence sentence>"}
}
\end{PromptBlock}

\subsection{Candidate-Reranking Schema}
\begin{PromptBlock}[prompt/schema]
{
  "candidates": [
    {"item_id": "C01", "match_score": <0-100 number>, "reason": "<short reason>"},
    {"item_id": "C02", "match_score": <0-100 number>, "reason": "<short reason>"},
    ...
  ]
}
\end{PromptBlock}

\subsection{Downstream Specialist Agent / Judge Schema}
\begin{PromptBlock}[prompt/schema]
{
  "spread_fast_bin": {"label": "slow_or_moderate|fast_or_very_fast", "reason": "<short reason>"},
  "peak_high_bin": {"label": "low|high", "reason": "<short reason>"},
  "burnscar_high_bin": {"label": "low|high", "reason": "<short reason>"},
  "vegetation_mean_canopy_cover_pct_median_bin": {"label": "low|high", "reason": "<short reason>"},
  "rapid_growth_and_large": {"label": "low|high", "reason": "<short reason>"},
  "severe_fire_behavior": {"label": "low|high", "reason": "<short reason>"}
}
\end{PromptBlock}

\section{Cross-Generator and External UAV Evaluation}
\label{app:generalization-tests}

\subsection{Cross-Generator Evaluation}

We use LTX-2.3 + AVControl as a video-generation backbone independent
of the original VACE generator. We evaluate both LTX-generated queries
against VACE-generated memory and VACE-generated queries against
LTX-generated memory. Results are reported in
Table~\ref{tab:generalization-compact}a. Our method reaches 41.00\%
and 44.44\% accuracy, respectively, compared with 28.00\% and 27.27\%
for the strongest direct VLM baselines in the two query sets.
This tests changing the video-generation backbone, not changing the
physical simulator or procedural proxy renderer. It isolates a visual-generator change from the separate real-UAV tests below.

\subsection{External UAV Evaluation}

Real wildfire videos jointly annotated with environmental drivers and fire behavior are scarce, motivating the simulation-based construction of labeled reference episodes. We complement the controlled physical-label evaluation with real-UAV tests using available annotations: smoke detection, fire presence, and visible fire-area assessment. These tests assess observable monitoring tasks; the annotations used here do not jointly specify the four environmental drivers needed to validate physical-driver recovery.

\paragraph{Boreal Forest Fire.}
We evaluate smoke detection using the smoke-detection labels in Boreal Forest Fire~\cite{pesonen2025boreal}.
Table~\ref{tab:boreal-results} reports accuracy, precision, recall,
and F1. Our method, Claude 4, and ForestFireVLM-7B reach 100\% on all
four metrics; this evaluation does not distinguish among those methods.

\begin{table}[h]
\centering
\small
\caption{Boreal Forest Fire smoke detection. All entries are percentages.}
\label{tab:boreal-results}
\begin{tabular}{@{}lrrrr@{}}
\toprule
Method & Accuracy & Precision & Recall & F1 \\
\midrule
Ours & 100.0 & 100.0 & 100.0 & 100.0 \\
GPT-5 & 94.4 & 90.0 & 100.0 & 94.7 \\
Claude 4 & 100.0 & 100.0 & 100.0 & 100.0 \\
ForestFireVLM-7B & 100.0 & 100.0 & 100.0 & 100.0 \\
Qwen2-Wildfire-2B & 94.4 & 100.0 & 88.9 & 94.1 \\
\bottomrule
\end{tabular}
\end{table}

\paragraph{FireMan-UAV-RGBT.}
We evaluate binary fire presence on FireMan-UAV-RGBT~\cite{kularatne2024fireman}. Overall accuracy is accompanied
by balanced accuracy, which assigns equal weight to fire and no-fire
classes in this imbalanced evaluation. Table~\ref{tab:fireman-results}
reports the complete per-method results.

\begin{table}[h]
\centering
\small
\caption{FireMan-UAV-RGBT fire-presence detection. Scores are on a
0--1 scale. Bold indicates the highest score.}
\label{tab:fireman-results}
\begin{tabular}{@{}lrr@{}}
\toprule
Method & Balanced accuracy & Accuracy \\
\midrule
GPT-5 mini & 0.7646 & 0.8173 \\
Claude Sonnet 4.6 & 0.8563 & 0.9135 \\
Gemini 2.5 Pro & 0.8125 & 0.9135 \\
ForestFireVLM-7B & 0.8625 & 0.9231 \\
Qwen2-Wildfire-2B & 0.8375 & 0.7500 \\
\midrule
Ours & \textbf{0.9625} & \textbf{0.9423} \\
\bottomrule
\end{tabular}
\end{table}

\paragraph{FLAME.}
For FLAME~\cite{shamsoshoara2021aerial}, we derive reference visible fire area by summing the fire segmentation regions.
In the ranking task, each prompt provides ten videos and asks the model
to rank them from the largest to the smallest fire region.
Table~\ref{tab:flame-results} reports mean AP, largest-fire Hit@1,
and fire-area NDCG. Our method has the highest NDCG and ties Claude
Sonnet 4.6 on Hit@1; Gemini 2.5 Pro has the highest mean AP.
These annotations concern visible fire regions, not a complete
physical characterization of fire severity.

\begin{table}[h]
\centering
\small
\caption{FLAME fire-area evaluation. Scores are on a 0--1 scale.
Bold indicates the highest score, including ties.}
\label{tab:flame-results}
\begin{tabular}{@{}lrrr@{}}
\toprule
Method & Mean AP & \shortstack{Largest-fire\\Hit@1} &
\shortstack{Fire-area\\NDCG} \\
\midrule
GPT-5 & 0.930 & 0.658 & 0.947 \\
Claude Sonnet 4.6 & 0.946 & \textbf{0.850} & 0.973 \\
Gemini 2.5 Pro & \textbf{0.952} & 0.564 & 0.891 \\
ForestFireVLM-7B & 0.939 & 0.767 & 0.957 \\
Qwen2-Wildfire-2B & 0.844 & 0.547 & 0.883 \\
\midrule
Ours & 0.950 & \textbf{0.850} & \textbf{0.984} \\
\bottomrule
\end{tabular}
\end{table}

\section{Additional Visual Results}
\label{app:additional-visuals}
\begin{figure}[htbp]
\centering
\includegraphics[width=\linewidth]{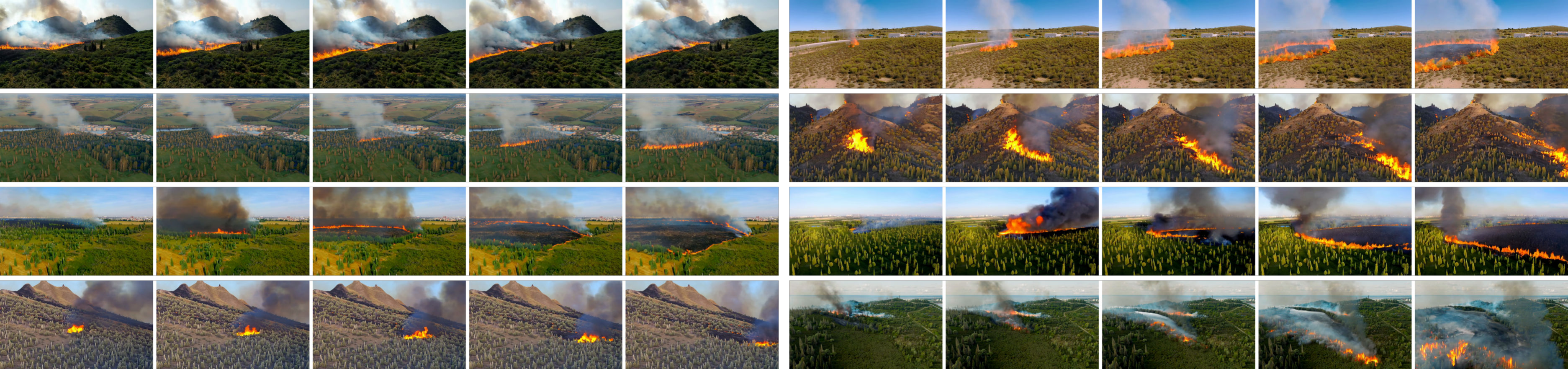}
\caption{Additional generated wildfire video frames conditioned on 2D simulations through procedural 3D proxies.}
\label{fig:more_vlm}
\end{figure}

Figure~\ref{fig:more_vlm} provides additional generated-video examples. Figure~\ref{fig:per_tag_accuracy} reports the per-attribute comparison complementing Table~\ref{tab:tag_accuracy} in the main text.

\end{document}